\documentclass[preprint,12pt]{elsarticle}
\usepackage{latexsym, graphicx, epsfig, amsmath, amssymb, amsfonts}
\usepackage{amsmath}
\usepackage{amssymb}
\usepackage{graphicx}
\usepackage{subfig}
\usepackage{tabu}
\usepackage{adjustbox}
\usepackage{bbold}
\usepackage{bm}
\usepackage{textcomp}
\usepackage[thinc]{esdiff}
\usepackage[dvipsnames]{xcolor}
\usepackage{comment}
\usepackage{textgreek}
\usepackage{caption}
\usepackage{lineno}
\usepackage{multirow}
\usepackage{adjustbox}

\usepackage{siunitx}
\usepackage{xcolor}
\usepackage{booktabs}
\usepackage{caption}
\usepackage{float}
\usepackage{color,soul}

\journal{Applied Mechanics Reviews}
\begin{document}
\begin{frontmatter}

\title{Recent Advances and Applications of Machine Learning in Experimental Solid Mechanics: A Review}

\author{Hanxun Jin\textsuperscript{a,d}}
\author{Enrui Zhang\textsuperscript{c}}
\author{Horacio D. Espinosa\textsuperscript{a,b}\corref{cor1}}
\cortext[cor1]{Corresponding authors. E-mail:  espinosa@northwestern.edu (H.D.E.)}

\address[1]{Department of Mechanical Engineering, Northwestern University, Evanston, IL 60208}
\address[2]{Theoretical and Applied Mechanics Program, Northwestern University, Evanston, IL 60208}
\address[3]{Division of Applied Mathematics, Brown University, Providence, RI 02912}
\address[4]{Present address: Division of Engineering and Applied Science, California Institute of Technology, Pasadena, CA 91125}

\begin{abstract}

For many decades, experimental solid mechanics has played a crucial role in characterizing and understanding the mechanical properties of natural and novel materials. Recent advances in machine learning (ML) provide new opportunities for the field, including experimental design, data analysis, uncertainty quantification, and inverse problems. As the number of papers published in recent years in this emerging field is exploding, it is timely to conduct a comprehensive and up-to-date review of recent ML applications in experimental solid mechanics. Here, we first provide an overview of common ML algorithms and terminologies that are pertinent to this review, with emphasis placed on physics-informed and physics-based ML methods. Then, we provide thorough coverage of recent ML applications in traditional and emerging areas of experimental mechanics, including fracture mechanics, biomechanics, nano- and micro-mechanics, architected materials, and 2D material. Finally, we highlight some current challenges of applying ML to multi-modality and multi-fidelity experimental datasets and propose several future research directions. This review aims to provide valuable insights into the use of ML methods as well as a variety of examples for researchers in solid mechanics to integrate into their experiments.
\end{abstract}

\end{frontmatter}



\newpage
\section{Introduction}


Over the years, the field of experimental solid mechanics has kept evolving because of the continuous demand to characterize and understand the mechanical properties of natural and novel artificial metamaterials \cite{RN94,RN98}. There are two main motivations for performing experiments in solid mechanics: (1) to provide experimental observations that can be used to advance universal mechanics laws; (2) to measure unknown mechanical properties of materials and structures, e.g., modulus, strength, phase changes, inelasticity, failure strain, and fracture toughness, under prescribed boundary or loading conditions. Measured fields and properties guide the construction of constitutive laws and interpretation of underlying physics. For example, in the 1670s, Robert Hooke performed experiments about adding masses to springs and discovered a linear relationship between forces and their produced deformations, laying the foundations of elastic mechanics \cite{RN111}. Another seminal example is A. A. Griffith’s rupture experiments on glass specimens, which led to his energetic interpretation of fracture in the 1920s \cite{RN7}. Experimental results supported his theory that “a crack will propagate when the reduction in potential energy that occurs due to crack growth is greater than or equal to the increase in surface energy due to the creation of new free surfaces.” This launched the century-old discipline of fracture mechanics.

Throughout the history of experimental solid mechanics, various apparatuses have been invented to measure mechanical properties, from quasi-static testing (e.g., the universal tensile testing machine \cite{RN97}) to high strain-rate testing (e.g., Kolsky bars \cite{RN95} and plate impact facility \cite{RN96}). In the past two decades, advances in nanomechanics tools like the nanoindenter and microelectromechanical systems (MEMS) for in situ microscopy testing \cite{RN101,RN103} have enabled nanoscale characterization of advanced materials \cite{RN102}. Likewise, various experimental measurement techniques have been developed, from local methods, e.g., strain gauges and displacement transducers \cite{RN99}, to full-field methods, such as high-resolution and high-speed imaging systems \cite{RN100}. These technical innovations provide an extensive and ever-increasing amount of data collected during a single experiment. To analysis data from full-field measurement, new analysis techniques like Moiré interferometry \cite{RN104}, digital image/volume correlations (DIC/DVC) \cite{RN105, RN106}, electronic speckle pattern interferometry (ESPI) \cite{RN107} have been developed to extract mechanical properties and deformation fields from experiments. Furthermore, inverse methods have been used to extract constitutive behavior and identify imperfections  \cite{RN185, RN186}. Indeed, the combination of experimental mechanics with fast and robust computational algorithms for inverse analysis has been growing in importance since it enables new approaches to mechanical property identification, from fracture properties under extreme conditions to anisotropic properties of biological tissues to superior mechanical properties of nanoarchitected and 2D materials. 

Recently, the concept of materials by design \cite{RN121} has been advanced for the design of multi-functional architected materials \cite{RN30} and 2D materials/devices \cite{RN109} exhibiting unprecedented performance. Such progress was possible due to the rapid development of modern experimental mechanics techniques, which include fabrication processes, e.g., Additive Manufacturing (AM) \cite{RN108} and Chemical Vapor Deposition (CVD), as well as the development of high-throughput testing methodologies \cite{RN187, RN188, RN189}. In such research strategy, experimental solid mechanics plays an essential role in providing valuable training and validation experimental data for extraction of physically-inspired reduced order models as well as advancing understanding of fabrication process-mechanical property relationships. Therefore, combining novel and intelligent algorithms together with advances in fabrication and experimental characterization methods has the potential to achieve a paradigm shift in discovering multi-functional and architected materials.

Recent advances in Machine Learning (ML) \cite{RN112}, in particular, Deep Learning (DL) \cite{lecun2015deep}, offer the opportunity to expand the field of experimental solid mechanics when combined with rapid data processing and inverse approaches. ML has played a significant role in computer science applications and technologies like computer vision \cite{krizhevsky2017imagenet}, natural language processing \cite{hinton2012deep}, and self-driving cars \cite{ramos2017detecting}. In engineering and applied physics disciplines, ML has been widely used in various areas of materials science, including material microstructure design \cite{RN35}, microscopic imaging detection \cite{RN120}, and force-field development \cite{RN119}. Several comprehensive reviews \cite{choudhary2022recent, mueller2016machine, butler2018machine, RN116, RN117} have thoroughly surveyed the potential of ML in materials science. In Solid Mechanics, ML has been successfully employed in a wide range of applications, such as constructing surrogate models for constitutive modeling \cite{RN209, RN208}, advancing multiscale modeling \cite{RN124, RN125}, designing architected materials \cite{RN123}, extracting unknown mechanical parameters \cite{RN77}, or obtaining the internal material information from externally measured fields \cite{ni2021deep, zhang2020physics, RN18, song2023identifying}. In these applications, most ML frameworks were trained on synthetic data from computational methods. Therefore, applying these ML frameworks to train and validate sparse and noisy experimental data with high fidelity and modality could require cautious quantification of uncertainty \cite{RN126, brodnik2023perspective}, from both experimental data and ML architectures like hyperparameters, optimization method, and overparameterization. For this purpose, uncertainty quantification methods like Bayesian methods and deep ensembles can be employed. Moreover, employing ML in the experimental design data process could not only potentially identify material properties, which could not be revealed otherwise, but also inspire experimentalists to develop new experimental techniques with metrology capable of big-data generation with high information content. With broad community interest, as reflected by the increasing number of publications in this field, it is timely to conduct a contemporary and thorough review of recent advances in the use of ML in experimental solid mechanics. Though this review will be focused on experimental aspects, multidisciplinary approaches, including computational and theoretical mechanics, as well as materials science, would be desirable to address engineering applications.

\emph{This review is dedicated to Prof. Kyung-Suk Kim, on his 70’s birthday, to celebrate his seminal contributions to the fields of experimental and theoretical mechanics.} The review is planned as follows. In \textbf{Section 2}, we will briefly review some key ML algorithms that can be employed in experimental mechanics. Then, in \textbf{Section 3}, we will review some recent progress of ML applications in experimental mechanics, covering the broad areas of fracture mechanics, biomechanics, nano- and micro-mechanics, mechanics of architected materials, and fracture toughness of 2D materials. In \textbf{Section 4}, we will discuss how to properly select ML models for experimental data with multi-fidelity and multi-modality. Finally, in \textbf{Section 5}, we will close our review with a discussion of potential future research directions. Please note that ML has also been extensively applied to AM and full-field optical measurement, which are two other important fields in experimental solid mechanics. Since extensive state-of-the-art reviews have been conducted in these two fields (see \cite{wang2020machine,jin2020machine, qin2022research} for AM and \cite{zuo2022deep} for full-field optical measurement), we will not cover these topics extensively in this review. 

\section{ML framework for experimental solid mechanics}

ML methods use algorithms mimicking human intelligence to perform optimization tasks for particular goals \cite{RN112}. Since the early development of NNs and back-propagation algorithms \cite{rumelhart1986learning} in the 1980s, the research field in ML has evolved significantly, leading to the discovery of various network architectures with distinct operational principles. Therefore, before applying these ML algorithms to experimental solid mechanics, it is crucial to understand their architectures, working principles, and potential limitations. Such fundamental knowledge will enable researchers to effectively understand these algorithms and utilize them appropriately in specific applications. 

In principle, ML methods can be divided into three main categories: supervised learning, unsupervised learning, and reinforcement learning. In supervised learning, the algorithm learns the mapping between the input dataset and their ground-truth labels, while in unsupervised learning, the algorithm aims to identify patterns and features in the data without being explicitly trained on labeled examples. In reinforced learning, an agent receives feedback in terms of rewards or punishments for each action, and then uses these feedbacks to improve future decision-making capabilities. In addition, there are other ML methods such as semi-supervised learning. In semi-supervised learning, the ML algorithm is trained with both labeled and unlabeled datasets. The model is first trained by the labeled data and then uses the model to label the unlabeled data. Examples of supervised learning algorithms include linear regression, decision trees, and neural networks (NNs). Examples of unsupervised learning algorithms include principal component analysis, K-mean clustering, and spectral clustering. Examples of RL include policy gradient and Q-learning. Examples of semi-supervised learning include generative models like generative adversarial networks (GANs). For more details on the fundamentals of ML methods, readers are referred to textbooks \cite{goodfellow2016deep} and online courses.

ML methods are typically data-driven, that is, the model is trained/informed by large datasets in forms including images, texts, audio, and so on. For example, the revolutionary AI chatbot software, ChatGPT \cite{openai2021chatgpt}, was trained based on a transformer language model \cite{vaswani2017attention} that uses self-attention mechanisms, allowing the model to weigh input words at different positions to predict the following words. In engineering and physics disciplines, however, many problems can be well-defined by some underlying physical laws such as partial differential equations (PDEs). For example, in solid mechanics, the equilibrium equations (a set of PDEs) define the kinematics of solid continuum bodies. Moreover, the physical laws themselves provide valuable priori temporal or spatial information, which can be integrated into the ML framework during training. To this end, the concept of Physics-Informed Neural Networks (PINNs) \cite{raissi2019physics} has been proposed by the Karniadakis group at Brown University. This breakthrough framework has paved a new pathway to solving physical-law-governed forward and inverse problems without the need to collect big datasets. With the notions of data-driven and physics-informed ML methods, we may interpolate between these two ends, formulating a spectrum of scientific ML methods that may be developed for experimental solid mechanics. Depending on the size of the dataset we could obtain, and how much physics is embedded in the algorithms, the scientific ML approaches can be categorized into three scenarios depicted in \textbf{Fig. 1}: (I) physics-informed learning method \cite{raissi2019physics, RN128}, (II) physics-based data-driven method, and (III) purely data-driven method. The second scenario of the physics-based data-driven approach, in particular, can be employed when a problem is too complex to be concretely described by a set of PDEs, such as cohesive fracture problems, but the training dataset can be relatively easily obtained from massive computational simulations like Finite Element Analysis (FEA) or Molecular dynamics (MD) within a reasonable time. 

In the following subsections, we will briefly review some scientific ML algorithms in the spectrum of methods and introduce terminologies that are useful for applications in experimental solid mechanics. In Section 2.1, we will discuss one of the unsupervised ML methods, the clustering method. We will then introduce various types of NNs in Sections 2.2-2.6. Then, in Sections 2.7 and 2.8, we will discuss RL and Bayesian inference. Lastly, methods used in the scientific ML community will be discussed in Sections 2.9-2.10.

\begin{figure}[!ht]
    \centering
    \includegraphics[width=1.0\textwidth]{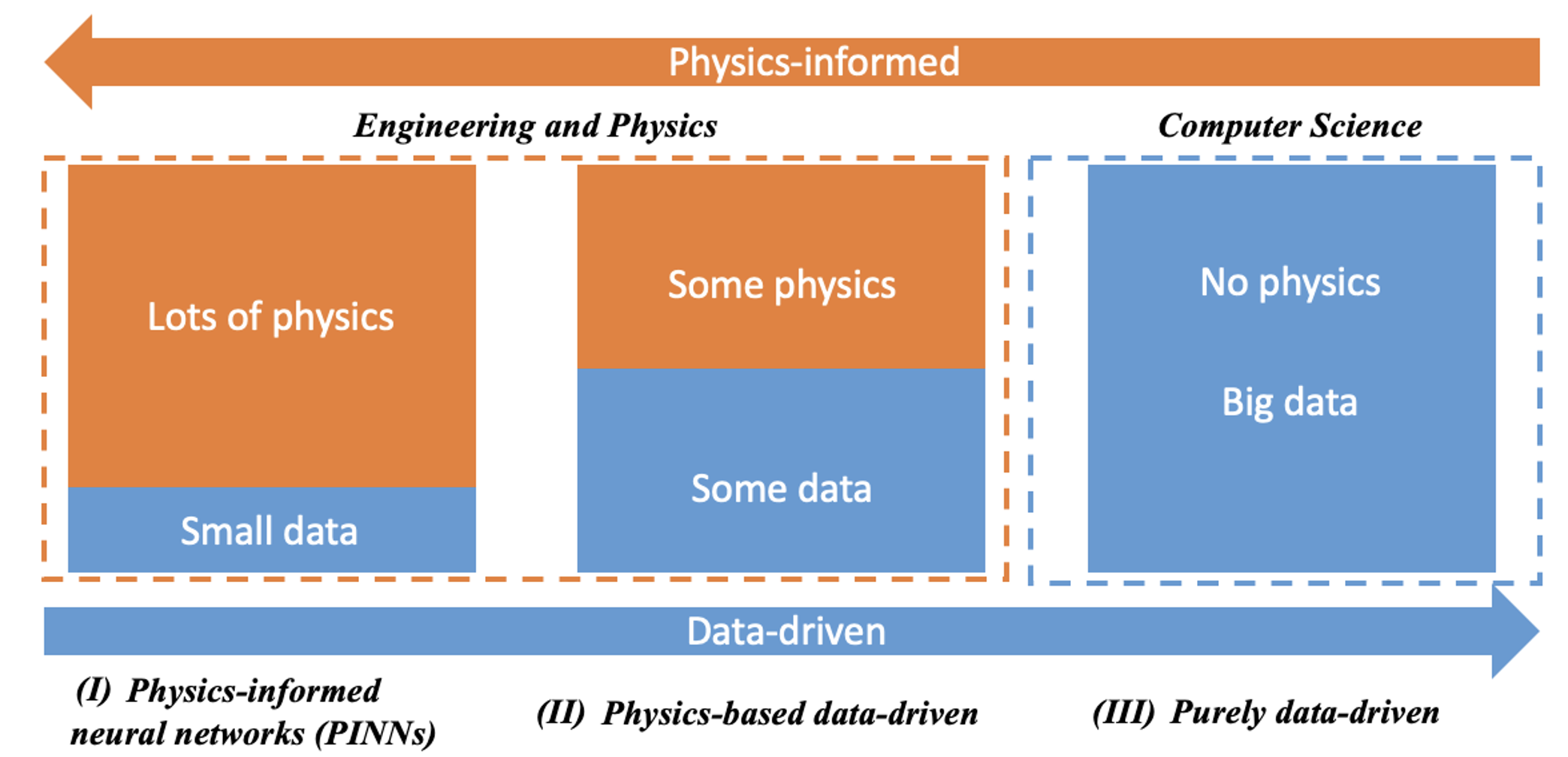}
    \caption{Schematics of three ML approaches based on available physics and data: (I) Physics-informed neural networks (PINNs); (II) Physics-based data-driven; and (III) Purely data-driven (Reproduced with permission from Ref. \cite{RN21}. Copyright 2021 by Hanxun Jin). Figure idea from Karniadakis et al. \cite{RN128}}
    \label{fig:abstract}
\end{figure}

\subsection{Clustering}
Clustering is an unsupervised ML technique that involves identifying data structure and grouping similar datasets into several clusters. The most common clustering algorithm is the K-means method, which groups total dataset into K clusters by minimizing the total distance between the data and computed cluster centroids. When K-means algorithm fails to cluster the data due to data non-linearities or complexity, the spectral clustering algorithm \cite{von2007tutorial}, which uses the spectral properties of the dataset to determine clusters, can be employed. The principle of spectral clustering is to transform a complex dataset into a low-dimensional representation by using the spectrum (eigenvalues) of the similarity matrix of the data. Then, the low-dimensional dataset can be clustered using traditional clustering techniques, such as K-means clustering. In experimental solid mechanics, spectral clustering is particularly applicable to acoustic signals such as acoustic emissions from mechanical response \cite{muir2021damage}. For example, Muir et al. \cite{muir2021machine} applied the spectral clustering technique to identify the damage mechanisms of SiC/SiC composites based on frequency information of acoustic emission signals.

\subsection{Neural Networks (NNs)}
NNs are an ML algorithm inspired by the function and structure of the human brain. They consist of an input layer, interconnected layer(s), and an output layer. These layers are interconnected with nodes, which are similar to neurons. Through passing information by interconnected layers with activation functions \cite{goodfellow2016deep} (such as Sigmoid, Softmax, and Rectified Linear Unit (ReLu) functions \cite{agarap2018deep}), the NN can learn a nonlinear mapping between inputs and outputs. There is a variety of NNs, including Dense Neural Networks (DNN), Convolutional Neural Networks (CNN), Generative Adversarial Networks (GANs), Recurrent Neural Networks (RNNs), transformer, and autoencoder,  which are useful in inverse problems and data generation for solid mechanics. Among these structures, the simplest architecture of NNs is the Fully-Connected Neural Networks, where all neurons in every layer are connected to all neurons in the adjacent layers. In the next subsections, we will briefly overview these NN structures and their current and potential applications in experimental solid mechanics.

\subsection{Convolutional Neural Networks (CNNs)}
CNN is a type of DL algorithm commonly used for image classification and feature extractions \cite{lecun1998gradient,lecun1989backpropagation}. The input images are processed through convolutional layers, subsequently passed through the polling layer, and fully connected layers for feature reduction and filtering. Multiple convolutional layers can be applied to increase the complexity of the feature extraction. CNN can be useful for both regression and classification of image-based experimental data. For example, CNN can be used for material property identification and microstructure characterization from experimental images \cite{RN129}. Furthermore, CNN can analyze images obtained from full-field measurement techniques like interferometric and DIC data. Using CNN for feature extraction from interferometric fringes can bypass the need for a conventional fringe unwrapping process while also increasing the feature extraction accuracy. For example, Jin et al. \cite{RN11} employed a CNN-based DL framework to extract dynamic cohesive properties and fracture toughness of polyurea directly from image-shearing interferometric fringes. Kaviani and Kolinski \cite{RN130} developed a CNN-based DL framework to convert fringes from Fizeau interferometry with low resolution into frustrated total internal reflection (FTIR) images with high resolution while studying droplet impact. Another important application for CNN in experimental mechanics is to analyze DIC data. DIC is a powerful full-field measurement tool to analyze local displacement and strain distribution \cite{RN105}. Many advanced DIC techniques like q-factor-based DIC (qDIC) \cite{RN132} and augmented-Lagrangian DIC (ALDIC) \cite{RN133, RN134} were developed to increase the accuracy, efficiency, and robustness of strain field calculation. Recently, Yang et al. \cite{RN131} showed the pre-trained CNN-based DL algorithms from synthetic data can accurately predict end-to-end measurement of displacement and strain fields from experimental speckle images. CNNs were also used by Espinosa and co-workers in the study of cell morphology upon biomolecular delivery into cells using localized electroporation \cite{RN245} and the localization of single cells within a population for single-cell gene editing \cite{RN191}.

\subsection{Recurrent neural networks (RNNs)}
RNNs are a type of NNs, that have been successfully used for processing sequential datasets such as natural language \cite{goodfellow2016deep}. Unlike feedforward NNs, RNNs can retain information about previous inputs. One of the seminal RNNs models is long short-term memory (LSTM) networks \cite{hochreiter1997long}, which was developed to solve the vanishing gradient problem for simple RNNs. RNNs have been widely applied in a variety of applications, such as speech recognition, machine translation, and natural language processing. In solid mechanics applications, RNNs have been used in structural health monitoring such as crack path detection. For example, in Buehler’s group, LSTM models were trained based on atomistic simulations to predict the fracture patterns of crystalline solids \cite{hsu2020using} and 2D materials \cite{lew2021deep,lew2021deep2}. Furthermore, RNNs have been successfully applied to model solids with plastic behaviors due to their capability to deal with time-dependent data. For example, Mozaffar et al. \cite{mozaffar2019deep} applied RNNs to model path-dependent plasticity for complex microstructures.

\subsection{Graph Neural Networks (GNNs)}
GNNs \cite{RN139} are a type of ML method that is usually employed on graph-structured data, which can be considered as a collection of nodes and edges. The nodes represent entities, and the edges represent the relationships between the entities. Therefore, GNNs are well-suited for handling problems where the relationships between entities are abstract, non-sequential, and highly interconnected. They have been successfully applied to a broad range of applications, including recommender systems \cite{RN140}, social networks \cite{RN142}, drug discovery \cite{RN141}, material property prediction \cite{RN146}, and protein nature frequency prediction \cite{RN143}. In solid mechanics, GNNs have been employed to characterize and design complex mechanical materials and structures based on the graph representation of their microstructure and/or crystallography. For example, Guo and Buehler \cite{RN144} applied GNN to design architected material through a semi-supervised approach. Recently, Xue et al. \cite{RN145} developed a GNN-based framework to predict the nonlinear dynamics of soft mechanical metamaterials. Hestroffer et al. \cite{hestroffer2023graph} applied GNNs to predict mechanical properties like stiffness and yield strength of polycrystalline materials. Thomas et al. \cite{thomas2023materials} employed GNNs to represent fatigue features in polycrystalline materials and predict high cycle fatigue damage formation.

\subsection{Generative Adversarial Networks (GANs) and Conditional Generative Adversarial Networks (cGANs)}

GAN and cGAN are ML algorithms initially used in the field of computer vision, like image generation. GAN consists of two NN, a generator and a discriminator, trained simultaneously in a game-theory-based framework to generate new synthetic datasets that mimic the original datasets \cite{RN135}. First, the generator creates new data from random Gaussian noise. Then, the discriminator evaluates if the generated data is true or false. The training finishes when a Nash Equilibrium \cite{RN136} is reached, where the generator produces authentic data that the discriminator could not identify as fake. Therefore, GAN can be a promising ML algorithm for training data generation and augmentation in solid mechanics, such as generating synthetic structures of metamaterials, which can be used for simulations and experiments. For example, GAN was applied to generate complex architected materials, among which many of them have extreme mechanical properties without prior knowledge \cite{RN46}. Similarly, GAN was used to generate 3D microstructures from 2D sliced images, which are as authentic as real microstructures of battery electrodes \cite{RN137}. While GANs have shown tremendous success in data augmentation, there are some common limitations such as mode collapse and training instability \cite{salimans2016improved}, which needs attention to generate high-quality synthetic data. Besides careful hyperparameters tunning, one possible solution is to introduce physics-informed constraints during training. For instance, in the context of microstructural generation, incorporating statistical information, such as geometry descriptors \cite{cang2017microstructure}, as a constraint, can help ensure the generated structures adhere to certain physical properties.

cGAN is an extension of GAN where the generation of data is conditioned on additional inputs or labels \cite{mirza2014conditional}. Beyond its data generation capability, cGAN can also be used for image-based end-to-end mapping and inverse problems, such as topology optimization \cite{RN138}. Applying cGAN to inverse problems enables the generation of solutions that are consistent with the desired properties. cGAN has been applied to image-to-image transitions like image inpainting \cite{isola2017image} or image semantic segmentation \cite{rezaei2018conditional}. For its application in Solid Mechanics, cGAN has been employed to inversely identify the material modulus map from the given strain/stress images \cite{ni2021deep} or predict strain and stress distributions for complex composites \cite{yang2021end, yang2021deep}. Furthermore, cGAN has been successfully applied to experimental data inpainting when partial experimental data is missing \cite{RN11}. 

\subsection{Reinforcement learning (RL)}
RL is a type of ML technique in which an agent learns to make the optimum decisions by interacting with its environment to achieve a specific goal [105]. During the training, the agent receives feedback in the form of rewards or penalties based on its actions, enabling it to adjust its time-dependent behavior to maximize the cumulative rewards. Therefore, there are three important aspects in RL: agent, environment, and reward. The RL methods can be categorized into value-based methods, policy-based methods, and actor-critic methods. Recent advances in deep reinforcement learning (DRL) \cite{mousavi2018deep}, have further expanded the capability of RL for sophisticated optimization tasks. Common DRL algorithms include Deep Q-networks (DQN) \cite{mnih2015human}, and deep policy gradient \cite{lillicrap2015continuous}. DRL has been widely used in game playing such as AlphaGo \cite{silver2016mastering}, autonomous robotics \cite{singh2022reinforcement}, chemical design \cite{popova2018deep}, and fluid flow optimization \cite{garnier2021review}. As its application in solid mechanics, RL has been employed in a wide range of optimization tasks, such as materials design and structural optimization. In materials design, the geometry and material parameters can be considered as the agent and the desired mechanical responses like stress-strain relationship can be considered as the environment. By sampling the design space, the agent can receive rewards when the structure reaches the desired properties. For example, Sui et al. \cite{sui2021deep} applied the DQN to design biphasic materials based on desired homogenized properties. More recently, Nguyen et al. \cite{nguyen2022synthesizing} developed a DL method by combining GANs and RL to generate realistic 3-dimensional microstructures with user-defined structural properties.

\subsection{Bayesian inference}
Bayesian inference, named after Thomas Bayes, is a statistical method that allows us to quantify uncertainty of unknown parameters based on observed data \cite{box2011bayesian}. The fundamental concept of Bayesian inference involves integrating prior knowledge about an unknown parameter, with the likelihood of the observed data given that parameter, to generate the posterior probability distribution. In contrast to ML methods that focus on identifying the optimal model parameters, i.e., the maximum likelihood estimate, Bayesian inference provides a comprehensive description of the uncertainty surrounding the parameters, allowing for robust uncertainty quantification. However, Bayesian inference also has certain drawbacks. First, the selection of prior distribution can be subjective and may influence the results, especially in the case of limited data. Second, computational cost can be intensive, especially for high-dimensional spaces, as it involves sampling and computing likelihood across the entire parameter space. Despite these challenges, Bayesian inference has been widely used in various fields, such as finance, environmental science, signal processing, and healthcare. In solid mechanics, Bayesian inference has been widely used to identify material parameters and quantify their uncertainty from various experimental data, such as uniaxial stress-strain data \cite{rappel2020tutorial}, force-indentation depth data from nanoindentation \cite{RN85}, and resonance frequency data from resonant ultrasound spectroscopy \cite{rossin2021bayesian,rossin2022single}.

\subsection{Physics-informed Neural Networks (PINNs)}

Since the landmark paper \cite{raissi2019physics} published by the Karniadakis group at in 2019 (arXiv preprint in 2017 \cite{RN192}), PINNs have played a significant role in scientific ML in engineering and physics disciplines \cite{RN128}. The fundamental idea of PINNs is to apply a NN to approximate the solution to a physical problem, where the governing physical principles (mathematically expressed by PDEs) are enforced as prior knowledge by penalizing the residuals of PDEs, similar to \cite{RN275, RN276}. PINNs have been successfully employed to solve scientific problems in a widespan of engineering disciplines such as heat transfer \cite{RN148}, fluid dynamics \cite{RN147, RN193, RN149}, wave propagation \cite{RN153}, nano-optics \cite{RN150}, AM \cite{RN156}, and biomaterials \cite{RN151}. Due to the injection of physical laws into the learning algorithm, PINNs require substantially less amount of data than data-driven neural network approaches for achieving similar predictive capability. For example, in the study on PINNs for fluid dynamics \cite{RN193}, by exploiting several snapshots of the concentration field of passive scalars, PINNs are capable of predicting the velocity and pressure fields. To fulfill the same task in a data-driven approach without integrating the fluid physics, one may need at least hundreds of paired snapshots of concentration, velocity, and pressure fields as training data. To facilitate the usage of PINNs in the research community, Lu et al. implemented various PINN algorithms in an open-source Python library called DeepXDE \cite{RN152}.

As for the applications in Solid Mechanics, PINNs have been successfully applied to both forward problems (i.e., solving boundary- and initial-value problems) and inverse problems (e.g., material characterization and defect detection). For example, Henkes et al. \cite{RN154} employed PINNs to model micromechanics for linear elastic materials. Haghighat et al. \cite{RN194} applied PINNs to build surrogate models for elastostatics and elastoplastic solids. Bastek and Kochmann \cite{RN195} employed PINNs to model the small-strain response of shell structures. Zhang et al. \cite{zhang2020physics, RN18} demonstrated that PINNs could effectively identify the inhomogeneous material and geometry distribution under plane-strain conditions. Due to the injection of mechanical laws into the learning algorithm, PINNs function with a limited amount of data. Though most of the current PINNs frameworks in solid mechanics were demonstrated using generated synthetic data as proofs of concept, these frameworks can be applied to experimental mechanics seamlessly where continuum mechanics theories apply.

\subsection{Neural Operators}

NNs are not only universal approximator of continuous function \cite{RN196}, but also of nonlinear continuous operator \cite{RN158}. Neural operators are neural network models that learn operators, which map functions to functions, such as differential operator, integral operator, and solution operator for parameterized PDEs. Learning operators is especially important in engineering and physics since many problems involving the relationship between functions, not simply between parameters. Within the scope of solid mechanics, examples of functions include displacement field, stress field, load distribution, stiffness distribution, crack propagation path, and so on. Karniadakis group proposed a neural operator architecture called Deep Operator Network (DeepONet) \cite{RN157}. DeepONet consists of two parts of NNs: a branch net to encode the discrete input function space and a trunk net to encode the domain of the output functions. Since then, some other neural operators have been developed \cite{kovachki2021neural, RN162}. For detailed explanations, adequate comparisons among these algorithms, and comparison between neural operators and PINNs, the readers can refer to the original papers and recent reviews \cite{RN159, RN197}. It is worth noting that, while the original versions of these neural operators are data-driven, physical principles may also be incorporated in a similar way as PINNs, making the neural operators informed by physics in addition to data \cite{RN277, RN278}. In the past few years, neural operators have been extensively applied to diverse engineering problems. In solid mechanics applications, neural operators have been successfully used for elastoplasticity \cite{RN279}, fracture mechanics \cite{RN17}, multiscale mechanics \cite{RN124}, and biomechanics \cite{RN164, RN218, RN165}. 

\newpage
\begin{table}[ht]
\centering

\caption{\textbf{Summary of ML applications in experimental solid mechanics.}}

\resizebox{0.9\textwidth}{!}{
\begin{tabular}{|c|c|c|c|}
    \hline
    Areas & Detailed applications & ML algorithms & Selected references \\ \hline
    \multirow{10}{*}{Fracture Mechanics} & \multirow{2}{*}{Fracture toughness} & NN, Decision trees & \cite{RN9, RN10} \\ \cline{3-4}
    &  & CNN & \cite{RN11} \\ \cline{2-4}
     
    & \multirow{3}{*}{Cohesive parameters} & CNN & \cite{RN11} \\ \cline{3-4}
    & & NN & \cite{RN14, RN13} \\ \cline{3-4}
    & & Deep-green inversion & \cite{wei2022deep}\\ \cline{2-4}

    & \multirow{2}{*}{Crack/flaw detection} & PINN & \cite{RN18}\\ \cline{3-4}
    & & CNN & \cite{RN19,RN20} \\ \cline{2-4}

    & \multirow{2}{*}{Crack path prediction} & LSTM & \cite{RN16,RN24} \\ \cline{3-4}
    & & DeepONet & \cite{RN17} \\ \cline{2-4}

    & Predict fracture instability & Gaussian process regressions & \cite{RN274} \\ \cline{1-4}

    \multirow{10}{*}{Biomechanics} & \multirow{2}{*}{Human motion} & NN & \cite{RN168} \\ \cline{3-4}
    & & CNN & \cite{RN167} \\ \cline{2-4}
    
    & \multirow{3}{*}{Constitutive parameters} & ResNet, CNN & \cite{RN181} \\ \cline{3-4}
    & & NN & \cite{RN183} \\ \cline{3-4}
    & & PINN & \cite{RN151, RN202} \\ \cline{2-4}
    
    & \multirow{4}{*}{Surrogate constitutive model} & Thermodynamics-based NN & \cite{RN209}\\ \cline{3-4}
    & & Constitutive artificial NN (CANN) & \cite{RN208} \\ \cline{3-4}
    & & DeepONet & \cite{RN164, RN220}\\ \cline{3-4}
    & & Neural operator & \cite{RN218}\\ \cline{2-4}

    & Cell Manipulation and Analysis & CNN & \cite{RN245, RN191, RN89, RN246}\\ \cline{1-4}

    \multirow{7}{*}{Micro and Nano-Mechanics} & \multirow{3}{*}{Nanoindentation} & NN & \cite{muliana2002artificial, RN62, RN63} \\ \cline{3-4}
    & & Multi-fidelity NN (MFNN) & \cite{RN77} \\ \cline{3-4}
    & & Bayesian method & \cite{RN77, RN84, RN85, RN81} \\ \cline{2-4}
    
    & AFM & data-driven NN & \cite{RN91} \\ \cline{2-4}
    
    & \multirow{3}{*}{Microstructure characterizations} & CNN, U-Net & \cite{RN229, RN243} \\ \cline{3-4}
    & & Random Forest statistical algorithm & \cite{RN231}\\ \cline{3-4}
    & & cGAN & \cite{ni2021deep, yang2021end} \\ \cline{1-4}

    \multirow{4}{*}{Architected materials} & \multirow{2}{*}{Verify computational design} & CNN, Autoencoder & \cite{RN38} \\ \cline{3-4}
    & & CNN, ResNet & \cite{RN40} \\ \cline{2-4}
    
    & \multirow{2}{*}{Training data generation} & GAN & \cite{RN46, RN244} \\ \cline{3-4}
    & & GNN & \cite{RN144} \\ \cline{1-4}

    \multirow{2}{*}{2D materials} & MD force-field parameterization & Multi-objective optimization algorithm & \cite{zhang2021multi} \\ \cline{2-4}
    & Fracture toughness & Integrated experiment-simulation framework & \cite{zhang2022atomistic} \\
    
    \hline

\end{tabular}
}
\end{table}

\newpage

\section{Applications of ML in experimental solid mechanics}

In this section, we will review recent advancements and applications of ML in experimental solid mechanics, covering a wide range of fields such as fracture mechanics, biomechanics, nano- and micro-mechanics, architected materials, and 2D materials. \textbf{Table 1} summarizes the ML applications, algorithms, and selected key references in these research areas.

\subsection{ML for fracture mechanics}

\begin{figure}[!ht]
    \centering
    \includegraphics[width=1.0\textwidth]{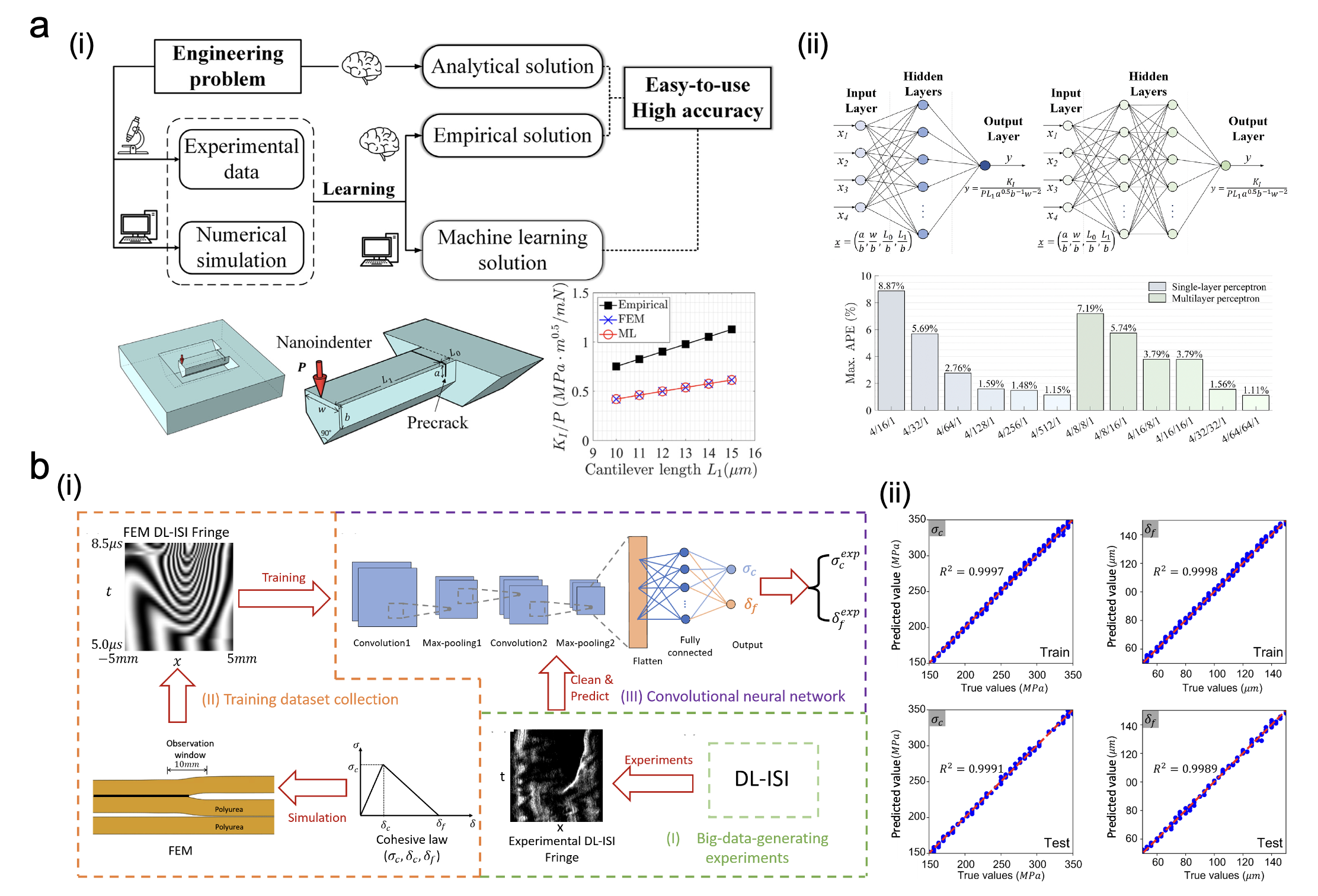}
    \caption{Applications of ML in characterizing fracture cohesive properties. (a) ML solutions can predict accurate fracture toughness comparable to simulations when an analytical solution is not available due to sample complexity: (i) ML framework for engineering problems; (ii) NNs-based prediction of fracture toughness (Reproduced with permission from Ref.  \cite{RN9}. Copyright 2020 by Elsevier). (b) A CNN-based DL algorithm can accurately determine dynamic fracture toughness and cohesive parameters under ultra-high strain rate loading: (i) DL framework for cohesive parameter inversion from dynamic big-data-generating experiments; (ii) Comparison between predicted cohesive parameters and ground-truth.  (Reproduced with permission from Ref. \cite{RN11}. Copyright 2022 by Elsevier).}
    \label{fig:abstract}
\end{figure}

Since the landmark paper from Griffith in 1921 \cite{RN7}, the century-old discipline of fracture mechanics has been established in advancing a wide range of technological advancements, from airplane structural integrity to novel materials with microscale architectures to biomimicry of natural materials. From an experimental viewpoint, there are applications of material failure we summarize here. The first deals with quantifying material intrinsic fracture properties such as fracture toughness or cohesive laws that enable the transfer of knowledge obtained from laboratory tests to engineering applications, e.g., designing new machines or structures with unprecedented fracture resistance. The second addresses the identification of non-visible crack-like defects by providing information such as their location and geometries from nondestructive evaluation (NDE) data. This enables failure and reliability analysis, carried out to prevent catastrophic failure in service.

Interestingly, ML methods can estimate fracture toughness when it cannot be easily measured using traditional methods, e.g., fracture toughness testing, under quasi-static loading, mature experimental protocols based on the American Society for Testing and Materials (ASTM) standard have been established. Readers are referred to a previous review for this subject \cite{RN8}. Typically, testing samples are machined into specific dimensions, and a load is applied to the sample to propagate a crack. Then, experimental data like load, displacement, and crack-tip opening distance (CTOD) are recorded to obtain the fracture toughness based on analytical solutions. Recently, these methods have also been extended to measure the fracture toughness of soft materials like hydrogen \cite{RN272, RN273}. However, such analytic solutions are not applicable when testing samples with more complex or irregular geometries and/or material nonlinearities. For example, Liu et al. \cite{RN9} proposed two ML approaches, decision trees and NNs, to obtain the fracture toughness for micro-fabricated ceramic cantilevers. The ML models mapped geometry descriptors of cantilevers to their fracture toughness calculated from FEA. As shown in \textbf{Fig. 2 (ai)}, when the analytical solution is not accessible due to geometry and material complexity, an ML solution trained from representative and sufficient FEA datasets can overcome these weaknesses and provide accurate fracture toughness with a mean error of 1\% (\textbf{Fig. 2 (aii)}). Furthermore, with knowledge extraction and transfer techniques, the fracture toughness of samples with 3D complexity can be efficiently predicted from simpler 2D simulations \cite{RN10}.

Beyond the fracture toughness prediction, ML can be employed to inversely extract cohesive law parameters (assuming its validity) from experimental measurements consisting of load-displacement curves \cite{RN14, RN13} or full-field measurements. As previously articulated, the prediction performance can be significantly improved by integrating physical governing laws like equilibrium equations into ML training. For example, Wei et al. \cite{wei2022deep} proposed a Green’s function embedded neural network (Deep-Green Inversion) that can extract mixed-mode cohesive zone properties using only far-field displacement data measured from experiments. This method integrates Green’s functions as a physical constraint, hence reducing the amount of training data and increasing the local model accuracy. Other computational frameworks using ML to characterize cohesive behaviors are also helpful in designing new experiments. For example, Liu \cite{RN22} proposed a deep material network (DMN) with cohesive layers, which enables accurate modeling of the material interface in heterogeneous materials. Wang and Sun \cite{RN23} proposed a meta-modeling method that employs deep reinforcement learning to model constitutive behaviors of interfaces.

Another important application of ML in fracture mechanics is the prediction of the crack path given the crack propagation history. Knowing the crack path is helpful to prevent catastrophic material failure by toughening the material along the path. For example, LSTM-based ML models were trained based on atomistic modeling to predict the fracture patterns of crystalline solids \cite{RN15} and 2D materials \cite{RN16, RN24}. The LSTM model is capable of learning the spatial-temporal relations from an atomic resolution of fracture, hence is effective in predicting the crack path. In another application, Goswami et al. \cite{RN17} developed a physics-informed variational formulation of DeepONet (V-DeepONet) to predict the crack path in quasi-brittle materials by mapping the initial crack configuration to the damage and displacement fields. More recently, Worthington and Chew \cite{worthington2023crack} applied NNs to predict the crack path of heterogenous materials by mapping the crack process zone information to the possible crack growth directions based on FEA training employing a micromechanics fracture model. It is also important to mention that current ML applications for crack path prediction are primarily trained and validated using computational datasets obtain from FEA or MD simulations. Thus, it remains uncertain how these methods would perform when applied to real experimental data. In the future, it is necessary to obtain high-fidelity crack propagation data from advanced diagnostic imaging techniques such as in-situ computed tomography \cite{chon2017lamellae, garcia2019using} or in-situ electron microscopy experiments \cite{RN54, RN261,ramachandramoorthy2015pushing}. This will enable researchers to validate and improve the current ML framework for crack path prediction.

When predicting dynamic fracture toughness under ultra-high loading rates, the conventional experimental measurement of load-displacement is not accessible. To this end, Jin et al. \cite{RN11} proposed an ML-assisted big-data-generating experimental framework that can accurately measure the dynamic fracture toughness and cohesive parameters of samples from plate impact experiments. As shown in \textbf{Fig. 2 (bi)}, a cohesive law identification experiment was developed using plate impact, a target polyurea sample containing a half-plane mid-crack, and a novel spatial-temporal interferometer that generates fringes associated with the sample rear surface motion history. By employing a physics-based data-driven method, using FEM, a CNN was trained to correlate the fringe images with corresponding cohesive law parameters. After the CNN was well-trained, the dynamic fracture toughness and cohesive parameters of polyurea were successfully determined from the experimental fringe image (\textbf{Fig. 2 (bii)}). This big-data-generating experiment framework can be easily extended to other mechanics problems, under extreme conditions, such as stress wave induced phase transformations, shear localization, and others where conventional measurement methods are not applicable. Furthermore, quantifying uncertainty for inhomogeneous materials like composites under dynamic loading is crucial to ensure an accurate assessment of dynamic fracture properties. Sharma et al. \cite{RN26} developed an ML framework bridging limited experimental data from advanced experimental techniques and data-driven models like Monte Carlo simulation to quantify the uncertainty of the dynamic fracture toughness of glass-filled epoxy composites.

\begin{figure}[!ht]
    \centering
    \includegraphics[width=0.75\textwidth]{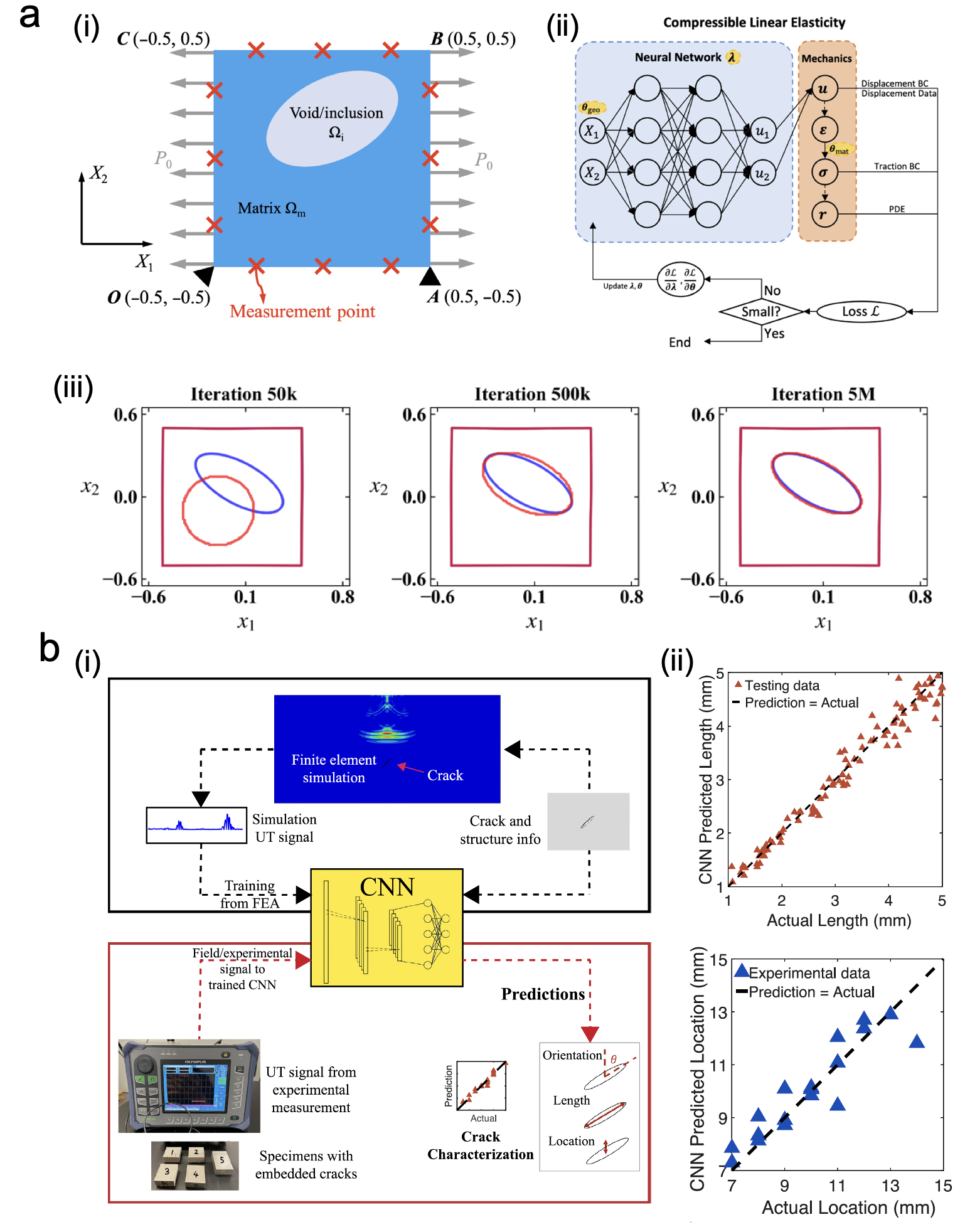}
    \caption{Applications of ML in crack/flaw detection. (a) PINN can identify internal voids/inclusions for linear and nonlinear solids: (i) General setup for geometric and material property identification; (ii) Architectures of PINNs for continuum solid mechanics. (iii) Inference of deformation patterns under different training epochs (Reproduced with permission from Ref. \cite{RN18}.  Copyright 2022, The authors, published by AAAS). (b) FEA simulation-trained CNN was used to determine crack locations and geometry in experiments: (i) Proposed CNN architecture, (ii) CNN predicted crack property compared to ground-truth (Reproduced with permission from Ref. \cite{RN19}. Copyright 2022 by Elsevier).}
    \label{fig:abstract}
\end{figure}

Beyond the fracture properties identification, ML is also effective in predicting internal crack/flaw geometry and locations from the experimental measurement. For example, Zhang et al. \cite{RN18} applied PINNs to identify internal cracks in linear and nonlinear solids. As shown in \textbf{Fig. 3 (ai)}, the framework uses the external boundary conditions as “sensors” to inversely identify the internal crack information under deformation. More importantly, the PINNs framework directly integrates the underlying physics, such as material compressibility and equilibrium equations, into the loss function (\textbf{Fig. 3 (aii)}), hence significantly reducing the amount of data required during training while achieving high prediction accuracy (\textbf{Fig. 3 (aiii)}). Using this PINNs method, the cracks in nonlinear solids like materials with elastoplastic behaviors can also be accurately identified \cite{RN27}. Moreover, by employing a similar PINNs framework, other material properties like modulus distribution can also be inferred \cite{zhang2020physics}. To predict material strength in solids with microcracks, Xu et al. \cite{RN28} trained an ML framework that maps the crack distribution morphology to the strength calculated from micromechanics theory. This framework can effectively predict the strength of solids with randomly distributed microcracks. ML can also be used with non-destructive measurement methods to characterize internal cracks without deforming the material. For example, Niu and Srivastava  \cite{RN19, RN20} used FEM-trained CNN to accurately identify the internal crack information from ultrasound measurement (\textbf{Fig. 3 (bi)}). Impressively, the simulation-trained CNN can make predictions of crack shape and locations on experimental data with a Mean Average Percentage Error (MAPE) of around 5\% (\textbf{Fig. 3 (bii)}). The accuracy of the prediction can be attributed to the fact that FEA can effectively simulate ultrasonic wave propagation processes. There is evidence that the method could potentially be extended to biomedical engineering applications, e.g., the identification of cancer tumors in soft breast tissue.

\subsection{ML for biomechanics}

Biomechanics is an important research field addressing the mechanics of biological systems, including organs and tissues \cite{RN166}. One important research topic in biomechanics is the understanding of human movement by analyzing motion data from sensors. Comprehensive and massive data have been collected over the past decades, including videos of human motion kinematics, force/displacement data from wearable devices like flexible electronics, and images obtained from computed tomography (CT) and magnetic resonance imaging (MRI). Understanding these data and developing applicable biomechanical models can guide the design of new devices and technologies to address body-related issues, e.g., predicting injury risk in sports and developing advanced medical devices. Recently, ML has been widely applied for data analysis from wearable sensors \cite{zhang2022deep}. For example, Komaris et al. \cite{RN168} successfully trained NNs based on a public dataset of 28 professional athletes to estimate the runner’s kinetics. Eerdekens et al. \cite{RN167} employed a CNN-based ML model trained from accelerometer data to understand equine activity. Recent reviews \cite{RN169, RN170, RN171, RN172} have extensively surveyed the application of ML in this area. Therefore, in this subsection, we focus on reviewing another important ML application in experimental biomechanics: characterizing and modeling biological materials such as tissues. Another important ML applications in biomechanics is cellular manipulation \cite{gerbin2021cell}, the readers are referred to a recent review paper \cite{RN89} for details.

\begin{figure}[!ht]
    \centering
    \includegraphics[width=1.0\textwidth]{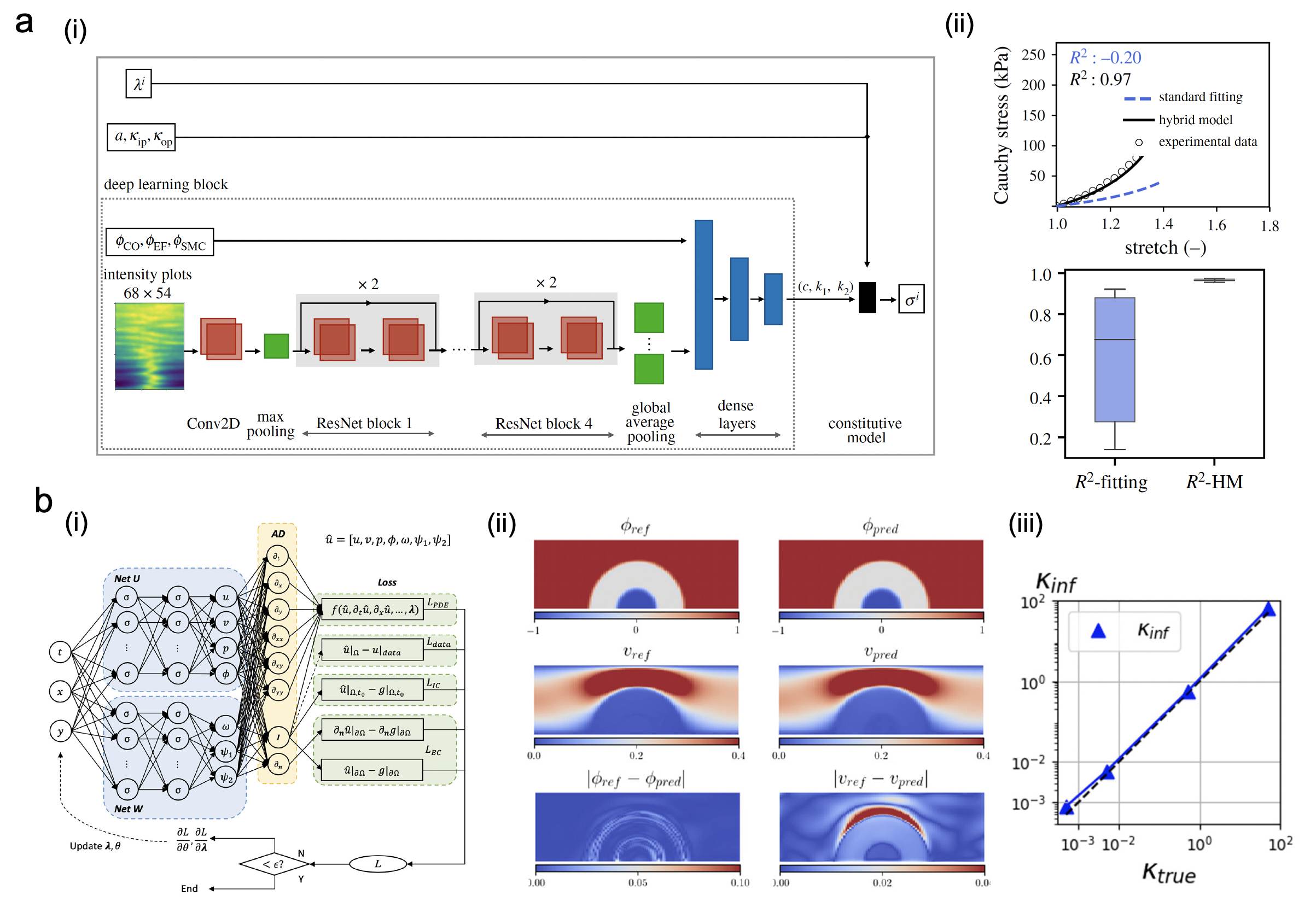}
    \caption{Applications of ML in constitutive parameter inversion for biomaterials. (a) A hybrid DL framework to identify unknown material parameters of arteries with high coefficient of determination: (i) Hybrid model architecture; (ii) Predicted stress-stretch curves from standard fitting method compared to the proposed hybrid model. (Reproduced with permission from Ref. \cite{RN181}. Copyright 2021, The authors, published by the Royal Society). (b) Non-invasive inference of thrombus material parameters using PINNs: (i) Schematic of PINNs for solving inverse problem; (ii) Prediction and ground-truth of 2D flow around a thrombus; (iii) Comparison of the inferred permeability of 2D flow with the ground-truth. (Reproduced with permission from Ref. \cite{RN151}.  Copyright 2021 by Elsevier).}
    \label{fig:abstract}
\end{figure}

Most biological tissues, such as blood vessels and brain matter, are soft materials. Their surface can easily form multi-mode instability, i.e., Ruga morphologies \cite{RN173, RN174, RN175, RN176}, under external loading. For example, Jin et al. \cite{RN173} performed FEM and experiments to systematically understand the surface instability and post-bifurcation phenomenon of soft matter containing orifices, e.g., arteries, when subjected to external pressure. Furthermore, many of these materials are anisotropic, i.e., their mechanical responses are dependent on the loading direction. Therefore, the characterization of these materials from well-planned experiments and identifying constitutive models are needed to understand and predict their mechanical behaviors. In turn, the information is used in the investigation of disease and the design of artificial organs. Many historical models have been developed to characterize various biomaterials, such as the neo-Hookean model \cite{RN177}, Ogden model \cite{RN178}, Fung-type model for blood vessels \cite{RN179}, and Holzapfel-Gasser-Ogden (HGO) model \cite{RN180} for anisotropic biomaterials. However, identifying these constitutive parameters for complex models from experimental data typically requires extensive nonlinear FEM and sophisticated optimization algorithms \cite{RN198, RN199, RN200}. ML can help infer material parameters from limited experimental data with multi-modality, such as mechanical testing and microstructure data obtained from images. For example, Liu et al. \cite{RN183} developed a ML framework to identify the HGO constitutive parameters of aortic walls based on synthetic microstructural data. Kakaletsis et al. \cite{RN201} compared the parameter identification accuracy from an iterative optimization framework and a stand-alone NN for isotropic and anisotropic biomaterials. Results suggest that replacing FEM with Gaussian process regression (GPR) or NN-based metamodels could accelerate the parameter prediction process. While replacing the entire optimization process with a stand-alone NN yielded unsatisfactory predictions. Recently, Holzapfel et al. \cite{RN181} developed a hybrid DL model based on a residual network (ResNet) and CNN to infer three unknown material parameters for a modified HGO model \cite{RN182} (\textbf{Fig. 4 (ai)}). This ML model maps second-harmonic generation (SHG) microstructure images, representing the orientation and dispersion of collagen fibers, to their mechanical stress-strain data for a total of 27 artery samples. By employing this hybrid ML model, the coefficient of determination, $R^{2}$, was 0.97, while conventional least square fitting gave $R^{2}$=0.676 with a much larger standard deviation (\textbf{Fig. 4 (aii)}). The exceptional high-accuracy prediction achieved from a limited biomechanics dataset can be attributed to two factors. First, the multi-modal nature of the dataset allows for the inference of material parameters beyond the conventional stress-strain data. Second, a priori knowledge of theoretical constitutive laws also contributes to reducing the necessary dataset size. As the authors suggested, expanding the experimental datasets, and incorporating biaxial extension experiments are required to validate these results. This research could have a transformative impact on soft tissue constitutive modeling, i.e., modeling soft tissues using prior physics laws and limited but multi-modal experimental datasets.

Recent advances in PINNs provide a promising alternative in constitutive parameter identification for biomaterials by encoding the underlying physics. As shown in \textbf{Fig. 4 (bi)}, Yin et al. \cite{RN151} employed PINNs to infer the permeability and viscoelastic modulus of the thrombus. The interaction between thrombus and blood flow can be described by sets of PDEs like Cahn–Hilliard and Navier-Stokes equations. The parameters can be accurately identified by encoding these governing physics during the PINNs training (\textbf{Fig. 4 (bii)\&(biii)}). Their results also demonstrated that PINNs could infer material properties from noisy data with complexity. Recently, Kamali et al. \cite{RN202} implemented PINNs to accurately identify Young’s modulus and Poisson’s ratio for heterogeneous materials like brain matter. This method has potential clinical applications such as noninvasive elastography.

Recently, a data-driven computation framework for constitutive modeling was proposed by Ortiz and coworkers \cite{RN204, RN205, RN206, RN207}. In this framework, a data-driven solver directly learns the mechanical responses of materials from experimental data, which eliminates the need for complex empirical constitutive modeling. In this approach, NNs can be employed to build surrogate models of biomaterials for constitutive modeling \cite{RN213, RN216}. For example, Linka et al. \cite{RN208} developed a constitutive artificial NN (CANN) to learn the constitutive models for hyperelastic materials directly from given stress-strain data. Masi et al. \cite{RN209} introduced a thermodynamics-based NN (TANNs) for constitutive modeling by coupling thermodynamics laws as constraints during training. Li and Chen \cite{RN210} developed an equilibrium-based CNN to extract local stress distribution based on strain measurement from DIC for hyperelastic materials. Wang et al. \cite{RN211} developed an ML algorithm based on singular value decomposition (SVD) and a Gaussian process to build metamodels of constitutive laws for time-dependent and nonlinear materials. This metamodeling method can be used to determine sets of material parameters that are best fit for experimental data. Liu et al. \cite{RN212} developed a physics-informed neural network material model (NNMat) to characterize soft biological tissues. Their model consists of a hierarchical learning strategy by first learning general characteristics for a class of materials and then determining parameters for each individual case.

Neural operators can also be employed to build data-driven surrogate models for constitutive modeling of biomaterials due to their advantage of generalizability and prediction efficiency to different inputs. For example, as shown in \textbf{Fig. 5 (ai)}, Zhang et al. \cite{RN164} developed a DeepONet-based model, genotype-to-biomechanical phenotype neural network (G2$\Phi$net), to characterize mechanical properties of soft tissues and classify their associated genotypes from sparse and noisy experimental data. With a 2-step training process consisting of a learning stage and an inference stage with an ensemble, G2$\Phi$net could effectively learn the constitutive models from biaxial testing data for 28 mice with 4 different genotypes with an L2 error of less than 5\% (\textbf{Fig. 5 (aii)}). Interestingly, it could also identify the correct genotype. This DL framework has important implications in biomechanics and related clinical applications, which is learning relationships between genotype and constitutive behaviors in biological materials from limited experimental data. Yin et al. \cite{RN165} employed DeepONet to build a data-driven surrogate model that could predict the damage progression of heterogeneous aortic walls. Goswami et al. \cite{RN220} developed a DeepONet-based surrogate model to identify pathological insults that could lead to thoracic aortic aneurysm (TAA) from a synthetic FEM database. More recently, You et al. \cite{RN218} employed a Fourier neural operator based method to model mechanical responses of soft tissues under different loading conditions directly from experimental data (\textbf{Fig. 5 (bi)}). The proposed physics-guided implicit Fourier neural operator architecture is shown in \textbf{Fig. 5 (bii)}. This method learned the material deformation model from DIC measurements and could predict the displacement field under unseen loading conditions with errors smaller than those ascertained in conventional constitutive models for soft tissues (\textbf{Fig. 5 (biii)}).

\begin{figure}[!ht]
    \centering
    \includegraphics[width=1.0\textwidth]{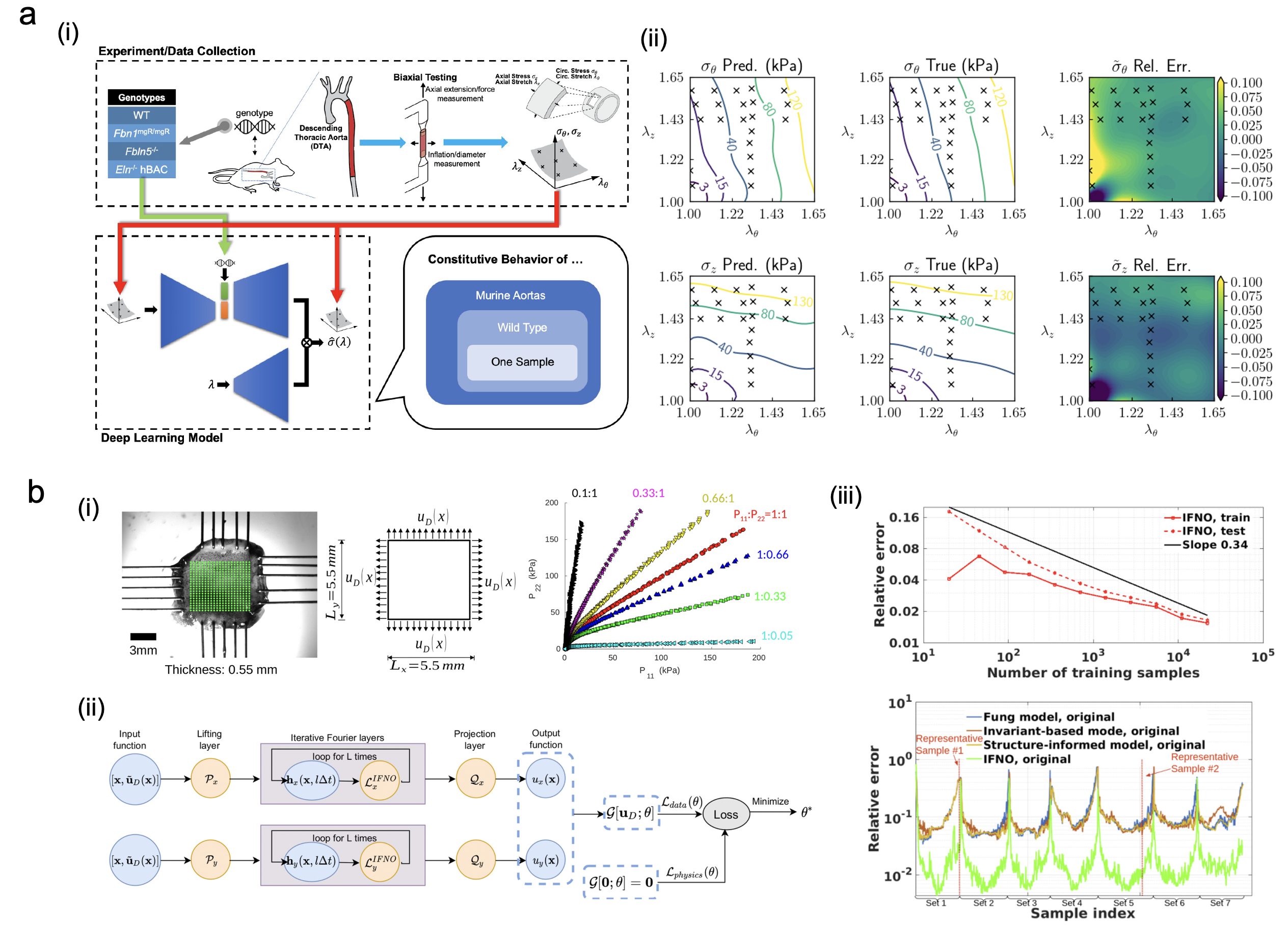}
    \caption{Applications of neural operator in constitutive modeling of biomaterials. (a) A deepONet-based DL framework to infer biomechanical response and associated genotype of tissues: (i) the DL framework; (ii) reconstructed stress-stretch relationships compared with their true values (Reproduced with permission from Ref. \cite{RN164}. Copyright 2022, The authors, published by PLOS). (b) A neural operator model to construct the mechanical response of biological tissues from displacement data measured from DIC: (i) The biaxial experimental setup; (ii) The architecture of the physics-guided Fourier neural operator; (iii) Error comparisons of proposed Fourier neural operator method and other mechanics models (Reproduced with permission from Ref. \cite{RN218}. Copyright 2022 by ASME).}
    \label{fig:abstract}
\end{figure}

\subsection{Micro and Nano-mechanics}

The emerging development of nanotechnology and biotechnology in recent several decades continuously requires a new understanding of micro- and nano-scale material behaviors. For example, much research has been conducted on understanding the mechanical properties of thin films \cite{RN52, RN53, RN251, RN247, RN252, RN255, RN254, RN256, RN257}, nanopillars\cite{greer2006nanoscale,chen2023exploring}, nanostructured metals \cite{RN59, RN60, RN223, RN258, RN259,jin2017distribution}, sub-micron-sized sensors \cite{RN55}, crystalline nanowires \cite{RN54, RN260, RN261, RN262, RN263, RN264}, 2D materials \cite{RN56, RN266, RN267, RN268, yang2021intrinsic}, origami \cite{RN265}, nanolattice metamaterials \cite{RN57, RN270, RN271}, and copolymers with nanoscale features \cite{RN11, RN58, RN221, RN222}. Conducting precise experiments to characterize these materials with small-scale features is essential to understand their properties, underlining mechanisms, and develop constitutive models. There are two major steps in this process. First, experimental measurements are taken from instrumentations with nanometer resolution. Advanced microscopes, including Scanning Electron Microscope (SEM), Transmission Electron Microscope (TEM), and Atomic Force Microscope (AFM), have been developed to capture images with nanometer resolution. Furthermore, nano-mechanical instruments like nanoindenters and MEMS, have been developed for nanomechanical characterization. Next, the properties of interest can be extracted from measurements via inverse algorithms. However, compared to macro-scale samples, the connection between measurable data and material properties is not straightforward for samples with nanoscale features due to local inhomogeneity \cite{RN61, RN79}, size effect \cite{RN245, nix1998indentation}, or chemo-mechanical coupling \cite{papakyriakou2022nanoindentation}. For example, in instrumented indentation \cite{{oliver1992improved, oliver2004measurement}}, identifying the material properties from measured load and indentation depth data is nontrivial and sometimes may not guarantee unique solutions when material constitutive law is elastoplastic \cite{alkorta2005absence, chen2007uniqueness} or the indentation tip is conical shape \cite{cheng1999can}. Therefore, there is a critical demand to identify material properties and quantify their uncertainty using nanomechanical experiments. Due to the popularity and accessibility of nanoindentation data, ML has been widely applied to such data. In this subsection, we will mainly focus on reviewing ML applications for material properties extraction from nanoindentation, and in particular, we will review two methods: the neural network (NN) approach and the Bayesian-based statistical approach. We note that the methods are also applicable to other nanomechanical testing, such as the membrane deflection experiments, which provide direct measurement of stresses and strains \cite{RN247}, and in-situ microscopy testing using MEMS technology \cite{RN248}. Lastly, ML applications in other nano and micro-mechanics, like microstructure characterization, will be briefly reviewed.

\begin{figure}[!ht]
    \centering
    \includegraphics[width=1.0\textwidth]{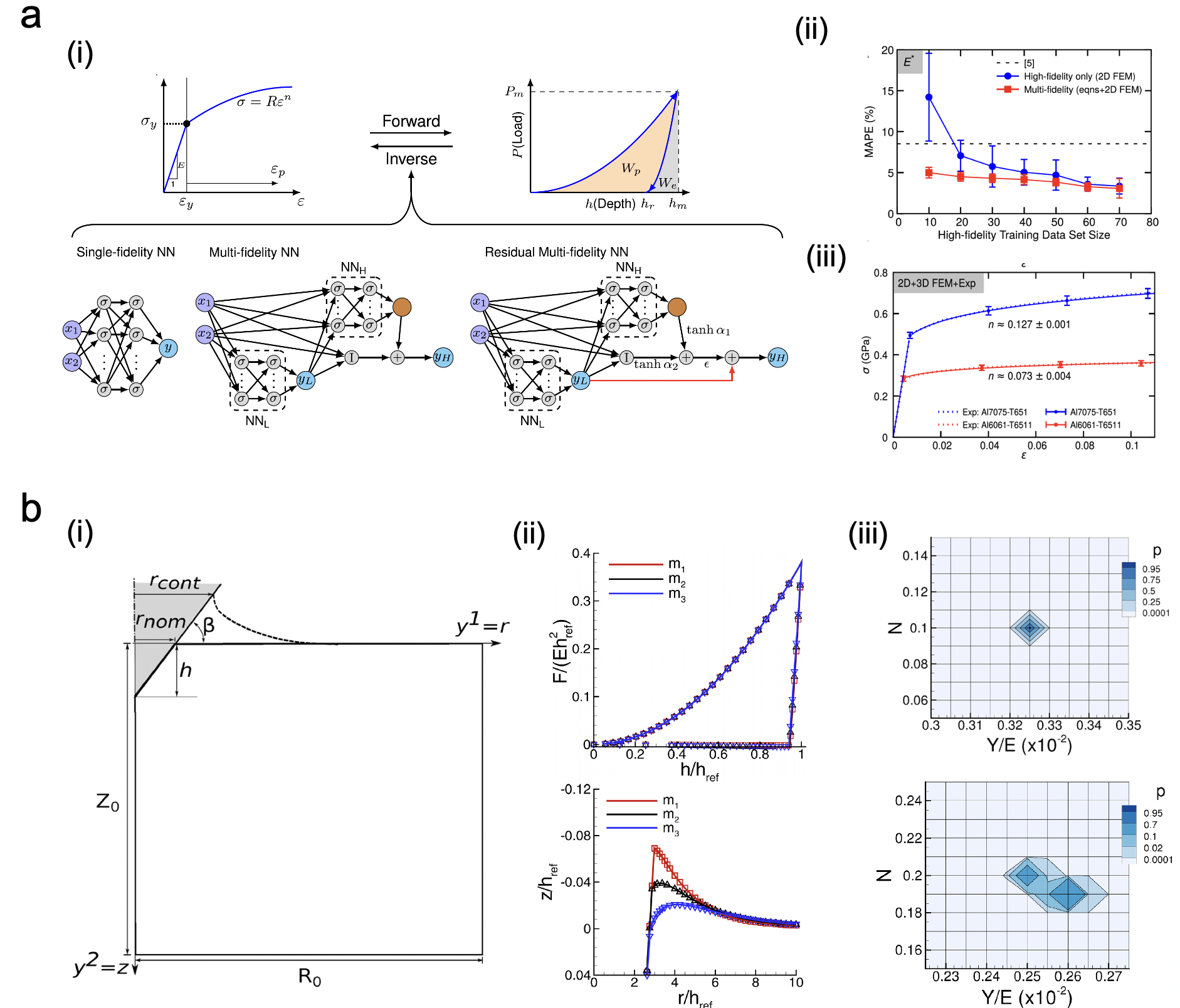}
    \caption{Applications of ML in nanoindentation. (a) DL methods including single-fidelity NNs, multi-fidelity NNs, and residual multi-fidelity NNs to identify material parameters from instrumented indentation: (i) architectures of these NNs; (ii) MAPE as a function of training dataset size for multi-fidelity NNs; (iii) identification of hardening exponent for two aluminum alloys from 2D, 3D FEA simulations and three experimental data points (Reproduced with permission from Ref. \cite{RN77}. Copyright 2020, The authors, published by PNAS). (b) A Bayesian-type statistical approach to identify material parameters and their uncertainty from conical indentation: (i) schematics of indentation configuration with conical tips; (ii) force-indentation depth curves and surface profiles for three different materials; (iii) posterior probability distribution of parameters with and without noisy data (Reproduced with permission from Ref. \cite{RN85}. Copyright 2019 by ASME).}
    \label{fig:abstract}
\end{figure}

For using the NNs method to identify material properties in nanoindentation, most of the training is based on FEM due to its flexibility and efficiency. Indentation, a well-defined contact mechanics problem, can be accurately simulated in either commercial FEM software or in-house codes \cite{needleman2015indentation, bower1993indentation}. In 2002, Muliana et al. \cite{muliana2002artificial} trained a neural network with hidden layers based on 2D and 3D FEM simulations to map nonlinear material properties from simulated load-displacement curves. They found that the trained NN can accurately predict the load-displacement curves of materials with properties not included in the training dataset. This work demonstrated the potential of using NNs to inversely obtain unknown material properties from experimental data. Huber et al. \cite{RN62, RN63} used FEM-trained NNs to identify the Poisson’s ratio of materials exhibiting plasticity with isotropic hardening, something not easily obtained before. Since then, FEM-trained NNs have been widely applied as an inverse algorithm to identify material properties from nanoindentation \cite{RN67, RN68, RN69, RN70, RN71, RN72, RN73, RN74, RN75, RN76}. However, in practice, these FEM-trained NNs could be cumbersome since they require a substantial amount of FEM training data to survey combinations of material properties within specific ranges. Such a training process is generally computationally expansive, especially when the unknown parameter space is large. Furthermore, the trained NNs usually have poor extrapolation performance, i.e., when material properties are outside the range used in the training dataset. Another issue in this approach is the lack of uncertainty quantification when identifying properties from experimental datasets based on FEM. To overcome these challenges, Lu et al. \cite{RN77} utilized a multi-fidelity NN (MFNN) \cite{RN78}, which trained low-fidelity FEM datasets together with a few high-fidelity experimental datasets together (\textbf{Fig. 6 (ai)}). As shown in \textbf{Fig. 6 (aii)\&(aiii)}, the MFNN can efficiently learn the correlations between these two datasets with different fidelities, hence significantly increasing the identification performance while reducing the size of FEM training datasets. The MFNN performance can be further improved by employing transfer learning when additional experimental data are obtained.

The NN method can identify unknown material parameters from indentation data via cost function minimization. However, this method could not systematically quantify the uncertainty of identified parameters, i.e., yielding the likelihood that other possible parameter sets also minimize the cost function. The Bayesian method, which was derived based on Bayes’ theorem \cite{RN80}, can be employed to quantify uncertainty for nanoindentation since it can provide a posterior probability for each set of parameters. For example, Fernandez-Zelaia et al. \cite{RN81} utilized the Bayesian framework to identify unknown material parameters from a FEM-trained Gaussian process surrogate model for spherical indentation experiments. The Bayesian framework has also been employed to identify unknown properties from spherical indentation for single crystal \cite{RN83}, plastic solids with exponential hardening laws \cite{RN82}, plastically compressible solids with Deshpande–Fleck constitutive laws \cite{RN84}. For the case of conical indentation, which is similar to experiments using a Berkovich indenter, the indentation force versus indentation depth (P-h) data sometimes could not yield unique sets of unknown parameters. To solve this issue, as shown in \textbf{Fig. 6 (bi)}, Zhang et al. \cite{RN85, RN86} employed a Bayesian framework to extract plastic properties and quantify the uncertainty from both P-h curves and surface profile datasets obtained from FE simulations (\textbf{Fig. 6 (bii)}). Furthermore, as shown in \textbf{Fig. 6 (biii)}, the posterior probability for possible material parameters could be calculated from both noise-free and noise-contaminated datasets. Later, Zhang and Needleman \cite{RN87} applied the Bayesian framework to infer the power-law creep constitutive parameters considering both the time-dependent indentation depth data and the residual surface profile. Though this framework was developed based on synthetic data from simulations, the parametric identification Bayesian approach offers valuable insights into uncertainty quantification resulting from nanoindentation experiments performed on materials with complex constitutive behaviors.

To understand the mechanical properties of soft materials like biological tissues \cite{RN90} and cells \cite{RN89}, one can conduct nanoindentation tests using AFM signatures based on high-resolution force detection and cantilever tip position obtained from a four quadrant position-sensitive photodiode (PSPD) \cite{RN249, RN250}. For example, Rajabifar et al. \cite{RN93} trained a multilayer NN to predict surface viscoelastic and adhesive properties of samples based on load-displacement curves obtained from AFM tapping mode. They generated the training data from a rigorous contact mechanics model, known as the enhanced Attard’s model. Here also, ML can be employed to extract the nanoscale force without complex modeling. For example, Chandrashekar et al. \cite{RN91} used a data-driven ML algorithm to capture the tip-sample force in dynamic AFM. The algorithm was also used to successfully identify the interaction forces for two-component polymer blends. Furthermore, the material parameter identification from AFM indentation data inherently involves uncertainty, which depends on the choice of contact mechanics model. To mitigate this uncertainty in modulus identification, Nguyen and Liu \cite{RN92} employed five conventional supervised ML techniques (decision tress, K-nearest neighbors, linear discriminant analysis, Naïve Bayes, and support vector machines) to classify AFM indentation curves for different materials into appropriate contact mechanics models. By choosing the appropriate contact mechanics model, the uncertainty of modulus identification can be potentially reduced.

Another important ML application for nano and micro-mechanics is microstructure characterizations \cite{RN225, RN226, RN230, RN235}. Since the rapid advances of high-resolution imaging techniques, ML algorithms like CNNs have been increasingly applied to microstructure characterizations. For example, ML can be employed in microstructural image segmentation to identify individual grains or phases \cite{RN228,chen2018deformation,durmaz2021deep,stuckner2022microstructure}. Additionally, ML can be applied to detect defects such as cracks based on microstructural images \cite{RN231} as well as predict unknown material properties like yield strength \cite{RN229} based on microstructure features such as grain size, orientation, and crystallographic features. Furthermore, ML can be applied to analyze evolving microstructure images to analyze time-dependent material behaviors during fracture \cite{RN232, RN236}, recrystallization \cite{RN233}, phase transformations \cite{RN234}, and cell morphology \cite{RN245, RN191, RN89, RN246}. These ML applications in microstructure characterizations could enable rapid and accurate analysis of complex material microstructures, leading to the development of next-generation materials with desired properties.

\subsection{ML for architected materials}

Recently, the concept of materials by design has enabled us to design multi-functional materials with unprecedented properties. Such progress can be attributed to the rapid development of computational tools for structural design and experimental techniques for the synthesis and characterization of materials \cite{RN30, RN29}. Among these novel materials, architected metamaterials, which combine the properties of material constitutes and architectural design, have demonstrated their superior mechanical properties like ultra-high specific strength \cite{RN34}, excellent recoverability \cite{RN271, RN33}, and impact resilience \cite{RN32}. Along this line, ML is becoming an important tool to systematically design these novel architected metamaterials with desired properties and functionality beyond laboratory trial-and-error \cite{RN123, RN41,RN43, RN44, RN45, RN47}. Comprehensive review papers have been published in recent years that reviewed and discussed the methodology and applications of ML in architected material design \cite{RN35, RN51, RN37, RN50}. Herein, in this subsection, we will focus on reviewing recent advances in experimental efforts in ML-enabled design of architected materials.

\begin{figure}[!ht]
    \centering
    \includegraphics[width=0.9\textwidth]{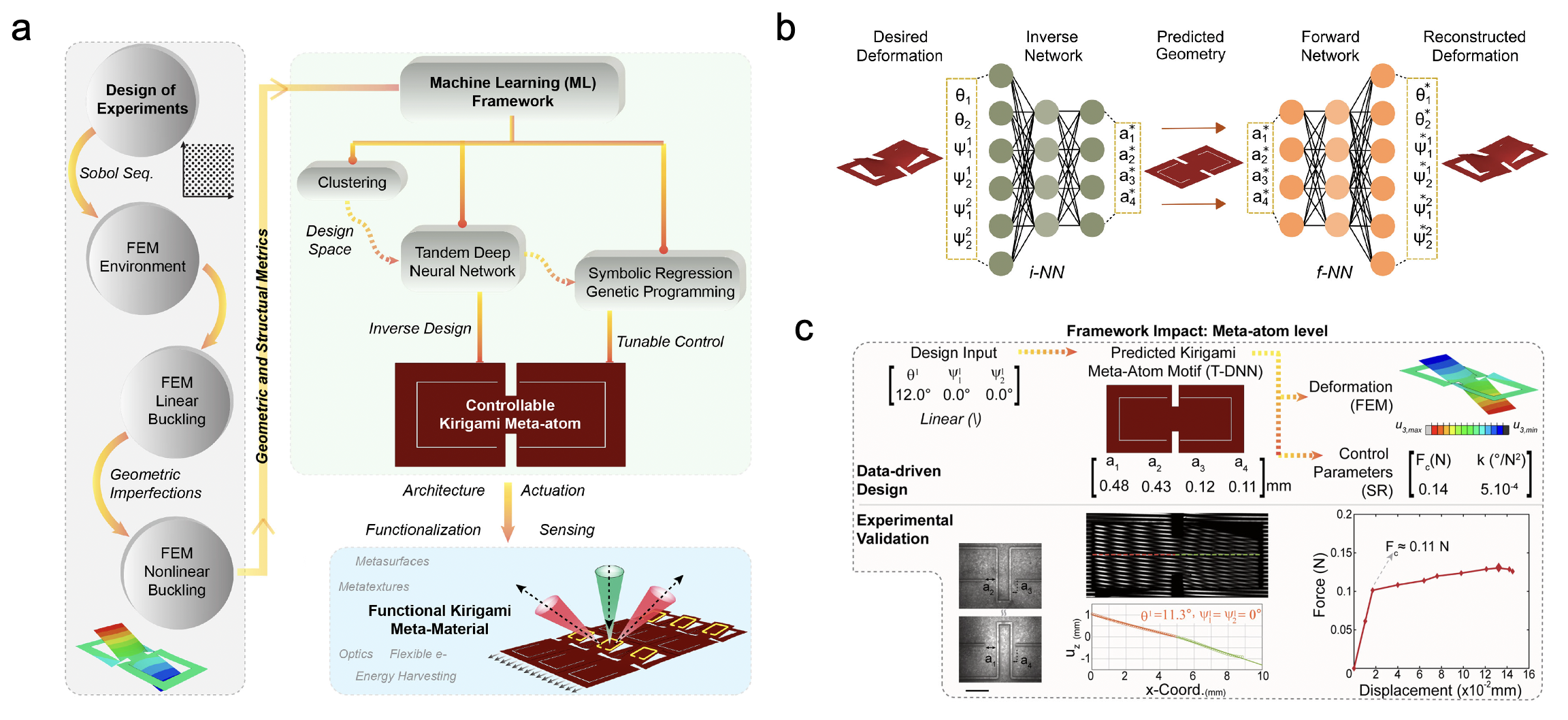}
    \caption{Applications of ML in designing shape-programable kirigami metamaterials. (a) The ML framework to inverse design kirigami metamaterials. (b) Schematics of the tandem network employed for inverse design. (c) Experimental verification of inverse design from shadow Moiré method (Reproduced with permission from Ref. \cite{RN38}. Copyright 2022, The authors, published by Springer Nature).}
    
    \label{fig:abstract}
\end{figure}

One significant aspect of experimental mechanics in architected materials is to verify the predicted power of computational methods used in the inverse design. Most of the current ML frameworks in material discovery were trained based on big datasets from reliable physically-based computer simulations, given the formidable challenges associated with cost and speed for acquisition of large-scale experimental data sets. Therefore, careful verifications from experiments are necessary to assess the ML framework’s performance in producing real structures/materials with desired properties. As shown in \textbf{Fig. 7 (a)}, Alderete and Pathak et al. \cite{RN38} have proposed an ML framework that combines the K-mean clustering methods for design space reduction and a tandem NN architecture to inversely design shape-programmable 3D Kirigami metamaterials. The tandem NN architecture was employed to circumvent the non-uniqueness issue during the inverse design process (\textbf{Fig. 7 (b)}). The framework was trained on finite element predictions of instabilities triggering 3D out-of-plane shapes validated by full-field experimental measurements using the shadow Moiré method \cite{RN39} (\textbf{Fig. 7 (c)}). A very good agreement between ML-designed cuts and predicted out-of-plane deformation and experiments was found under mechanical actuation (stretching). Moreover, using symbolic regression, the authors could predict the onset of actuation and needed stretching to achieve specific 3D shapes. These programmable 3D Kirigami metamaterials can be used in various engineering applications over a large range of size scales, from microscale particle trapping to macroscale solar tracking.

\begin{figure}[!ht]
    \centering
    \includegraphics[width=0.9\textwidth]{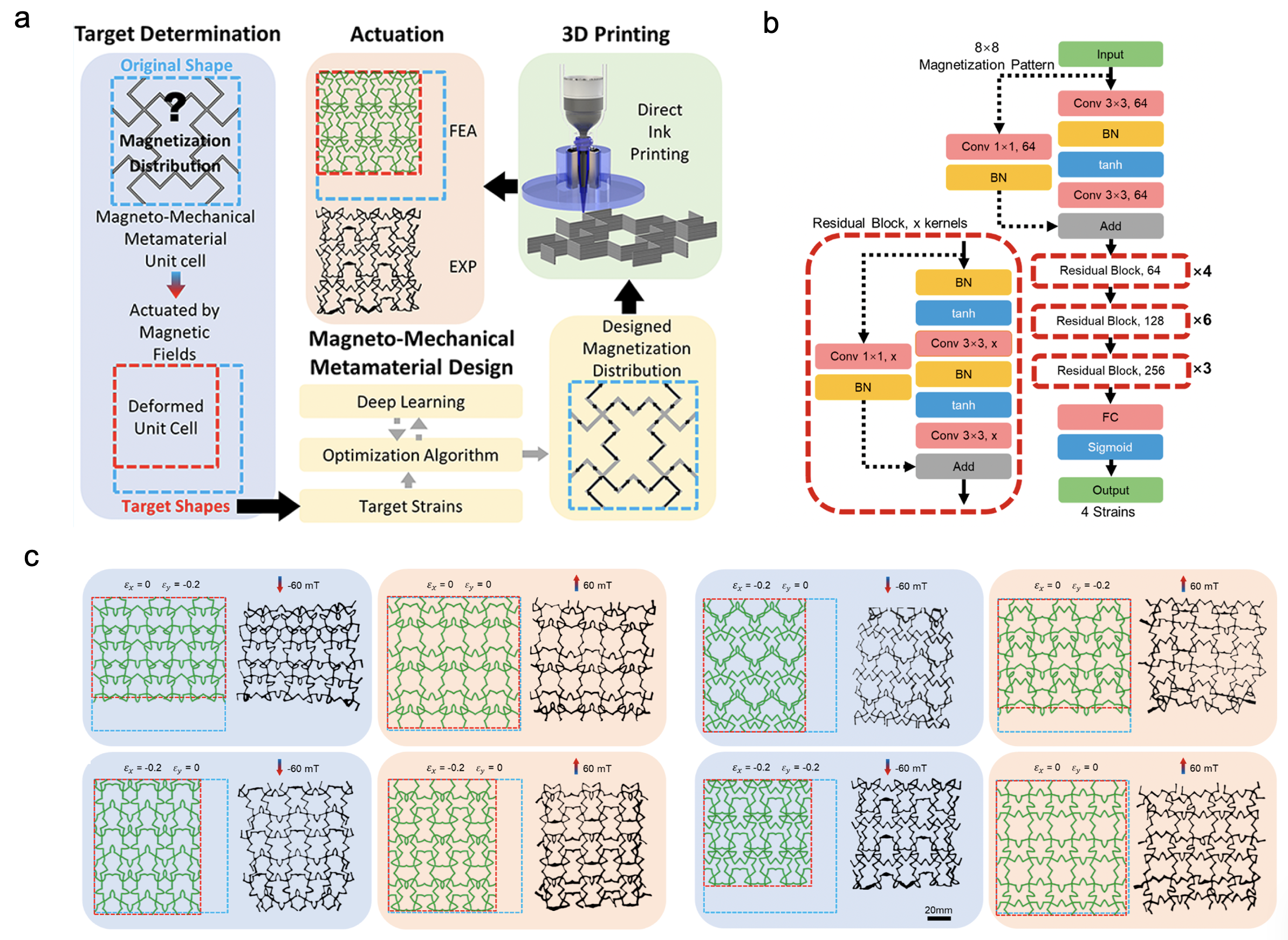}
    \caption{(a) Applications of ML in the inverse design of magneto-activate mechanical metamaterials. (b) The deep residual network (ResNet) architecture. (c) Comparisons between FEA and experiments (Reproduced with permission from Ref. \cite{RN40}. Copyright 2022 by ACS).}
    
    \label{fig:abstract}
\end{figure}

\begin{figure}[!ht]
    \centering
    \includegraphics[width=0.9\textwidth]{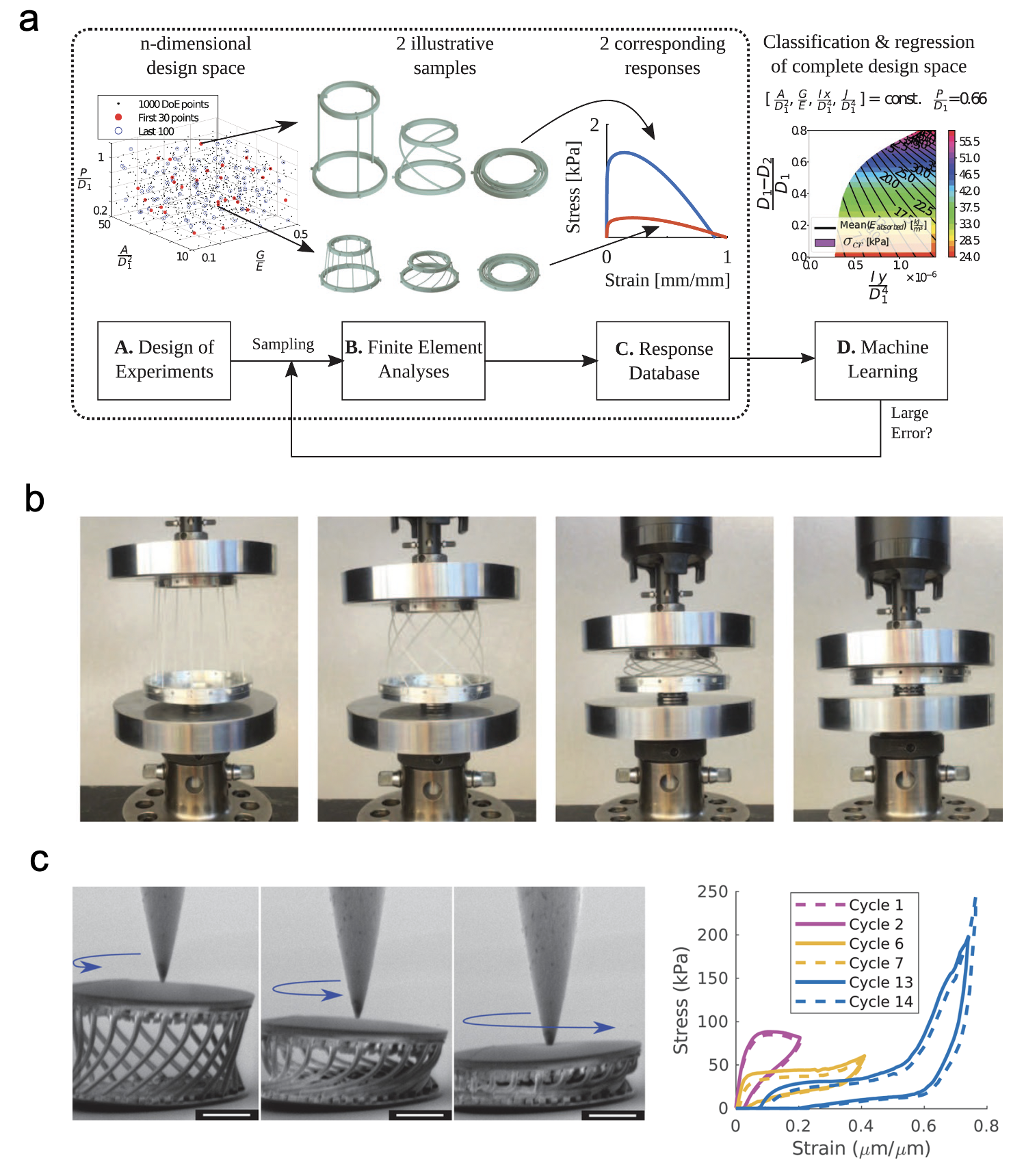}
    \caption{(a) Data-driven Bayesian ML framework for supercompressible metamaterials design. (b) Experimental validation for designed structure from fused filament fabrication using polylactic acid. (c) Experimental validation for designed structure with micro-scale size from two-photon lithography. The scale bar in (c) is 50 µm. (Reproduced with permission from Ref. \cite{RN48}. Copyright 2019, The Authors, published by WILEY).}
    
    \label{fig:abstract}
\end{figure}

Likewise, ML can also assist in the inverse design of complex soft materials. For example, as shown in \textbf{Fig. 8 (a)}, Ma et al. \cite{RN40} developed a deep residual network (ResNet) trained with FEM to inversely design tunable magneto-mechanical metamaterials. The ResNet model was chosen due to its capability to preserve information between shallow and deeper layers (\textbf{Fig. 8 (b)}). The predicted structures were printed and tested in good agreement with ML predictions (\textbf{Fig. 8 (c)}). Moreover, GANs could also be employed to design architected structures without prior experience. For example, Mao et al. \cite{RN46} used GANs to systematically design complex architected materials and found that some structures can reach Hashin-Shtrikman (HS) upper bounds. Experimental verification was also conducted by the authors to show the robustness of the framework. Furthermore, due to the stochastic deformation modes of metamaterials like buckling under compression, uncertainty quantification is necessary to design reliable structures. For example, Bessa et al. \cite{RN48} demonstrated that a data-driven Bayesian ML framework could enable the design of super compressible metamaterials (\textbf{Fig. 9 (a)}). Experiments on multiscale 3D printed structures have also demonstrated the super compressibility predicted by simulations (\textbf{Fig. 9 (b)\&(c)}).

Another important feature in this field is to ability to conduct high-throughput experiments to generate high-fidelity data needed for ML training. Until now, most of the ML applications of the inverse design of architected metamaterials were based on training from computer simulations. However, such a computer data-driven approach assumes the model is accurate and readily obtainable. This may not be the case in more complex behaviors arising from material nonlinearities and rate dependences for which accurate constitutive descriptions do not exist. A solution would be to run autonomous experiments, based on a large number of additively manufactured structures/samples, to obtain valuable representative data such as stress-strain curves or deformation patterns \cite{gongora2020bayesian, gongora2021using, stach2021autonomous}. As such, there is a need to develop new ML frameworks to inverse design new metamaterials directly from limited and noisy experimental data. For example, Lew and Buehler \cite{RN49} trained an ML framework called DeepBuckle that combines the Variational Autoencoder (VAE) model and Long Short-Term Memory (LSTM) model to quantitatively learn buckling behaviors of polymer beams from simple and limited mechanical testing on 3D printed structures.

\subsection{ML for 2D materials fracture toughness characterization}

In the past decades, significant progress has been made in the synthesis of 2D materials such as graphene, hexagonal Boron Nitride (h-BN), and transition metal dichalcogenides (TMDs), e.g., molybdenum disulfide (MoS2), including fabrication, chemical functionalization, transfer, and device assembly \cite{akinwande2019graphene, geim2009graphene, dong2022facile}. Therefore, quantifying their mechanical properties, like fracture toughness, is crucial to ensure the durability and reliability of these 2D material devices. Ni et al. thoroughly reviewed the recent experimental, theoretical, and computational progress on quantifying 2D materials’ fracture properties \cite{ni2022fracture}. In addition, the progress on ML toward 2D material fracture, including fracture pattern characterization, has also been discussed. Here, in this subsection, we will focus on reviewing the most recent experimental mechanics effort from the authors’ group to quantify the fracture toughness from experiments with atomistic resolution. Meanwhile, we will review ML-based parametrization of interatomic potentials for 2D materials and discuss our integrated experimental-computational framework not only to understand 2D material fracture (\textbf{Fig. 10 (a)}) but also as a case study for the use of ML and atomistic experimentation in advancing the predictive power of atomistic models employed in the design space exploration of family of 2D materials (TMDs, MXenes) in the spirit of the materials genome initiative.

\begin{figure}[!ht]
    \centering
    \includegraphics[width=1.0\textwidth]{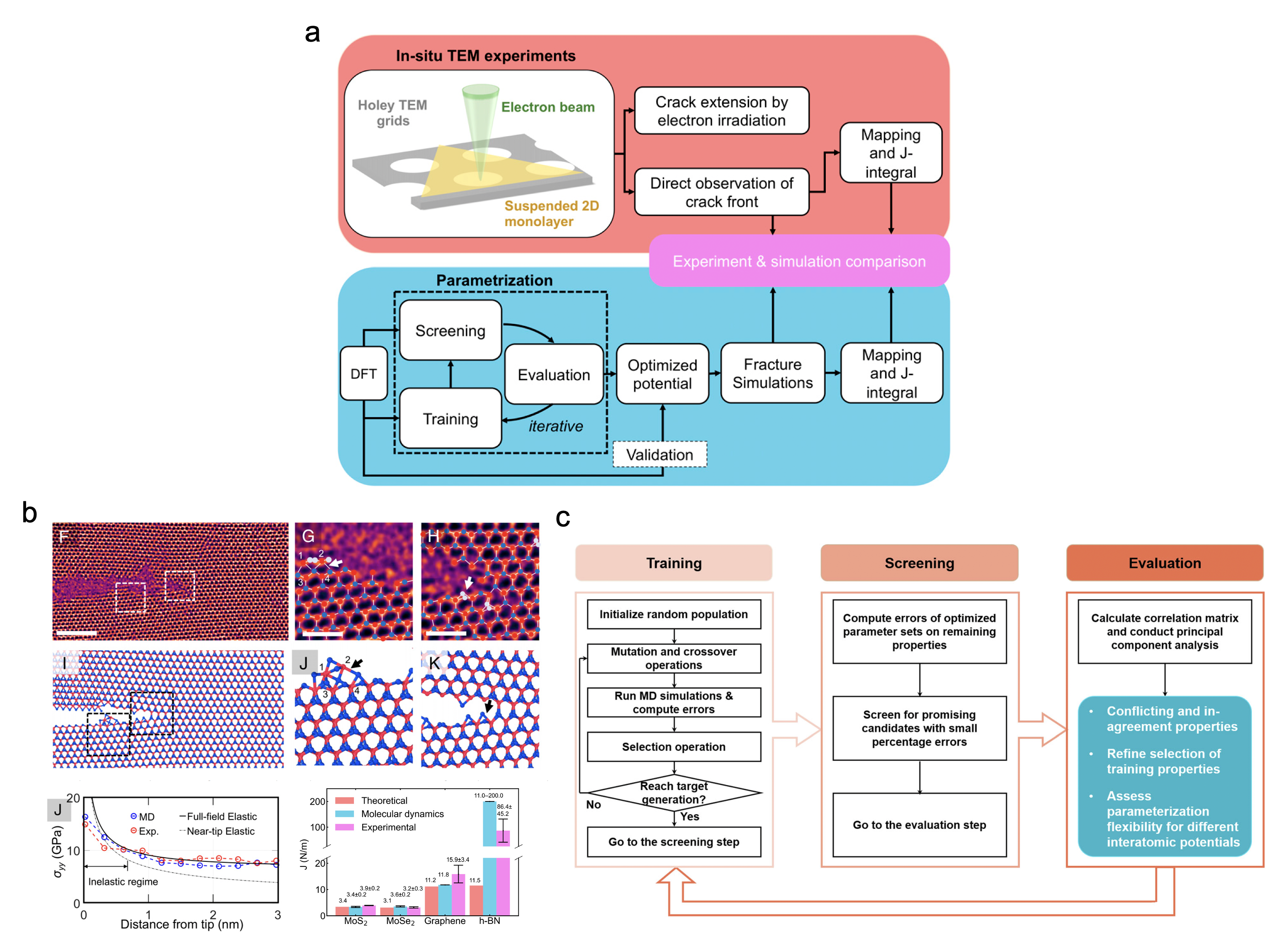}
    \caption{Applications of ML in fracture toughness characterization of 2D materials. (a) An integrated experiment-simulation framework to quantitively measure intrinsic fracture toughness of 2D materials. (b) In-situ HRTEM experiments of MoSe2 compared with MD simulations (Reproduced with permission from Ref. \cite{zhang2022atomistic}. Copyright 2022 by National Academy of Sciences). (c) Schematic of ML-based interatomic potential parametrization approach consisting of three steps: training, screening, and evaluation (Reproduced with permission from Ref. \cite{zhang2021multi}. Copyright 2021, The authors, published by Springer Nature).}
    \label{fig:abstract}
\end{figure}

The fracture toughness of monolayer graphene \cite{zhang2014fracture} and h-BN \cite{yang2021intrinsic} have been measured from in-situ SEM and TEM experiments. It was found that both materials obey Griffith’s brittle fracture criterion \cite{RN7}. Assisted by molecular dynamics simulations, researchers attributed their high toughness to different intrinsic toughening mechanisms like pre-existing grain boundaries for graphene and structural-asymmetry induced crack branching and deflection for h-BN. However, the direct observations of atomistic information like internal defects and lattice deformation are absent due to the resolution limits of the previous experiments. To overcome these challenges, in our group, Zhang et al. \cite{zhang2022atomistic} performed in-situ high-resolution transmission electron microscopy (HRTEM) fracture experiments to investigate the fracture toughness of two TMDs, MoS2 and MoSe2 (\textbf{Fig. 10 (b)}). The J-integral was computed from experimental stress-strain fields obtained from an affine transformation (deformation gradient) using high-resolution transmission electron microscope (HRTEM) images of atomic structures surrounding the crack tip. The experimental measurements revealed a nonlinear region near the crack tip, where bond dissociation occurs, and confirmed brittle fracture and the applicability of Griffith’s fracture criterion.

To verify the experimental observations, Zhang et al. also performed MD simulations with a newly developed ML-based parametrization framework \cite{zhang2021multi}. Force-field accuracy in MD simulations plays a crucial role in studying large atomic deformation as they occur near the crack tip. Although several successful parametrizations have been developed for various 2D materials \cite{jiang2013molecular, ostadhossein2017reaxff, wen2017force}, a force field that can accurately predict phase transition and fracture toughness of 2D materials was not reported. While ML has immense potential for force-field parameterization directly from large datasets based on first principal calculation like density functional theory (DFT) \cite{botu2017machine, li2017machine}, a suitable methodology that can incorporate atomic configurations far from equilibrium, essential to the prediction of fracture, is not common. Zhang et al. \cite{zhang2021multi} proposed a parametrization framework trained from DFT datasets and an evolutionary multi-objective optimization algorithm (\textbf{Fig. 10 (c)}). This parametrization was performed iteratively and consisted of three essential steps: training, screening, and evaluation. The force field was trained to capture both near-equilibrium properties like cohesive energy and nonequilibrium properties such as bond dissociation energy landscapes and vacancy formation energies. Therefore, the parametrized potential can accurately capture the bond breakage during fracture. MD simulations with this potential gave similar fracture toughness as those obtained from experimental measurements. We anticipate that by integrating in-situ experiments with atomistic resolution and MD simulations with force-field parametrized based on physical ML training, the fracture behaviors of other 2D materials beyond the TMDs family as well as other functional crystals can be identified. Moreover, MD simulations with ML parameterized force fields should accurately predict the effect of defects such as random vacancies, line vacancies, and grain boundaries, enabling the exploration of other constituents as well as thermos-mechanical properties.

\section{Discussions}

\subsection{Selection of appropriate ML model for a specific experimental mechanics application}

Selecting an appropriate ML model for a given task in experimental mechanics can be challenging since the broad range of ML algorithms and the complexity of experimental data. Here, we provide some practical guidelines on ML model selection based on the type and amount of available data, as well as the underlying physics of the problem. Before searching for a specific ML model, researchers and practitioners should ask themselves a crucial question: is ML necessary for the problem at hand? It is important to avoid the unnecessary use of ML when conventional methods and techniques are sufficient. For instance, using ML for constitutive parameter fitting for simple models like neo-Hookean solids from stress-strain data is not necessary. In such a case, nonlinear regression is preferred instead.

Next, it is important to properly define the types of problems that the user wishes to tackle. Are we using ML for an inverse problem like predicting unknown material parameters from experimental data? Or is it an optimization problem like the optimal design of architected materials subject to specific constraints? Having these questions and others answered before searching for suitable ML algorithms is important. Furthermore, one should be aware of the quantity and quality of available information before choosing an appropriate ML model. What forms of experimental data are available (e.g., stress-strain curves, images, full-field displacement)? What is the precision of these data? Is there any prior knowledge or simulation tool that may enrich the data? When a large amount of experimental data can be collected, a complex ML model like NNs can be applied. On the other hand, when there is data scarcity due to experimental constraints, the user may opt for employing PINNs or Gaussian Processes to embed additional physics and quantify uncertainty. 

When experimental data is not adequate, possible data augmentation techniques can be employed. One common approach is to interpolate among the original datasets. For image-based data, methods such as rotation, flipping or noise injection can be employed \cite{shorten2019survey}. It is crucial to exercise caution when applying these data augmentation methods, ensuring the key characteristics of the original data are maintained while increasing its quantity. Furthermore, if experiments are well-defined and can be simulated efficiently using some computational tools like FEA, MD, and DFT, we can generate synthetic data from these simulations. However, it is crucial to thoroughly examine and understand the differences and biases present in both computational and experimental datasets. We should ensure that the computational data accurately captures the experimental data's key features. During training, it is important to monitor the model's performance on both datasets during training to identify any overfitting to the synthetic data. Understanding differences between the fidelity of computational and experimental data is another important aspect, which will be discussed in the subsequent subsection.

The selection of an appropriate ML algorithm is also based on the experimental data type. If the data consists of images or video-based datasets, CNNs could be the most efficient. If the problem involves time-dependent data, RNNs and transformers could be a better choice. When the problem can be framed in terms of physical laws, PINNs should be employed. Given the speed at which the field is evolving, literature searches should be performed to identify the most suitable ML models for the problem at hand. Once the ML algorithms have been selected, the user can start to build an ML pipeline using open-source ML platforms such as TensorFlow \cite{RN237}, PyTorch \cite{RN239}, or JAX \cite{RN280}. Iterative refinement of the ML model by hyperparameter-tuning based on testing and validation datasets is necessary to reach the desired accuracy. However, it is worth noting that over-tuning hyperparameters can potentially result in overfitting. To address this issue, we can employ robust hyperparameter optimization methods such as k-fold cross-validation \cite{kohavi1995study}. In k-fold cross-validation, the original dataset is split into k equal folds. During each iteration, one-fold is used as the validation set while the other k-1 folds are used for training. After all iterations are completed, we can compute the average performance metric, such as mean squared error, from all k validation sets. This method can efficiently reduce the risk of overfitting since it trains and evaluates the model on different subsets of the data. Furthermore, it is important to keep in mind that there may be more than one possible choice of ML model for a specific experimental mechanics problem. Therefore, it is practical to carefully evaluate the prediction accuracy and efficiency of each model and select the model that is the best fit for a specific application.

\subsection{Integrating multi-modality and multi-fidelity experimental data into ML methods}

Conducting mechanical experiments, particularly those utilizing cutting-edge facilities and techniques for extremely large/small time and length scales, can be both costly and time-consuming. In many cases, researchers need to combine different experimental methods, hence gaining a better understanding of the mechanics problem. Furthermore, computer simulations may be employed to provide additional insights for the experiments. As a result, one may obtain experimental data with multi-modality and/or multi-fidelity. Multi-modal data refers to the data on an object comprising different forms and patterns, hence providing information from different channels (e.g., language data in the forms of text and speech, data on the mechanical test sample in the forms of images, and stress-strain curves). Multi-fidelity data refers to the measurement data with different levels of accuracy (e.g., high/low-resolution images of a test sample; stress-strain curves measured with load cells of different accuracies; data from real experiments and from computer simulations). Typically, high-fidelity data are expensive and hence limited, while low-fidelity data are cheap and plentiful.

To maximize the available information injected into learning algorithms, it is important to propose ML models that are capable of handling data with multi-modality \cite{RN281} and multi-fidelity \cite{RN78}. There have been some studies in applying such ML models to mechanics problems. For example,  Holzapfel et al. \cite{RN181} seek to develop an ML method that combines microstructural information and biomechanical tests. Trask et al. \cite{RN282} propose a framework that is capable of conducting multimodal inference for lattice metamaterials, relating their lattice design, stress-strain curves, and microstructural images. Lu et al. \cite{RN77} designed a multi-fidelity neural network for characterizing the mechanical properties of materials in instrumented indentation. While these studies, among others, have explored multi-modality and multi-fidelity ML methods for solid mechanics problems, further investigation is still needed to better integrate data with multi-modality and multi-fidelity from experiments and/or simulations to provide deeper insights for understanding the mechanics of materials and structures. 

\subsection{Estimating and reducing the uncertainty of ML predictions}

Most ML applications in experimental solid mechanics provide a point estimation – that is, a single value as the best estimate. To further acquire information regarding the reliability and confidence of such an estimation, one sometimes needs to quantify and/or reduce the uncertainty of the ML predictions. In the context of experimental solid mechanics, there are diverse sources of uncertainties coming from data and models, related to almost every component in the research workflow, including: (1) experimental implementation, such as the uncertainty of material properties caused by the manufacturing of specimens, inaccurate enforcement of the experimental setup (e.g., boundaries that are not perfectly clamped, approximate fulfillment of plane strain/stress condition), representativeness and noisiness of data; (2) theoretical modelling, such as the misspecification/oversimplification of constitutive models, ignoring dynamic effects, ignoring length scale effects, typical in micro- and nano-mechanics, as well as continuum assumptions, neglecting material and/or geometric nonlinearity, and stochasticity; (3) numerical modeling, such as finite element discretization, inaccurate force fields in molecular dynamics simulation. On top of these three aspects of uncertainty, ML methods (especially NN-based methods) introduce a few additional sources of uncertainty, including the choices of model architecture and hyperparameters, stochasticity in the training process, and transferability of the trained model, making the accurate quantification of total uncertainty a complex and time-consuming endeavor.

Uncertainty quantification (UQ) is a discipline of science focusing on identifying, quantifying, and reducing uncertainties associated with models, numerical algorithms, experiments, and predicted outcomes or quantities of interest \cite{smith2013uncertainty}. It is a broad area that has been studied extensively, which is not exclusively applicable to machine learning methods. For detailed UQ methods and their applications in ML, readers are referred to related textbooks and review papers \cite{abdar2021review, soize2017uncertainty, sullivan2015introduction}. Here, we briefly review a few UQ methods that are extensively applied to estimating and reducing uncertainty in the context of ML applications in experimental solid mechanics. To quantify the uncertainty of mechanics systems, one of the most widely adopted classes of methods is the Bayesian procedure. Built upon the well-established, century-old Bayes’ theorem \cite{RN80}, the Bayesian procedure seeks to infer the posterior distribution of variables based on prior knowledge and measured data. Specific examples of methods involving the Bayesian approach include the use of Gaussian process regression for modeling the nonlinear behavior of solids \cite{cicci2023uncertainty}, the creep behavior of concrete \cite{liang2022interpretable}, and metamaterial design \cite{RN48}, as well as NNs for crystal plasticity \cite{de2022predicting} and multiscale modeling of nanocomposite \cite{pyrialakos2021neural}. These studies, by employing the Bayesian procedure, provide a distribution of the quantities of interest rather than a single output. Another prevalent technique is the ensemble method, which combines a group of base (weak) models, which differ by the algorithm, hyperparameter, training data, and/or random seeds, to provide more accurate, more robust predictions. They may be applied to machine learning models such as decision tree (random forest), support vector machines, and neural networks. Through this method, one may estimate the uncertainty of the prediction according to the spread among base models as well as reducing the uncertainty by combining these models into an ensemble for enhanced predictive capability.

After training a machine learning model, one often needs to validate it by testing the performance of the model on unseen data. Aside from the simplest way of train-test split of the dataset, techniques including k-fold cross-validation, leave-one-out cross-validation, and bootstrapping are often employed. These techniques help to better assess the performance of the trained model and estimate the uncertainty of predictions, by efficiently utilizing the available dataset. In addition to foregoing methods related to uncertainty quantification, some other commonly invoked techniques and procedures include sensitivity analysis \cite{nguyen2021hybrid}, Monte-Carlo simulations \cite{huang2022microstructure}, and computation of confidence interval \cite{RN283}. Note that the field of UQ has developed for decades, with numerous methods proposed and investigated. In the foregoing context, we have only provided a very brief, non-exhaustive list of UQ methods that have been commonly used in mechanics. Finally, we comment that there are still a lot of problems related to quantification of uncertainty associated to ML applications in mechanics. As we analyzed at the beginning of this section, there are different sources of uncertainty throughout the research workflow of experimental mechanics, many of which have not yet been well investigated. It is a challenging yet worthwhile task to quantify the total uncertainty of the experimental mechanics workflow addressing all relevant sources of uncertainty.

\section{Outlook: Future opportunities of ML in experimental solid mechanics }

\subsection{ML for experimental mechanics under extreme conditions}

In recent years, there has been an emerging interest in characterizing material properties under extreme conditions like high strain rate, high pressure, and high temperature. However, such experiments are often considered laborious and costly, requiring significant experiment preparation time. For example, in conventional plate impact experiments, which characterize dynamic properties of materials under high strain rates, the experiment preparation time (e.g., sample preparation, optical alignment, triggering circuits connection, interferometry) alone can take several days. Therefore, there is an increasing demand to design big-data-generating experiments where new high-throughput experimental techniques can be coupled with ML methods to improve the efficiency of data collection and analysis. One recent study by Jin et al. \cite{RN11} demonstrated the benefits of big-data-generating experiments in the context of material dynamic fracture toughness and cohesive parameter determination. By leveraging a high-throughput optical interferometer and CNN-based ML model, the researchers were able to significantly increase the experimental efficiency, reducing the required number of experiments by order of magnitude. These results highlight the potential for ML to transform the field of experimental mechanics, enabling researchers to characterize material properties more efficiently and effectively under extreme conditions. Moving forward, it will be crucial to continue to develop new experimental full-field measurement techniques for experiments under extreme conditions like spatial-temporal interferometer \cite{RN11} or stereo DIC \cite{RN240} that can be effectively coupled with ML to generate and analyze large volumes of high-fidelity data. By leveraging ML for extreme mechanics, experimentalists will be capable of characterizing extreme material properties more efficiently and accurately, enabling new insights and applications in designing next-generation materials and technologies, e.g., earthquake protective coatings used in architectural design.

\subsection{Design intelligent architected materials with in-situ decision-making capabilities}

The design of materials with decision-making capability is a relatively new concept that has the potential to revolutionize material design by offering unprecedented properties. With the help of ML, architected materials can be programmed to respond to real-time stimuli based on external input. Here, we present some potential research areas in which experimental mechanics and ML can contribute to the development of these intelligent architected materials. One such area is the development of new fabrication techniques to build novel material systems with high resolution. For example, new high-throughput AM techniques, such as the hydrogel infusion AM (HIAM) method \cite{RN241}, can be employed to fabricate architected materials with complex geometries at a variety of scales. Furthermore, bottom-up approaches such as self-assembly can be employed to spontaneously organize nanoscale constituents into ordered structures through intermolecular forces. Recently, scaling-up fabrication techniques \cite{kagiasmetasurface}, like holographic lithography, were employed to build centimeter-size samples with nanoscale features. Another area of focus is the development of new actuation methods. Current actuation methods are mainly focused on passive actuation techniques like mechanical or electromagnetic actuation while designing materials with active actuations that can deform on demand according to local environmental stimuli would confer intelligence to these materials. Designing materials with active actuation would involve the development of ML algorithms capable of optimizing material structures and properties based on a set of design objectives and constraints. For example, reinforcement learning algorithms could be used to train materials to learn and respond to different loading scenarios, leading to enhanced structural performance and durability. Moreover, ML algorithms could enable materials to make intelligent decisions in real time based on environmental conditions, such as changes in temperature, humidity, or mechanical loads. This opens a wide range of potential applications for intelligent architected materials in fields such as aerospace and robotics.

\section{Conclusions}

Recent advances in ML have revolutionized the field of experimental solid mechanics, allowing for efficient and accurate experimental design, data analysis, and solutions to inverse problems. In this review paper, we highlight recent advances and applications of ML in experimental solid mechanics. We start by providing an overview of common ML algorithms and terminologies relevant to experimental mechanics, with a particular emphasis on physics-informed and physics-based scientific ML methods. Then, we reviewed recent applications of ML in traditional and emerging areas of experimental solid mechanics, including fracture mechanics, biomechanics, nano- and micro-mechanics, architected materials, and 2D materials. Furthermore, the review discusses current challenges in applying ML to problems involving data scarcity, multi-modality, and multi-fidelity experimental datasets. It also advances several future research directions to address such challenges. It is hoped that this comprehensive and up-to-date review will provide valuable insights for researchers and practitioners in solid mechanics who are interested in employing ML to design and analyze their experiments. As the field continues to evolve, it will also be essential to build bridges across disciplines, with the most obvious being computational mechanics and materials sciences, to address challenges and opportunities.

\section*{Acknowledgment}
H.D.E. acknowledges the financial support from the Air Force Office of Scientific Research (AFOSR-FA9550-20-1-0258), National Science Foundation (grant CMMI-1953806), Office of Naval Research (grant N000142212133).

\section*{Author Contribution}
H.J. and H.D.E. conceived the initial idea to write this review. H.J. wrote the initial draft except for Section 4.2 and 4.3. E.Z. wrote the initial draft for Section 4.2 and 4.3 and revised other sections. H.D.E. revised the entire manuscript. H.D.E. supervised the project. All authors gave final approval for the publication.

\bibliographystyle{elsarticle-num-names}
\bibliography{reference.bib}

\begin{thebibliography}{350}
\expandafter\ifx\csname natexlab\endcsname\relax\def\natexlab#1{#1}\fi
\providecommand{\url}[1]{\texttt{#1}}
\providecommand{\href}[2]{#2}
\providecommand{\path}[1]{#1}
\providecommand{\DOIprefix}{doi:}
\providecommand{\ArXivprefix}{arXiv:}
\providecommand{\URLprefix}{URL: }
\providecommand{\Pubmedprefix}{pmid:}
\providecommand{\doi}[1]{\href{http://dx.doi.org/#1}{\path{#1}}}
\providecommand{\Pubmed}[1]{\href{pmid:#1}{\path{#1}}}
\providecommand{\bibinfo}[2]{#2}
\ifx\xfnm\relax \def\xfnm[#1]{\unskip,\space#1}\fi
\bibitem[{Sciammarella and Sciammarella(2012)}]{RN94}
\bibinfo{author}{C.~A. Sciammarella}, \bibinfo{author}{F.~M. Sciammarella},
  \bibinfo{title}{Experimental mechanics of solids}, \bibinfo{publisher}{John
  Wiley \& Sons}, \bibinfo{year}{2012}.
\bibitem[{Kassner et~al.(2005)Kassner, Nemat-Nasser, Suo, Bao, Barbour,
  Brinson, Espinosa, Gao, Granick, Gumbsch, Kim, Knauss, Kubin, Langer, Larson,
  Mahadevan, Majumdar, Torquato, and van Swol}]{RN98}
\bibinfo{author}{M.~E. Kassner}, \bibinfo{author}{S.~Nemat-Nasser},
  \bibinfo{author}{Z.~G. Suo}, \bibinfo{author}{G.~Bao}, \bibinfo{author}{J.~C.
  Barbour}, \bibinfo{author}{L.~C. Brinson}, \bibinfo{author}{H.~Espinosa},
  \bibinfo{author}{H.~J. Gao}, \bibinfo{author}{S.~Granick},
  \bibinfo{author}{P.~Gumbsch}, \bibinfo{author}{K.~S. Kim},
  \bibinfo{author}{W.~Knauss}, \bibinfo{author}{L.~Kubin},
  \bibinfo{author}{J.~Langer}, \bibinfo{author}{B.~C. Larson},
  \bibinfo{author}{L.~Mahadevan}, \bibinfo{author}{A.~Majumdar},
  \bibinfo{author}{S.~Torquato}, \bibinfo{author}{F.~van Swol},
\newblock \bibinfo{title}{New directions in mechanics},
\newblock \bibinfo{journal}{Mechanics of Materials} \bibinfo{volume}{37}
  (\bibinfo{year}{2005}) \bibinfo{pages}{231--259}.
  \DOIprefix\doi{10.1016/j.mechmat.2004.04.009}.
\bibitem[{Hooke(1678)}]{RN111}
\bibinfo{author}{R.~Hooke}, \bibinfo{title}{Potentia Restitutiva, or Spring},
  \bibinfo{year}{1678}.
\bibitem[{Griffith(1921)}]{RN7}
\bibinfo{author}{A.~A. Griffith},
\newblock \bibinfo{title}{Vi. the phenomena of rupture and flow in solids},
\newblock \bibinfo{journal}{Philosophical transactions of the royal society of
  london. Series A} \bibinfo{volume}{221} (\bibinfo{year}{1921})
  \bibinfo{pages}{163--198}.
\bibitem[{Davis(2004)}]{RN97}
\bibinfo{author}{J.~R. Davis}, \bibinfo{title}{Tensile testing},
  \bibinfo{publisher}{ASM international}, \bibinfo{year}{2004}.
\bibitem[{Chen and Song(2010)}]{RN95}
\bibinfo{author}{W.~W. Chen}, \bibinfo{author}{B.~Song}, \bibinfo{title}{Split
  Hopkinson (Kolsky) bar: design, testing and applications},
  \bibinfo{publisher}{Springer Science \& Business Media},
  \bibinfo{year}{2010}.
\bibitem[{Abou-Sayed et~al.(1976)Abou-Sayed, Clifton, and Hermann}]{RN96}
\bibinfo{author}{A.~S. Abou-Sayed}, \bibinfo{author}{R.~J. Clifton},
  \bibinfo{author}{L.~Hermann},
\newblock \bibinfo{title}{The oblique-plate impact experiment},
\newblock \bibinfo{journal}{Experimental Mechanics} \bibinfo{volume}{16}
  (\bibinfo{year}{1976}) \bibinfo{pages}{127--132}.
  \DOIprefix\doi{10.1007/BF02321106}.
\bibitem[{Zhu and Espinosa(2005)}]{RN101}
\bibinfo{author}{Y.~Zhu}, \bibinfo{author}{H.~D. Espinosa},
\newblock \bibinfo{title}{An electromechanical material testing system for in
  situ electron microscopy and applications},
\newblock \bibinfo{journal}{Proceedings of the National Academy of Sciences of
  the United States of America} \bibinfo{volume}{102} (\bibinfo{year}{2005})
  \bibinfo{pages}{14503--14508}. \DOIprefix\doi{10.1073/pnas.0506544102}.
\bibitem[{Espinosa et~al.(2007)Espinosa, Zhu, and Moldovan}]{RN103}
\bibinfo{author}{H.~D. Espinosa}, \bibinfo{author}{Y.~Zhu},
  \bibinfo{author}{N.~Moldovan},
\newblock \bibinfo{title}{Design and operation of a mems-based material testing
  system for nanomechanical characterization},
\newblock \bibinfo{journal}{Journal of Microelectromechanical Systems}
  \bibinfo{volume}{16} (\bibinfo{year}{2007}) \bibinfo{pages}{1219--1231}.
\bibitem[{Prorok et~al.(2004)Prorok, Zhu, Espinosa, Guo, Bazant, Zhao, and
  Yakobson}]{RN102}
\bibinfo{author}{B.~C. Prorok}, \bibinfo{author}{Y.~Zhu},
  \bibinfo{author}{H.~D. Espinosa}, \bibinfo{author}{Z.~Guo},
  \bibinfo{author}{Z.~Bazant}, \bibinfo{author}{Y.~Zhao},
  \bibinfo{author}{B.~I. Yakobson}, \bibinfo{title}{Micro-and nanomechanics},
  volume~\bibinfo{volume}{5}, \bibinfo{publisher}{Citeseer},
  \bibinfo{year}{2004}, pp. \bibinfo{pages}{561--606}.
\bibitem[{Higson(1964)}]{RN99}
\bibinfo{author}{G.~Higson},
\newblock \bibinfo{title}{Recent advances in strain gauges},
\newblock \bibinfo{journal}{Journal of Scientific Instruments}
  \bibinfo{volume}{41} (\bibinfo{year}{1964}) \bibinfo{pages}{405}.
\bibitem[{Tiwari et~al.(2007)Tiwari, Sutton, and McNeill}]{RN100}
\bibinfo{author}{V.~Tiwari}, \bibinfo{author}{M.~A. Sutton},
  \bibinfo{author}{S.~R. McNeill},
\newblock \bibinfo{title}{Assessment of high speed imaging systems for 2d and
  3d deformation measurements: Methodology development and validation},
\newblock \bibinfo{journal}{Experimental Mechanics} \bibinfo{volume}{47}
  (\bibinfo{year}{2007}) \bibinfo{pages}{561--579}.
  \DOIprefix\doi{10.1007/s11340-006-9011-y}.
\bibitem[{Walker(1994)}]{RN104}
\bibinfo{author}{C.~A. Walker},
\newblock \bibinfo{title}{A historical review of moire interferometry},
\newblock \bibinfo{journal}{Experimental Mechanics} \bibinfo{volume}{34}
  (\bibinfo{year}{1994}) \bibinfo{pages}{281--299}. \DOIprefix\doi{Doi
  10.1007/Bf02325143}.
\bibitem[{Chu et~al.(1985)Chu, Ranson, Sutton, and Peters}]{RN105}
\bibinfo{author}{T.~C. Chu}, \bibinfo{author}{W.~F. Ranson},
  \bibinfo{author}{M.~A. Sutton}, \bibinfo{author}{W.~H. Peters},
\newblock \bibinfo{title}{Applications of digital-image-correlation techniques
  to experimental mechanics},
\newblock \bibinfo{journal}{Experimental Mechanics} \bibinfo{volume}{25}
  (\bibinfo{year}{1985}) \bibinfo{pages}{232--244}. \DOIprefix\doi{Doi
  10.1007/Bf02325092}.
\bibitem[{Bay et~al.(1999)Bay, Smith, Fyhrie, and Saad}]{RN106}
\bibinfo{author}{B.~K. Bay}, \bibinfo{author}{T.~S. Smith},
  \bibinfo{author}{D.~P. Fyhrie}, \bibinfo{author}{M.~Saad},
\newblock \bibinfo{title}{Digital volume correlation: Three-dimensional strain
  mapping using x-ray tomography},
\newblock \bibinfo{journal}{Experimental Mechanics} \bibinfo{volume}{39}
  (\bibinfo{year}{1999}) \bibinfo{pages}{217--226}. \DOIprefix\doi{Doi
  10.1007/Bf02323555}.
\bibitem[{Petzing and Tyrer(1998)}]{RN107}
\bibinfo{author}{J.~N. Petzing}, \bibinfo{author}{J.~R. Tyrer},
\newblock \bibinfo{title}{Recent developments and applications in electronic
  speckle pattern interferometry},
\newblock \bibinfo{journal}{Journal of Strain Analysis for Engineering Design}
  \bibinfo{volume}{33} (\bibinfo{year}{1998}) \bibinfo{pages}{153--169}.
  \DOIprefix\doi{Doi 10.1243/0309324981512887}.
\bibitem[{Doyle(2002)}]{RN185}
\bibinfo{author}{J.~Doyle},
\newblock \bibinfo{title}{Inverse methods in experimental mechanics},
\newblock \bibinfo{journal}{Recent Advances in Experimental Mechanics: In Honor
  of Isaac M. Daniel}  (\bibinfo{year}{2002}) \bibinfo{pages}{585--594}.
\bibitem[{Tanaka and Dulikravich(1998)}]{RN186}
\bibinfo{author}{M.~Tanaka}, \bibinfo{author}{G.~S. Dulikravich},
  \bibinfo{title}{Inverse problems in engineering mechanics},
  \bibinfo{publisher}{Elsevier}, \bibinfo{year}{1998}.
\bibitem[{Montemayor et~al.(2015)Montemayor, Chernow, and Greer}]{RN121}
\bibinfo{author}{L.~Montemayor}, \bibinfo{author}{V.~Chernow},
  \bibinfo{author}{J.~R. Greer},
\newblock \bibinfo{title}{Materials by design: Using architecture in material
  design to reach new property spaces},
\newblock \bibinfo{journal}{Mrs Bulletin} \bibinfo{volume}{40}
  (\bibinfo{year}{2015}) \bibinfo{pages}{1122--1129}.
  \DOIprefix\doi{10.1557/mrs.2015.263}.
\bibitem[{Xia et~al.(2022)Xia, Spadaccini, and Greer}]{RN30}
\bibinfo{author}{X.~X. Xia}, \bibinfo{author}{C.~M. Spadaccini},
  \bibinfo{author}{J.~R. Greer},
\newblock \bibinfo{title}{Responsive materials architected in space and time},
\newblock \bibinfo{journal}{Nature Reviews Materials} \bibinfo{volume}{7}
  (\bibinfo{year}{2022}) \bibinfo{pages}{683--701}.
  \DOIprefix\doi{10.1038/s41578-022-00450-z}.
\bibitem[{Mas-Balleste et~al.(2011)Mas-Balleste, Gomez-Navarro, Gomez-Herrero,
  and Zamora}]{RN109}
\bibinfo{author}{R.~Mas-Balleste}, \bibinfo{author}{C.~Gomez-Navarro},
  \bibinfo{author}{J.~Gomez-Herrero}, \bibinfo{author}{F.~Zamora},
\newblock \bibinfo{title}{2d materials: to graphene and beyond},
\newblock \bibinfo{journal}{Nanoscale} \bibinfo{volume}{3}
  (\bibinfo{year}{2011}) \bibinfo{pages}{20--30}.
  \DOIprefix\doi{10.1039/c0nr00323a}.
\bibitem[{Greer and Park(2018)}]{RN108}
\bibinfo{author}{J.~R. Greer}, \bibinfo{author}{J.~Park},
\newblock \bibinfo{title}{Additive manufacturing of nano- and microarchitected
  materials},
\newblock \bibinfo{journal}{Nano Letters} \bibinfo{volume}{18}
  (\bibinfo{year}{2018}) \bibinfo{pages}{2187--2188}.
  \DOIprefix\doi{10.1021/acs.nanolett.8b00724}.
\bibitem[{Stewart et~al.(2020)Stewart, Murray, Suzuki, Pollock, and
  Levi}]{RN187}
\bibinfo{author}{C.~A. Stewart}, \bibinfo{author}{S.~P. Murray},
  \bibinfo{author}{A.~Suzuki}, \bibinfo{author}{T.~M. Pollock},
  \bibinfo{author}{C.~G. Levi},
\newblock \bibinfo{title}{Accelerated discovery of oxidation resistant
  coni-base γ/γ’alloys with high l12 solvus and low density},
\newblock \bibinfo{journal}{Materials \& Design} \bibinfo{volume}{189}
  (\bibinfo{year}{2020}) \bibinfo{pages}{108445}.
\bibitem[{Noh et~al.(2019)Noh, Kim, ho~Gu, Shinde, Zhou, Gregoire, and
  Jung}]{RN188}
\bibinfo{author}{J.~Noh}, \bibinfo{author}{S.~Kim}, \bibinfo{author}{G.~ho~Gu},
  \bibinfo{author}{A.~Shinde}, \bibinfo{author}{L.~Zhou},
  \bibinfo{author}{J.~M. Gregoire}, \bibinfo{author}{Y.~Jung},
\newblock \bibinfo{title}{Unveiling new stable manganese based photoanode
  materials via theoretical high-throughput screening and experiments},
\newblock \bibinfo{journal}{Chemical Communications} \bibinfo{volume}{55}
  (\bibinfo{year}{2019}) \bibinfo{pages}{13418--13421}.
\bibitem[{Lin et~al.(2021)Lin, Magagnosc, Wen, Oh, Kim, and Espinosa}]{RN189}
\bibinfo{author}{Z.~Lin}, \bibinfo{author}{D.~Magagnosc},
  \bibinfo{author}{J.~Wen}, \bibinfo{author}{C.-S. Oh}, \bibinfo{author}{S.-M.
  Kim}, \bibinfo{author}{H.~Espinosa},
\newblock \bibinfo{title}{In-situ sem high strain rate testing of large
  diameter micropillars followed by tem and ebsd postmortem analysis},
\newblock \bibinfo{journal}{Experimental Mechanics} \bibinfo{volume}{61}
  (\bibinfo{year}{2021}) \bibinfo{pages}{739--752}.
\bibitem[{Mitchell and Mitchell(1997)}]{RN112}
\bibinfo{author}{T.~M. Mitchell}, \bibinfo{author}{T.~M. Mitchell},
  \bibinfo{title}{Machine learning}, volume~\bibinfo{volume}{1},
  \bibinfo{publisher}{McGraw-hill New York}, \bibinfo{year}{1997}.
\bibitem[{LeCun et~al.(2015)LeCun, Bengio, and Hinton}]{lecun2015deep}
\bibinfo{author}{Y.~LeCun}, \bibinfo{author}{Y.~Bengio},
  \bibinfo{author}{G.~Hinton},
\newblock \bibinfo{title}{Deep learning},
\newblock \bibinfo{journal}{nature} \bibinfo{volume}{521}
  (\bibinfo{year}{2015}) \bibinfo{pages}{436--444}.
\bibitem[{Krizhevsky et~al.(2017)Krizhevsky, Sutskever, and
  Hinton}]{krizhevsky2017imagenet}
\bibinfo{author}{A.~Krizhevsky}, \bibinfo{author}{I.~Sutskever},
  \bibinfo{author}{G.~E. Hinton},
\newblock \bibinfo{title}{Imagenet classification with deep convolutional
  neural networks},
\newblock \bibinfo{journal}{Communications of the ACM} \bibinfo{volume}{60}
  (\bibinfo{year}{2017}) \bibinfo{pages}{84--90}.
\bibitem[{Hinton et~al.(2012)Hinton, Deng, Yu, Dahl, Mohamed, Jaitly, Senior,
  Vanhoucke, Nguyen, Sainath et~al.}]{hinton2012deep}
\bibinfo{author}{G.~Hinton}, \bibinfo{author}{L.~Deng},
  \bibinfo{author}{D.~Yu}, \bibinfo{author}{G.~E. Dahl}, \bibinfo{author}{A.-r.
  Mohamed}, \bibinfo{author}{N.~Jaitly}, \bibinfo{author}{A.~Senior},
  \bibinfo{author}{V.~Vanhoucke}, \bibinfo{author}{P.~Nguyen},
  \bibinfo{author}{T.~N. Sainath}, et~al.,
\newblock \bibinfo{title}{Deep neural networks for acoustic modeling in speech
  recognition: The shared views of four research groups},
\newblock \bibinfo{journal}{IEEE Signal processing magazine}
  \bibinfo{volume}{29} (\bibinfo{year}{2012}) \bibinfo{pages}{82--97}.
\bibitem[{Ramos et~al.(2017)Ramos, Gehrig, Pinggera, Franke, and
  Rother}]{ramos2017detecting}
\bibinfo{author}{S.~Ramos}, \bibinfo{author}{S.~Gehrig},
  \bibinfo{author}{P.~Pinggera}, \bibinfo{author}{U.~Franke},
  \bibinfo{author}{C.~Rother},
\newblock \bibinfo{title}{Detecting unexpected obstacles for self-driving cars:
  Fusing deep learning and geometric modeling},
\newblock in: \bibinfo{booktitle}{2017 IEEE Intelligent Vehicles Symposium
  (IV)}, \bibinfo{organization}{IEEE}, \bibinfo{year}{2017}, pp.
  \bibinfo{pages}{1025--1032}.
\bibitem[{Guo et~al.(2021)Guo, Yang, Yu, and Buehler}]{RN35}
\bibinfo{author}{K.~Guo}, \bibinfo{author}{Z.~Z. Yang}, \bibinfo{author}{C.~H.
  Yu}, \bibinfo{author}{M.~J. Buehler},
\newblock \bibinfo{title}{Artificial intelligence and machine learning in
  design of mechanical materials},
\newblock \bibinfo{journal}{Materials Horizons} \bibinfo{volume}{8}
  (\bibinfo{year}{2021}) \bibinfo{pages}{1153--1172}.
  \DOIprefix\doi{10.1039/d0mh01451f}.
\bibitem[{Yang et~al.(2021)Yang, Choi, Cho, Agyapong-Fordjour, Park, Yun, Kim,
  Han, Lee, Kim et~al.}]{RN120}
\bibinfo{author}{S.-H. Yang}, \bibinfo{author}{W.~Choi}, \bibinfo{author}{B.~W.
  Cho}, \bibinfo{author}{F.~O.-T. Agyapong-Fordjour},
  \bibinfo{author}{S.~Park}, \bibinfo{author}{S.~J. Yun},
  \bibinfo{author}{H.-J. Kim}, \bibinfo{author}{Y.-K. Han},
  \bibinfo{author}{Y.~H. Lee}, \bibinfo{author}{K.~K. Kim}, et~al.,
\newblock \bibinfo{title}{Deep learning-assisted quantification of atomic
  dopants and defects in 2d materials},
\newblock \bibinfo{journal}{Advanced Science} \bibinfo{volume}{8}
  (\bibinfo{year}{2021}) \bibinfo{pages}{2101099}.
\bibitem[{Unke et~al.(2021)Unke, Chmiela, Sauceda, Gastegger, Poltaysky,
  Schutt, Tkatchenko, and Muller}]{RN119}
\bibinfo{author}{O.~T. Unke}, \bibinfo{author}{S.~Chmiela},
  \bibinfo{author}{H.~E. Sauceda}, \bibinfo{author}{M.~Gastegger},
  \bibinfo{author}{I.~Poltaysky}, \bibinfo{author}{K.~T. Schutt},
  \bibinfo{author}{A.~Tkatchenko}, \bibinfo{author}{K.~R. Muller},
\newblock \bibinfo{title}{Machine learning force fields},
\newblock \bibinfo{journal}{Chemical Reviews} \bibinfo{volume}{121}
  (\bibinfo{year}{2021}) \bibinfo{pages}{10142--10186}.
  \DOIprefix\doi{10.1021/acs.chemrev.0c01111}.
\bibitem[{Choudhary et~al.(2022)Choudhary, DeCost, Chen, Jain, Tavazza, Cohn,
  Park, Choudhary, Agrawal, Billinge et~al.}]{choudhary2022recent}
\bibinfo{author}{K.~Choudhary}, \bibinfo{author}{B.~DeCost},
  \bibinfo{author}{C.~Chen}, \bibinfo{author}{A.~Jain},
  \bibinfo{author}{F.~Tavazza}, \bibinfo{author}{R.~Cohn},
  \bibinfo{author}{C.~W. Park}, \bibinfo{author}{A.~Choudhary},
  \bibinfo{author}{A.~Agrawal}, \bibinfo{author}{S.~J. Billinge}, et~al.,
\newblock \bibinfo{title}{Recent advances and applications of deep learning
  methods in materials science},
\newblock \bibinfo{journal}{npj Computational Materials} \bibinfo{volume}{8}
  (\bibinfo{year}{2022}) \bibinfo{pages}{59}.
\bibitem[{Mueller et~al.(2016)Mueller, Kusne, and
  Ramprasad}]{mueller2016machine}
\bibinfo{author}{T.~Mueller}, \bibinfo{author}{A.~G. Kusne},
  \bibinfo{author}{R.~Ramprasad},
\newblock \bibinfo{title}{Machine learning in materials science: Recent
  progress and emerging applications},
\newblock \bibinfo{journal}{Reviews in computational chemistry}
  \bibinfo{volume}{29} (\bibinfo{year}{2016}) \bibinfo{pages}{186--273}.
\bibitem[{Butler et~al.(2018)Butler, Davies, Cartwright, Isayev, and
  Walsh}]{butler2018machine}
\bibinfo{author}{K.~T. Butler}, \bibinfo{author}{D.~W. Davies},
  \bibinfo{author}{H.~Cartwright}, \bibinfo{author}{O.~Isayev},
  \bibinfo{author}{A.~Walsh},
\newblock \bibinfo{title}{Machine learning for molecular and materials
  science},
\newblock \bibinfo{journal}{Nature} \bibinfo{volume}{559}
  (\bibinfo{year}{2018}) \bibinfo{pages}{547--555}.
\bibitem[{Wang et~al.(2020)Wang, Murdock, Kauwe, Oliynyk, Gurlo, Brgoch,
  Persson, and Sparks}]{RN116}
\bibinfo{author}{A.~Y.~T. Wang}, \bibinfo{author}{R.~J. Murdock},
  \bibinfo{author}{S.~K. Kauwe}, \bibinfo{author}{A.~O. Oliynyk},
  \bibinfo{author}{A.~Gurlo}, \bibinfo{author}{J.~Brgoch},
  \bibinfo{author}{K.~A. Persson}, \bibinfo{author}{T.~D. Sparks},
\newblock \bibinfo{title}{Machine learning for materials scientists: An
  introductory guide toward best practices},
\newblock \bibinfo{journal}{Chemistry of Materials} \bibinfo{volume}{32}
  (\bibinfo{year}{2020}) \bibinfo{pages}{4954--4965}.
  \DOIprefix\doi{10.1021/acs.chemmater.0c01907}.
\bibitem[{Himanen et~al.(2019)Himanen, Geurts, Foster, and Rinke}]{RN117}
\bibinfo{author}{L.~Himanen}, \bibinfo{author}{A.~Geurts},
  \bibinfo{author}{A.~S. Foster}, \bibinfo{author}{P.~Rinke},
\newblock \bibinfo{title}{Data-driven materials science: Status, challenges,
  and perspectives},
\newblock \bibinfo{journal}{Advanced Science} \bibinfo{volume}{6}
  (\bibinfo{year}{2019}). \DOIprefix\doi{ARTN 1900808 10.1002/advs.201900808}.
\bibitem[{Masi et~al.(2021)Masi, Stefanou, Vannucci, and
  Maffi-Berthier}]{RN209}
\bibinfo{author}{F.~Masi}, \bibinfo{author}{I.~Stefanou},
  \bibinfo{author}{P.~Vannucci}, \bibinfo{author}{V.~Maffi-Berthier},
\newblock \bibinfo{title}{Thermodynamics-based artificial neural networks for
  constitutive modeling},
\newblock \bibinfo{journal}{Journal of the Mechanics and Physics of Solids}
  \bibinfo{volume}{147} (\bibinfo{year}{2021}) \bibinfo{pages}{104277}.
\bibitem[{Linka et~al.(2021)Linka, Hillgärtner, Abdolazizi, Aydin, Itskov, and
  Cyron}]{RN208}
\bibinfo{author}{K.~Linka}, \bibinfo{author}{M.~Hillgärtner},
  \bibinfo{author}{K.~P. Abdolazizi}, \bibinfo{author}{R.~C. Aydin},
  \bibinfo{author}{M.~Itskov}, \bibinfo{author}{C.~J. Cyron},
\newblock \bibinfo{title}{Constitutive artificial neural networks: A fast and
  general approach to predictive data-driven constitutive modeling by deep
  learning},
\newblock \bibinfo{journal}{Journal of Computational Physics}
  \bibinfo{volume}{429} (\bibinfo{year}{2021}) \bibinfo{pages}{110010}.
\bibitem[{Yin et~al.(2022)Yin, Zhang, Yu, and Karniadakis}]{RN124}
\bibinfo{author}{M.~Yin}, \bibinfo{author}{E.~Zhang}, \bibinfo{author}{Y.~Yu},
  \bibinfo{author}{G.~E. Karniadakis},
\newblock \bibinfo{title}{Interfacing finite elements with deep neural
  operators for fast multiscale modeling of mechanics problems},
\newblock \bibinfo{journal}{Computer Methods in Applied Mechanics and
  Engineering} \bibinfo{volume}{402} (\bibinfo{year}{2022}).
  \DOIprefix\doi{10.1016/j.cma.2022.115027}.
\bibitem[{Alber et~al.(2019)Alber, Buganza~Tepole, Cannon, De, Dura-Bernal,
  Garikipati, Karniadakis, Lytton, Perdikaris, and Petzold}]{RN125}
\bibinfo{author}{M.~Alber}, \bibinfo{author}{A.~Buganza~Tepole},
  \bibinfo{author}{W.~R. Cannon}, \bibinfo{author}{S.~De},
  \bibinfo{author}{S.~Dura-Bernal}, \bibinfo{author}{K.~Garikipati},
  \bibinfo{author}{G.~Karniadakis}, \bibinfo{author}{W.~W. Lytton},
  \bibinfo{author}{P.~Perdikaris}, \bibinfo{author}{L.~Petzold},
\newblock \bibinfo{title}{Integrating machine learning and multiscale
  modeling—perspectives, challenges, and opportunities in the biological,
  biomedical, and behavioral sciences},
\newblock \bibinfo{journal}{NPJ digital medicine} \bibinfo{volume}{2}
  (\bibinfo{year}{2019}) \bibinfo{pages}{115}.
\bibitem[{Kumar et~al.(2020)Kumar, Tan, Zheng, and Kochmann}]{RN123}
\bibinfo{author}{S.~Kumar}, \bibinfo{author}{S.~H. Tan},
  \bibinfo{author}{L.~Zheng}, \bibinfo{author}{D.~M. Kochmann},
\newblock \bibinfo{title}{Inverse-designed spinodoid metamaterials},
\newblock \bibinfo{journal}{Npj Computational Materials} \bibinfo{volume}{6}
  (\bibinfo{year}{2020}). \DOIprefix\doi{10.1038/s41524-020-0341-6}.
\bibitem[{Lu et~al.(2020)Lu, Dao, Kumar, Ramamurty, Karniadakis, and
  Suresh}]{RN77}
\bibinfo{author}{L.~Lu}, \bibinfo{author}{M.~Dao}, \bibinfo{author}{P.~Kumar},
  \bibinfo{author}{U.~Ramamurty}, \bibinfo{author}{G.~E. Karniadakis},
  \bibinfo{author}{S.~Suresh},
\newblock \bibinfo{title}{Extraction of mechanical properties of materials
  through deep learning from instrumented indentation},
\newblock \bibinfo{journal}{Proceedings of the National Academy of Sciences of
  the United States of America} \bibinfo{volume}{117} (\bibinfo{year}{2020})
  \bibinfo{pages}{7052--7062}. \DOIprefix\doi{10.1073/pnas.1922210117}.
\bibitem[{Ni and Gao(2021)}]{ni2021deep}
\bibinfo{author}{B.~Ni}, \bibinfo{author}{H.~Gao},
\newblock \bibinfo{title}{A deep learning approach to the inverse problem of
  modulus identification in elasticity},
\newblock \bibinfo{journal}{MRS Bulletin} \bibinfo{volume}{46}
  (\bibinfo{year}{2021}) \bibinfo{pages}{19--25}.
\bibitem[{Zhang et~al.(2020)Zhang, Yin, and Karniadakis}]{zhang2020physics}
\bibinfo{author}{E.~Zhang}, \bibinfo{author}{M.~Yin}, \bibinfo{author}{G.~E.
  Karniadakis},
\newblock \bibinfo{title}{Physics-informed neural networks for nonhomogeneous
  material identification in elasticity imaging},
\newblock \bibinfo{journal}{arXiv preprint arXiv:2009.04525}
  (\bibinfo{year}{2020}).
\bibitem[{Zhang et~al.(2022)Zhang, Dao, Karniadakis, and Suresh}]{RN18}
\bibinfo{author}{E.~Zhang}, \bibinfo{author}{M.~Dao}, \bibinfo{author}{G.~E.
  Karniadakis}, \bibinfo{author}{S.~Suresh},
\newblock \bibinfo{title}{Analyses of internal structures and defects in
  materials using physics-informed neural networks},
\newblock \bibinfo{journal}{Science Advances} \bibinfo{volume}{8}
  (\bibinfo{year}{2022}). \DOIprefix\doi{10.1126/sciadv.abk0644}.
\bibitem[{Song and Jin(2023)}]{song2023identifying}
\bibinfo{author}{S.~Song}, \bibinfo{author}{H.~Jin},
\newblock \bibinfo{title}{Identifying constitutive parameters for complex
  hyperelastic solids using physics-informed neural networks},
\newblock \bibinfo{journal}{arXiv preprint arXiv:2308.15640}
  (\bibinfo{year}{2023}).
\bibitem[{Psaros et~al.(2023)Psaros, Meng, Zou, Guo, and Karniadakis}]{RN126}
\bibinfo{author}{A.~F. Psaros}, \bibinfo{author}{X.~Meng},
  \bibinfo{author}{Z.~Zou}, \bibinfo{author}{L.~Guo}, \bibinfo{author}{G.~E.
  Karniadakis},
\newblock \bibinfo{title}{Uncertainty quantification in scientific machine
  learning: Methods, metrics, and comparisons},
\newblock \bibinfo{journal}{Journal of Computational Physics}
  (\bibinfo{year}{2023}) \bibinfo{pages}{111902}.
\bibitem[{Brodnik et~al.(2023)Brodnik, Muir, Tulshibagwale, Rossin, Echlin,
  Hamel, Kramer, Pollock, Kiser, Smith et~al.}]{brodnik2023perspective}
\bibinfo{author}{N.~Brodnik}, \bibinfo{author}{C.~Muir},
  \bibinfo{author}{N.~Tulshibagwale}, \bibinfo{author}{J.~Rossin},
  \bibinfo{author}{M.~Echlin}, \bibinfo{author}{C.~Hamel},
  \bibinfo{author}{S.~Kramer}, \bibinfo{author}{T.~Pollock},
  \bibinfo{author}{J.~Kiser}, \bibinfo{author}{C.~Smith}, et~al.,
\newblock \bibinfo{title}{Perspective: Machine learning in experimental solid
  mechanics},
\newblock \bibinfo{journal}{Journal of the Mechanics and Physics of Solids}
  \bibinfo{volume}{173} (\bibinfo{year}{2023}) \bibinfo{pages}{105231}.
\bibitem[{Wang et~al.(2020)Wang, Tan, Tor, and Lim}]{wang2020machine}
\bibinfo{author}{C.~Wang}, \bibinfo{author}{X.~Tan}, \bibinfo{author}{S.~B.
  Tor}, \bibinfo{author}{C.~Lim},
\newblock \bibinfo{title}{Machine learning in additive manufacturing:
  State-of-the-art and perspectives},
\newblock \bibinfo{journal}{Additive Manufacturing} \bibinfo{volume}{36}
  (\bibinfo{year}{2020}) \bibinfo{pages}{101538}.
\bibitem[{Jin et~al.(2020)Jin, Zhang, Demir, and Gu}]{jin2020machine}
\bibinfo{author}{Z.~Jin}, \bibinfo{author}{Z.~Zhang},
  \bibinfo{author}{K.~Demir}, \bibinfo{author}{G.~X. Gu},
\newblock \bibinfo{title}{Machine learning for advanced additive
  manufacturing},
\newblock \bibinfo{journal}{Matter} \bibinfo{volume}{3} (\bibinfo{year}{2020})
  \bibinfo{pages}{1541--1556}.
\bibitem[{Qin et~al.(2022)Qin, Hu, Liu, Witherell, Wang, Rosen, Simpson, Lu,
  and Tang}]{qin2022research}
\bibinfo{author}{J.~Qin}, \bibinfo{author}{F.~Hu}, \bibinfo{author}{Y.~Liu},
  \bibinfo{author}{P.~Witherell}, \bibinfo{author}{C.~C. Wang},
  \bibinfo{author}{D.~W. Rosen}, \bibinfo{author}{T.~W. Simpson},
  \bibinfo{author}{Y.~Lu}, \bibinfo{author}{Q.~Tang},
\newblock \bibinfo{title}{Research and application of machine learning for
  additive manufacturing},
\newblock \bibinfo{journal}{Additive Manufacturing} \bibinfo{volume}{52}
  (\bibinfo{year}{2022}) \bibinfo{pages}{102691}.
\bibitem[{Zuo et~al.(2022)Zuo, Qian, Feng, Yin, Li, Fan, Han, Qian, and
  Chen}]{zuo2022deep}
\bibinfo{author}{C.~Zuo}, \bibinfo{author}{J.~Qian}, \bibinfo{author}{S.~Feng},
  \bibinfo{author}{W.~Yin}, \bibinfo{author}{Y.~Li}, \bibinfo{author}{P.~Fan},
  \bibinfo{author}{J.~Han}, \bibinfo{author}{K.~Qian},
  \bibinfo{author}{Q.~Chen},
\newblock \bibinfo{title}{Deep learning in optical metrology: a review},
\newblock \bibinfo{journal}{Light: Science \& Applications}
  \bibinfo{volume}{11} (\bibinfo{year}{2022}) \bibinfo{pages}{39}.
\bibitem[{Rumelhart et~al.(1986)Rumelhart, Hinton, and
  Williams}]{rumelhart1986learning}
\bibinfo{author}{D.~E. Rumelhart}, \bibinfo{author}{G.~E. Hinton},
  \bibinfo{author}{R.~J. Williams},
\newblock \bibinfo{title}{Learning representations by back-propagating errors},
\newblock \bibinfo{journal}{nature} \bibinfo{volume}{323}
  (\bibinfo{year}{1986}) \bibinfo{pages}{533--536}.
\bibitem[{Goodfellow et~al.(2016)Goodfellow, Bengio, and
  Courville}]{goodfellow2016deep}
\bibinfo{author}{I.~Goodfellow}, \bibinfo{author}{Y.~Bengio},
  \bibinfo{author}{A.~Courville}, \bibinfo{title}{Deep learning},
  \bibinfo{publisher}{MIT press}, \bibinfo{year}{2016}.
\bibitem[{OpenAI(2021)}]{openai2021chatgpt}
\bibinfo{author}{OpenAI}, \bibinfo{title}{{ChatGPT: AI Language Model (Version
  GPT-4)}}, \bibinfo{howpublished}{Available from
  \url{https://www.openai.com/chatgpt}}, \bibinfo{year}{2021}.
  \bibinfo{note}{Accessed: 2023-05-06}.
\bibitem[{Vaswani et~al.(2017)Vaswani, Shazeer, Parmar, Uszkoreit, Jones,
  Gomez, Kaiser, and Polosukhin}]{vaswani2017attention}
\bibinfo{author}{A.~Vaswani}, \bibinfo{author}{N.~Shazeer},
  \bibinfo{author}{N.~Parmar}, \bibinfo{author}{J.~Uszkoreit},
  \bibinfo{author}{L.~Jones}, \bibinfo{author}{A.~N. Gomez},
  \bibinfo{author}{{\L}.~Kaiser}, \bibinfo{author}{I.~Polosukhin},
\newblock \bibinfo{title}{Attention is all you need},
\newblock \bibinfo{journal}{Advances in neural information processing systems}
  \bibinfo{volume}{30} (\bibinfo{year}{2017}).
\bibitem[{Raissi et~al.(2019)Raissi, Perdikaris, and
  Karniadakis}]{raissi2019physics}
\bibinfo{author}{M.~Raissi}, \bibinfo{author}{P.~Perdikaris},
  \bibinfo{author}{G.~E. Karniadakis},
\newblock \bibinfo{title}{Physics-informed neural networks: A deep learning
  framework for solving forward and inverse problems involving nonlinear
  partial differential equations},
\newblock \bibinfo{journal}{Journal of Computational physics}
  \bibinfo{volume}{378} (\bibinfo{year}{2019}) \bibinfo{pages}{686--707}.
\bibitem[{Karniadakis et~al.(2021)Karniadakis, Kevrekidis, Lu, Perdikaris,
  Wang, and Yang}]{RN128}
\bibinfo{author}{G.~E. Karniadakis}, \bibinfo{author}{I.~G. Kevrekidis},
  \bibinfo{author}{L.~Lu}, \bibinfo{author}{P.~Perdikaris},
  \bibinfo{author}{S.~Wang}, \bibinfo{author}{L.~Yang},
\newblock \bibinfo{title}{Physics-informed machine learning},
\newblock \bibinfo{journal}{Nature Reviews Physics} \bibinfo{volume}{3}
  (\bibinfo{year}{2021}) \bibinfo{pages}{422--440}.
\bibitem[{Jin(2022)}]{RN21}
\bibinfo{author}{H.~Jin}, \bibinfo{title}{Big-data-driven multi-scale
  experimental study of nanostructured block copolymer’s dynamic toughness},
  Ph.D. thesis, Brown University, \bibinfo{year}{2022}.
\bibitem[{Von~Luxburg(2007)}]{von2007tutorial}
\bibinfo{author}{U.~Von~Luxburg},
\newblock \bibinfo{title}{A tutorial on spectral clustering},
\newblock \bibinfo{journal}{Statistics and computing} \bibinfo{volume}{17}
  (\bibinfo{year}{2007}) \bibinfo{pages}{395--416}.
\bibitem[{Muir et~al.(2021{\natexlab{a}})Muir, Swaminathan, Almansour, Sevener,
  Smith, Presby, Kiser, Pollock, and Daly}]{muir2021damage}
\bibinfo{author}{C.~Muir}, \bibinfo{author}{B.~Swaminathan},
  \bibinfo{author}{A.~Almansour}, \bibinfo{author}{K.~Sevener},
  \bibinfo{author}{C.~Smith}, \bibinfo{author}{M.~Presby},
  \bibinfo{author}{J.~Kiser}, \bibinfo{author}{T.~Pollock},
  \bibinfo{author}{S.~Daly},
\newblock \bibinfo{title}{Damage mechanism identification in composites via
  machine learning and acoustic emission},
\newblock \bibinfo{journal}{npj Computational Materials} \bibinfo{volume}{7}
  (\bibinfo{year}{2021}{\natexlab{a}}) \bibinfo{pages}{95}.
\bibitem[{Muir et~al.(2021{\natexlab{b}})Muir, Swaminathan, Fields, Almansour,
  Sevener, Smith, Presby, Kiser, Pollock, and Daly}]{muir2021machine}
\bibinfo{author}{C.~Muir}, \bibinfo{author}{B.~Swaminathan},
  \bibinfo{author}{K.~Fields}, \bibinfo{author}{A.~Almansour},
  \bibinfo{author}{K.~Sevener}, \bibinfo{author}{C.~Smith},
  \bibinfo{author}{M.~Presby}, \bibinfo{author}{J.~Kiser},
  \bibinfo{author}{T.~Pollock}, \bibinfo{author}{S.~Daly},
\newblock \bibinfo{title}{A machine learning framework for damage mechanism
  identification from acoustic emissions in unidirectional sic/sic composites},
\newblock \bibinfo{journal}{npj Computational Materials} \bibinfo{volume}{7}
  (\bibinfo{year}{2021}{\natexlab{b}}) \bibinfo{pages}{146}.
\bibitem[{Agarap(2018)}]{agarap2018deep}
\bibinfo{author}{A.~F. Agarap},
\newblock \bibinfo{title}{Deep learning using rectified linear units (relu)},
\newblock \bibinfo{journal}{arXiv preprint arXiv:1803.08375}
  (\bibinfo{year}{2018}).
\bibitem[{LeCun et~al.(1998)LeCun, Bottou, Bengio, and
  Haffner}]{lecun1998gradient}
\bibinfo{author}{Y.~LeCun}, \bibinfo{author}{L.~Bottou},
  \bibinfo{author}{Y.~Bengio}, \bibinfo{author}{P.~Haffner},
\newblock \bibinfo{title}{Gradient-based learning applied to document
  recognition},
\newblock \bibinfo{journal}{Proceedings of the IEEE} \bibinfo{volume}{86}
  (\bibinfo{year}{1998}) \bibinfo{pages}{2278--2324}.
\bibitem[{LeCun et~al.(1989)LeCun, Boser, Denker, Henderson, Howard, Hubbard,
  and Jackel}]{lecun1989backpropagation}
\bibinfo{author}{Y.~LeCun}, \bibinfo{author}{B.~Boser}, \bibinfo{author}{J.~S.
  Denker}, \bibinfo{author}{D.~Henderson}, \bibinfo{author}{R.~E. Howard},
  \bibinfo{author}{W.~Hubbard}, \bibinfo{author}{L.~D. Jackel},
\newblock \bibinfo{title}{Backpropagation applied to handwritten zip code
  recognition},
\newblock \bibinfo{journal}{Neural computation} \bibinfo{volume}{1}
  (\bibinfo{year}{1989}) \bibinfo{pages}{541--551}.
\bibitem[{Holm et~al.(2020)Holm, Cohn, Gao, Kitahara, Matson, Lei, and
  Yarasi}]{RN129}
\bibinfo{author}{E.~A. Holm}, \bibinfo{author}{R.~Cohn},
  \bibinfo{author}{N.~Gao}, \bibinfo{author}{A.~R. Kitahara},
  \bibinfo{author}{T.~P. Matson}, \bibinfo{author}{B.~Lei},
  \bibinfo{author}{S.~R. Yarasi},
\newblock \bibinfo{title}{Overview: Computer vision and machine learning for
  microstructural characterization and analysis},
\newblock \bibinfo{journal}{Metallurgical and Materials Transactions a-Physical
  Metallurgy and Materials Science} \bibinfo{volume}{51} (\bibinfo{year}{2020})
  \bibinfo{pages}{5985--5999}. \DOIprefix\doi{10.1007/s11661-020-06008-4}.
\bibitem[{Jin et~al.(2022)Jin, Jiao, Clifton, and Kim}]{RN11}
\bibinfo{author}{H.~Jin}, \bibinfo{author}{T.~Jiao},
  \bibinfo{author}{R.~Clifton}, \bibinfo{author}{K.-S. Kim},
\newblock \bibinfo{title}{Dynamic fracture of a bicontinuously nanostructured
  copolymer: A deep-learning analysis of big-data-generating experiment},
\newblock \bibinfo{journal}{Journal of the Mechanics and Physics of Solids}
  \bibinfo{volume}{164} (\bibinfo{year}{2022}).
  \DOIprefix\doi{10.1016/j.jmps.2022.104898}.
\bibitem[{Kaviani and Kolinski(2022)}]{RN130}
\bibinfo{author}{R.~Kaviani}, \bibinfo{author}{J.~M. Kolinski},
\newblock \bibinfo{title}{High resolution interferometric imaging of
  liquid-solid interfaces with hotnnet},
\newblock \bibinfo{journal}{Experimental Mechanics}  (\bibinfo{year}{2022}).
  \DOIprefix\doi{10.1007/s11340-022-00912-z}.
\bibitem[{Landauer et~al.(2018)Landauer, Patel, Henann, and Franck}]{RN132}
\bibinfo{author}{A.~K. Landauer}, \bibinfo{author}{M.~Patel},
  \bibinfo{author}{D.~L. Henann}, \bibinfo{author}{C.~Franck},
\newblock \bibinfo{title}{A q-factor-based digital image correlation algorithm
  (qdic) for resolving finite deformations with degenerate speckle patterns},
\newblock \bibinfo{journal}{Experimental Mechanics} \bibinfo{volume}{58}
  (\bibinfo{year}{2018}) \bibinfo{pages}{815--830}.
  \DOIprefix\doi{10.1007/s11340-018-0377-4}.
\bibitem[{Yang and Bhattacharya(2021)}]{RN133}
\bibinfo{author}{J.~Yang}, \bibinfo{author}{K.~Bhattacharya},
\newblock \bibinfo{title}{Fast adaptive mesh augmented lagrangian digital image
  correlation},
\newblock \bibinfo{journal}{Experimental Mechanics} \bibinfo{volume}{61}
  (\bibinfo{year}{2021}) \bibinfo{pages}{719--735}.
  \DOIprefix\doi{10.1007/s11340-021-00695-9}.
\bibitem[{Yang and Bhattacharya(2019)}]{RN134}
\bibinfo{author}{J.~Yang}, \bibinfo{author}{K.~Bhattacharya},
\newblock \bibinfo{title}{Augmented lagrangian digital image correlation},
\newblock \bibinfo{journal}{Experimental Mechanics} \bibinfo{volume}{59}
  (\bibinfo{year}{2019}) \bibinfo{pages}{187--205}.
  \DOIprefix\doi{10.1007/s11340-018-00457-0}.
\bibitem[{Yang et~al.(2022)Yang, Li, Zeng, and Guo}]{RN131}
\bibinfo{author}{R.~Yang}, \bibinfo{author}{Y.~Li}, \bibinfo{author}{D.~Zeng},
  \bibinfo{author}{P.~Guo},
\newblock \bibinfo{title}{Deep dic: Deep learning-based digital image
  correlation for end-to-end displacement and strain measurement},
\newblock \bibinfo{journal}{Journal of Materials Processing Technology}
  \bibinfo{volume}{302} (\bibinfo{year}{2022}). \DOIprefix\doi{ARTN 117474
  10.1016/j.jmatprotec.2021.117474}.
\bibitem[{Patino et~al.(2022)Patino, Pathak, Mukherjee, Park, Bao, and
  Espinosa}]{RN245}
\bibinfo{author}{C.~A. Patino}, \bibinfo{author}{N.~Pathak},
  \bibinfo{author}{P.~Mukherjee}, \bibinfo{author}{S.~H. Park},
  \bibinfo{author}{G.~Bao}, \bibinfo{author}{H.~D. Espinosa},
\newblock \bibinfo{title}{Multiplexed high-throughput localized electroporation
  workflow with deep learning–based analysis for cell engineering},
\newblock \bibinfo{journal}{Science Advances} \bibinfo{volume}{8}
  (\bibinfo{year}{2022}) \bibinfo{pages}{eabn7637}.
\bibitem[{Mukherjee et~al.(2022)Mukherjee, Patino, Pathak, Lemaitre, and
  Espinosa}]{RN191}
\bibinfo{author}{P.~Mukherjee}, \bibinfo{author}{C.~A. Patino},
  \bibinfo{author}{N.~Pathak}, \bibinfo{author}{V.~Lemaitre},
  \bibinfo{author}{H.~D. Espinosa},
\newblock \bibinfo{title}{Deep learning‐assisted automated single cell
  electroporation platform for effective genetic manipulation of
  hard‐to‐transfect cells},
\newblock \bibinfo{journal}{Small} \bibinfo{volume}{18} (\bibinfo{year}{2022})
  \bibinfo{pages}{2107795}.
\bibitem[{Hochreiter and Schmidhuber(1997)}]{hochreiter1997long}
\bibinfo{author}{S.~Hochreiter}, \bibinfo{author}{J.~Schmidhuber},
\newblock \bibinfo{title}{Long short-term memory},
\newblock \bibinfo{journal}{Neural computation} \bibinfo{volume}{9}
  (\bibinfo{year}{1997}) \bibinfo{pages}{1735--1780}.
\bibitem[{Hsu et~al.(2020)Hsu, Yu, and Buehler}]{hsu2020using}
\bibinfo{author}{Y.-C. Hsu}, \bibinfo{author}{C.-H. Yu}, \bibinfo{author}{M.~J.
  Buehler},
\newblock \bibinfo{title}{Using deep learning to predict fracture patterns in
  crystalline solids},
\newblock \bibinfo{journal}{Matter} \bibinfo{volume}{3} (\bibinfo{year}{2020})
  \bibinfo{pages}{197--211}.
\bibitem[{Lew et~al.(2021)Lew, Yu, Hsu, and Buehler}]{lew2021deep}
\bibinfo{author}{A.~J. Lew}, \bibinfo{author}{C.-H. Yu}, \bibinfo{author}{Y.-C.
  Hsu}, \bibinfo{author}{M.~J. Buehler},
\newblock \bibinfo{title}{Deep learning model to predict fracture mechanisms of
  graphene},
\newblock \bibinfo{journal}{npj 2D Materials and Applications}
  \bibinfo{volume}{5} (\bibinfo{year}{2021}) \bibinfo{pages}{48}.
\bibitem[{Lew and Buehler(2021)}]{lew2021deep2}
\bibinfo{author}{A.~J. Lew}, \bibinfo{author}{M.~J. Buehler},
\newblock \bibinfo{title}{A deep learning augmented genetic algorithm approach
  to polycrystalline 2d material fracture discovery and design},
\newblock \bibinfo{journal}{Applied Physics Reviews} \bibinfo{volume}{8}
  (\bibinfo{year}{2021}) \bibinfo{pages}{041414}.
\bibitem[{Mozaffar et~al.(2019)Mozaffar, Bostanabad, Chen, Ehmann, Cao, and
  Bessa}]{mozaffar2019deep}
\bibinfo{author}{M.~Mozaffar}, \bibinfo{author}{R.~Bostanabad},
  \bibinfo{author}{W.~Chen}, \bibinfo{author}{K.~Ehmann},
  \bibinfo{author}{J.~Cao}, \bibinfo{author}{M.~Bessa},
\newblock \bibinfo{title}{Deep learning predicts path-dependent plasticity},
\newblock \bibinfo{journal}{Proceedings of the National Academy of Sciences}
  \bibinfo{volume}{116} (\bibinfo{year}{2019}) \bibinfo{pages}{26414--26420}.
\bibitem[{Scarselli et~al.(2008)Scarselli, Gori, Tsoi, Hagenbuchner, and
  Monfardini}]{RN139}
\bibinfo{author}{F.~Scarselli}, \bibinfo{author}{M.~Gori},
  \bibinfo{author}{A.~C. Tsoi}, \bibinfo{author}{M.~Hagenbuchner},
  \bibinfo{author}{G.~Monfardini},
\newblock \bibinfo{title}{The graph neural network model},
\newblock \bibinfo{journal}{IEEE transactions on neural networks}
  \bibinfo{volume}{20} (\bibinfo{year}{2008}) \bibinfo{pages}{61--80}.
\bibitem[{Wu et~al.(2022)Wu, Sun, Zhang, Xie, and Cui}]{RN140}
\bibinfo{author}{S.~Wu}, \bibinfo{author}{F.~Sun}, \bibinfo{author}{W.~Zhang},
  \bibinfo{author}{X.~Xie}, \bibinfo{author}{B.~Cui},
\newblock \bibinfo{title}{Graph neural networks in recommender systems: a
  survey},
\newblock \bibinfo{journal}{ACM Computing Surveys} \bibinfo{volume}{55}
  (\bibinfo{year}{2022}) \bibinfo{pages}{1--37}.
\bibitem[{Fan et~al.(2019)Fan, Ma, Li, He, Zhao, Tang, and Yin}]{RN142}
\bibinfo{author}{W.~Fan}, \bibinfo{author}{Y.~Ma}, \bibinfo{author}{Q.~Li},
  \bibinfo{author}{Y.~He}, \bibinfo{author}{E.~Zhao},
  \bibinfo{author}{J.~Tang}, \bibinfo{author}{D.~Yin},
\newblock \bibinfo{title}{Graph neural networks for social recommendation},
\newblock in: \bibinfo{booktitle}{The world wide web conference},
  \bibinfo{year}{2019}, pp. \bibinfo{pages}{417--426}.
\bibitem[{Xiong et~al.(2019)Xiong, Wang, Liu, Zhong, Wan, Li, Li, Luo, Chen,
  and Jiang}]{RN141}
\bibinfo{author}{Z.~Xiong}, \bibinfo{author}{D.~Wang},
  \bibinfo{author}{X.~Liu}, \bibinfo{author}{F.~Zhong},
  \bibinfo{author}{X.~Wan}, \bibinfo{author}{X.~Li}, \bibinfo{author}{Z.~Li},
  \bibinfo{author}{X.~Luo}, \bibinfo{author}{K.~Chen},
  \bibinfo{author}{H.~Jiang},
\newblock \bibinfo{title}{Pushing the boundaries of molecular representation
  for drug discovery with the graph attention mechanism},
\newblock \bibinfo{journal}{Journal of medicinal chemistry}
  \bibinfo{volume}{63} (\bibinfo{year}{2019}) \bibinfo{pages}{8749--8760}.
\bibitem[{Xie and Grossman(2018)}]{RN146}
\bibinfo{author}{T.~Xie}, \bibinfo{author}{J.~C. Grossman},
\newblock \bibinfo{title}{Crystal graph convolutional neural networks for an
  accurate and interpretable prediction of material properties},
\newblock \bibinfo{journal}{Physical review letters} \bibinfo{volume}{120}
  (\bibinfo{year}{2018}) \bibinfo{pages}{145301}.
\bibitem[{Guo and Buehler(2022)}]{RN143}
\bibinfo{author}{K.~Guo}, \bibinfo{author}{M.~J. Buehler},
\newblock \bibinfo{title}{Rapid prediction of protein natural frequencies using
  graph neural networks},
\newblock \bibinfo{journal}{Digital Discovery} \bibinfo{volume}{1}
  (\bibinfo{year}{2022}) \bibinfo{pages}{277--285}.
\bibitem[{Guo and Buehler(2020)}]{RN144}
\bibinfo{author}{K.~Guo}, \bibinfo{author}{M.~J. Buehler},
\newblock \bibinfo{title}{A semi-supervised approach to architected materials
  design using graph neural networks},
\newblock \bibinfo{journal}{Extreme Mechanics Letters} \bibinfo{volume}{41}
  (\bibinfo{year}{2020}) \bibinfo{pages}{101029}.
\bibitem[{Xue et~al.(2023)Xue, Adriaenssens, and Mao}]{RN145}
\bibinfo{author}{T.~Xue}, \bibinfo{author}{S.~Adriaenssens},
  \bibinfo{author}{S.~Mao},
\newblock \bibinfo{title}{Learning the nonlinear dynamics of mechanical
  metamaterials with graph networks},
\newblock \bibinfo{journal}{International Journal of Mechanical Sciences}
  \bibinfo{volume}{238} (\bibinfo{year}{2023}) \bibinfo{pages}{107835}.
\bibitem[{Hestroffer et~al.(2023)Hestroffer, Charpagne, Latypov, and
  Beyerlein}]{hestroffer2023graph}
\bibinfo{author}{J.~M. Hestroffer}, \bibinfo{author}{M.-A. Charpagne},
  \bibinfo{author}{M.~I. Latypov}, \bibinfo{author}{I.~J. Beyerlein},
\newblock \bibinfo{title}{Graph neural networks for efficient learning of
  mechanical properties of polycrystals},
\newblock \bibinfo{journal}{Computational Materials Science}
  \bibinfo{volume}{217} (\bibinfo{year}{2023}) \bibinfo{pages}{111894}.
\bibitem[{Thomas et~al.(2023)Thomas, Durmaz, Alam, Gumbsch, Sack, and
  Eberl}]{thomas2023materials}
\bibinfo{author}{A.~Thomas}, \bibinfo{author}{A.~Durmaz},
  \bibinfo{author}{M.~Alam}, \bibinfo{author}{P.~Gumbsch},
  \bibinfo{author}{H.~Sack}, \bibinfo{author}{C.~Eberl},
\newblock \bibinfo{title}{Materials fatigue prediction using graph neural
  networks on microstructure representations}  (\bibinfo{year}{2023}).
\bibitem[{Goodfellow et~al.(2020)Goodfellow, Pouget-Abadie, Mirza, Xu,
  Warde-Farley, Ozair, Courville, and Bengio}]{RN135}
\bibinfo{author}{I.~Goodfellow}, \bibinfo{author}{J.~Pouget-Abadie},
  \bibinfo{author}{M.~Mirza}, \bibinfo{author}{B.~Xu},
  \bibinfo{author}{D.~Warde-Farley}, \bibinfo{author}{S.~Ozair},
  \bibinfo{author}{A.~Courville}, \bibinfo{author}{Y.~Bengio},
\newblock \bibinfo{title}{Generative adversarial networks},
\newblock \bibinfo{journal}{Communications of the ACM} \bibinfo{volume}{63}
  (\bibinfo{year}{2020}) \bibinfo{pages}{139--144}.
\bibitem[{Holt and Roth(2004)}]{RN136}
\bibinfo{author}{C.~A. Holt}, \bibinfo{author}{A.~E. Roth},
\newblock \bibinfo{title}{The nash equilibrium: A perspective},
\newblock \bibinfo{journal}{Proceedings of the National Academy of Sciences}
  \bibinfo{volume}{101} (\bibinfo{year}{2004}) \bibinfo{pages}{3999--4002}.
\bibitem[{Mao et~al.(2020)Mao, He, and Zhao}]{RN46}
\bibinfo{author}{Y.~W. Mao}, \bibinfo{author}{Q.~He}, \bibinfo{author}{X.~H.
  Zhao},
\newblock \bibinfo{title}{Designing complex architectured materials with
  generative adversarial networks},
\newblock \bibinfo{journal}{Science Advances} \bibinfo{volume}{6}
  (\bibinfo{year}{2020}). \DOIprefix\doi{ARTN eaaz4169 10.1126/sciadv.aaz4169}.
\bibitem[{Kench and Cooper(2021)}]{RN137}
\bibinfo{author}{S.~Kench}, \bibinfo{author}{S.~J. Cooper},
\newblock \bibinfo{title}{Generating three-dimensional structures from a
  two-dimensional slice with generative adversarial network-based
  dimensionality expansion},
\newblock \bibinfo{journal}{Nature Machine Intelligence} \bibinfo{volume}{3}
  (\bibinfo{year}{2021}) \bibinfo{pages}{299--305}.
\bibitem[{Salimans et~al.(2016)Salimans, Goodfellow, Zaremba, Cheung, Radford,
  and Chen}]{salimans2016improved}
\bibinfo{author}{T.~Salimans}, \bibinfo{author}{I.~Goodfellow},
  \bibinfo{author}{W.~Zaremba}, \bibinfo{author}{V.~Cheung},
  \bibinfo{author}{A.~Radford}, \bibinfo{author}{X.~Chen},
\newblock \bibinfo{title}{Improved techniques for training gans},
\newblock \bibinfo{journal}{Advances in neural information processing systems}
  \bibinfo{volume}{29} (\bibinfo{year}{2016}).
\bibitem[{Cang et~al.(2017)Cang, Xu, Chen, Liu, Jiao, and
  Yi~Ren}]{cang2017microstructure}
\bibinfo{author}{R.~Cang}, \bibinfo{author}{Y.~Xu}, \bibinfo{author}{S.~Chen},
  \bibinfo{author}{Y.~Liu}, \bibinfo{author}{Y.~Jiao},
  \bibinfo{author}{M.~Yi~Ren},
\newblock \bibinfo{title}{Microstructure representation and reconstruction of
  heterogeneous materials via deep belief network for computational material
  design},
\newblock \bibinfo{journal}{Journal of Mechanical Design} \bibinfo{volume}{139}
  (\bibinfo{year}{2017}) \bibinfo{pages}{071404}.
\bibitem[{Mirza and Osindero(2014)}]{mirza2014conditional}
\bibinfo{author}{M.~Mirza}, \bibinfo{author}{S.~Osindero},
\newblock \bibinfo{title}{Conditional generative adversarial nets},
\newblock \bibinfo{journal}{arXiv preprint arXiv:1411.1784}
  (\bibinfo{year}{2014}).
\bibitem[{Nie et~al.(2021)Nie, Lin, Jiang, and Kara}]{RN138}
\bibinfo{author}{Z.~Nie}, \bibinfo{author}{T.~Lin}, \bibinfo{author}{H.~Jiang},
  \bibinfo{author}{L.~B. Kara},
\newblock \bibinfo{title}{Topologygan: Topology optimization using generative
  adversarial networks based on physical fields over the initial domain},
\newblock \bibinfo{journal}{Journal of Mechanical Design} \bibinfo{volume}{143}
  (\bibinfo{year}{2021}).
\bibitem[{Isola et~al.(2017)Isola, Zhu, Zhou, and Efros}]{isola2017image}
\bibinfo{author}{P.~Isola}, \bibinfo{author}{J.-Y. Zhu},
  \bibinfo{author}{T.~Zhou}, \bibinfo{author}{A.~A. Efros},
\newblock \bibinfo{title}{Image-to-image translation with conditional
  adversarial networks},
\newblock in: \bibinfo{booktitle}{Proceedings of the IEEE conference on
  computer vision and pattern recognition}, \bibinfo{year}{2017}, pp.
  \bibinfo{pages}{1125--1134}.
\bibitem[{Rezaei et~al.(2018)Rezaei, Harmuth, Gierke, Kellermeier, Fischer,
  Yang, and Meinel}]{rezaei2018conditional}
\bibinfo{author}{M.~Rezaei}, \bibinfo{author}{K.~Harmuth},
  \bibinfo{author}{W.~Gierke}, \bibinfo{author}{T.~Kellermeier},
  \bibinfo{author}{M.~Fischer}, \bibinfo{author}{H.~Yang},
  \bibinfo{author}{C.~Meinel},
\newblock \bibinfo{title}{A conditional adversarial network for semantic
  segmentation of brain tumor},
\newblock in: \bibinfo{booktitle}{Brainlesion: Glioma, Multiple Sclerosis,
  Stroke and Traumatic Brain Injuries: Third International Workshop, BrainLes
  2017, Held in Conjunction with MICCAI 2017, Quebec City, QC, Canada,
  September 14, 2017, Revised Selected Papers 3},
  \bibinfo{organization}{Springer}, \bibinfo{year}{2018}, pp.
  \bibinfo{pages}{241--252}.
\bibitem[{Yang et~al.(2021{\natexlab{a}})Yang, Yu, Guo, and
  Buehler}]{yang2021end}
\bibinfo{author}{Z.~Yang}, \bibinfo{author}{C.-H. Yu},
  \bibinfo{author}{K.~Guo}, \bibinfo{author}{M.~J. Buehler},
\newblock \bibinfo{title}{End-to-end deep learning method to predict complete
  strain and stress tensors for complex hierarchical composite
  microstructures},
\newblock \bibinfo{journal}{Journal of the Mechanics and Physics of Solids}
  \bibinfo{volume}{154} (\bibinfo{year}{2021}{\natexlab{a}})
  \bibinfo{pages}{104506}.
\bibitem[{Yang et~al.(2021{\natexlab{b}})Yang, Yu, and Buehler}]{yang2021deep}
\bibinfo{author}{Z.~Yang}, \bibinfo{author}{C.-H. Yu}, \bibinfo{author}{M.~J.
  Buehler},
\newblock \bibinfo{title}{Deep learning model to predict complex stress and
  strain fields in hierarchical composites},
\newblock \bibinfo{journal}{Science Advances} \bibinfo{volume}{7}
  (\bibinfo{year}{2021}{\natexlab{b}}) \bibinfo{pages}{eabd7416}.
\bibitem[{Mousavi et~al.(2018)Mousavi, Schukat, and Howley}]{mousavi2018deep}
\bibinfo{author}{S.~S. Mousavi}, \bibinfo{author}{M.~Schukat},
  \bibinfo{author}{E.~Howley},
\newblock \bibinfo{title}{Deep reinforcement learning: an overview},
\newblock in: \bibinfo{booktitle}{Proceedings of SAI Intelligent Systems
  Conference (IntelliSys) 2016: Volume 2}, \bibinfo{organization}{Springer},
  \bibinfo{year}{2018}, pp. \bibinfo{pages}{426--440}.
\bibitem[{Mnih et~al.(2015)Mnih, Kavukcuoglu, Silver, Rusu, Veness, Bellemare,
  Graves, Riedmiller, Fidjeland, Ostrovski et~al.}]{mnih2015human}
\bibinfo{author}{V.~Mnih}, \bibinfo{author}{K.~Kavukcuoglu},
  \bibinfo{author}{D.~Silver}, \bibinfo{author}{A.~A. Rusu},
  \bibinfo{author}{J.~Veness}, \bibinfo{author}{M.~G. Bellemare},
  \bibinfo{author}{A.~Graves}, \bibinfo{author}{M.~Riedmiller},
  \bibinfo{author}{A.~K. Fidjeland}, \bibinfo{author}{G.~Ostrovski}, et~al.,
\newblock \bibinfo{title}{Human-level control through deep reinforcement
  learning},
\newblock \bibinfo{journal}{nature} \bibinfo{volume}{518}
  (\bibinfo{year}{2015}) \bibinfo{pages}{529--533}.
\bibitem[{Lillicrap et~al.(2015)Lillicrap, Hunt, Pritzel, Heess, Erez, Tassa,
  Silver, and Wierstra}]{lillicrap2015continuous}
\bibinfo{author}{T.~P. Lillicrap}, \bibinfo{author}{J.~J. Hunt},
  \bibinfo{author}{A.~Pritzel}, \bibinfo{author}{N.~Heess},
  \bibinfo{author}{T.~Erez}, \bibinfo{author}{Y.~Tassa},
  \bibinfo{author}{D.~Silver}, \bibinfo{author}{D.~Wierstra},
\newblock \bibinfo{title}{Continuous control with deep reinforcement learning},
\newblock \bibinfo{journal}{arXiv preprint arXiv:1509.02971}
  (\bibinfo{year}{2015}).
\bibitem[{Silver et~al.(2016)Silver, Huang, Maddison, Guez, Sifre, Van
  Den~Driessche, Schrittwieser, Antonoglou, Panneershelvam, Lanctot
  et~al.}]{silver2016mastering}
\bibinfo{author}{D.~Silver}, \bibinfo{author}{A.~Huang}, \bibinfo{author}{C.~J.
  Maddison}, \bibinfo{author}{A.~Guez}, \bibinfo{author}{L.~Sifre},
  \bibinfo{author}{G.~Van Den~Driessche}, \bibinfo{author}{J.~Schrittwieser},
  \bibinfo{author}{I.~Antonoglou}, \bibinfo{author}{V.~Panneershelvam},
  \bibinfo{author}{M.~Lanctot}, et~al.,
\newblock \bibinfo{title}{Mastering the game of go with deep neural networks
  and tree search},
\newblock \bibinfo{journal}{nature} \bibinfo{volume}{529}
  (\bibinfo{year}{2016}) \bibinfo{pages}{484--489}.
\bibitem[{Singh et~al.(2022)Singh, Kumar, and Singh}]{singh2022reinforcement}
\bibinfo{author}{B.~Singh}, \bibinfo{author}{R.~Kumar}, \bibinfo{author}{V.~P.
  Singh},
\newblock \bibinfo{title}{Reinforcement learning in robotic applications: a
  comprehensive survey},
\newblock \bibinfo{journal}{Artificial Intelligence Review}
  (\bibinfo{year}{2022}) \bibinfo{pages}{1--46}.
\bibitem[{Popova et~al.(2018)Popova, Isayev, and Tropsha}]{popova2018deep}
\bibinfo{author}{M.~Popova}, \bibinfo{author}{O.~Isayev},
  \bibinfo{author}{A.~Tropsha},
\newblock \bibinfo{title}{Deep reinforcement learning for de novo drug design},
\newblock \bibinfo{journal}{Science advances} \bibinfo{volume}{4}
  (\bibinfo{year}{2018}) \bibinfo{pages}{eaap7885}.
\bibitem[{Garnier et~al.(2021)Garnier, Viquerat, Rabault, Larcher, Kuhnle, and
  Hachem}]{garnier2021review}
\bibinfo{author}{P.~Garnier}, \bibinfo{author}{J.~Viquerat},
  \bibinfo{author}{J.~Rabault}, \bibinfo{author}{A.~Larcher},
  \bibinfo{author}{A.~Kuhnle}, \bibinfo{author}{E.~Hachem},
\newblock \bibinfo{title}{A review on deep reinforcement learning for fluid
  mechanics},
\newblock \bibinfo{journal}{Computers \& Fluids} \bibinfo{volume}{225}
  (\bibinfo{year}{2021}) \bibinfo{pages}{104973}.
\bibitem[{Sui et~al.(2021)Sui, Guo, Zhang, Gu, and Lin}]{sui2021deep}
\bibinfo{author}{F.~Sui}, \bibinfo{author}{R.~Guo}, \bibinfo{author}{Z.~Zhang},
  \bibinfo{author}{G.~X. Gu}, \bibinfo{author}{L.~Lin},
\newblock \bibinfo{title}{Deep reinforcement learning for digital materials
  design},
\newblock \bibinfo{journal}{ACS Materials Letters} \bibinfo{volume}{3}
  (\bibinfo{year}{2021}) \bibinfo{pages}{1433--1439}.
\bibitem[{Nguyen et~al.(2022)Nguyen, Vlassis, Bahmani, Sun, Udaykumar, and
  Baek}]{nguyen2022synthesizing}
\bibinfo{author}{P.~C. Nguyen}, \bibinfo{author}{N.~N. Vlassis},
  \bibinfo{author}{B.~Bahmani}, \bibinfo{author}{W.~Sun},
  \bibinfo{author}{H.~Udaykumar}, \bibinfo{author}{S.~S. Baek},
\newblock \bibinfo{title}{Synthesizing controlled microstructures of porous
  media using generative adversarial networks and reinforcement learning},
\newblock \bibinfo{journal}{Scientific reports} \bibinfo{volume}{12}
  (\bibinfo{year}{2022}) \bibinfo{pages}{9034}.
\bibitem[{Box and Tiao(2011)}]{box2011bayesian}
\bibinfo{author}{G.~E. Box}, \bibinfo{author}{G.~C. Tiao},
  \bibinfo{title}{Bayesian inference in statistical analysis},
  \bibinfo{publisher}{John Wiley \& Sons}, \bibinfo{year}{2011}.
\bibitem[{Rappel et~al.(2020)Rappel, Beex, Hale, Noels, and
  Bordas}]{rappel2020tutorial}
\bibinfo{author}{H.~Rappel}, \bibinfo{author}{L.~A. Beex},
  \bibinfo{author}{J.~S. Hale}, \bibinfo{author}{L.~Noels},
  \bibinfo{author}{S.~Bordas},
\newblock \bibinfo{title}{A tutorial on bayesian inference to identify material
  parameters in solid mechanics},
\newblock \bibinfo{journal}{Archives of Computational Methods in Engineering}
  \bibinfo{volume}{27} (\bibinfo{year}{2020}) \bibinfo{pages}{361--385}.
\bibitem[{Zhang et~al.(2019)Zhang, Hart, and Needleman}]{RN85}
\bibinfo{author}{Y.~Zhang}, \bibinfo{author}{J.~D. Hart},
  \bibinfo{author}{A.~Needleman},
\newblock \bibinfo{title}{Identification of plastic properties from conical
  indentation using a bayesian-type statistical approach},
\newblock \bibinfo{journal}{Journal of Applied Mechanics-Transactions of the
  Asme} \bibinfo{volume}{86} (\bibinfo{year}{2019}).
  \DOIprefix\doi{10.1115/1.4041352}.
\bibitem[{Rossin et~al.(2021)Rossin, Leser, Pusch, Frey, Murray, Torbet, Smith,
  Daly, and Pollock}]{rossin2021bayesian}
\bibinfo{author}{J.~Rossin}, \bibinfo{author}{P.~Leser},
  \bibinfo{author}{K.~Pusch}, \bibinfo{author}{C.~Frey}, \bibinfo{author}{S.~P.
  Murray}, \bibinfo{author}{C.~J. Torbet}, \bibinfo{author}{S.~Smith},
  \bibinfo{author}{S.~Daly}, \bibinfo{author}{T.~M. Pollock},
\newblock \bibinfo{title}{Bayesian inference of elastic constants and texture
  coefficients in additively manufactured cobalt-nickel superalloys using
  resonant ultrasound spectroscopy},
\newblock \bibinfo{journal}{Acta Materialia} \bibinfo{volume}{220}
  (\bibinfo{year}{2021}) \bibinfo{pages}{117287}.
\bibitem[{Rossin et~al.(2022)Rossin, Leser, Pusch, Frey, Vogel, Saville,
  Torbet, Clarke, Daly, and Pollock}]{rossin2022single}
\bibinfo{author}{J.~Rossin}, \bibinfo{author}{P.~Leser},
  \bibinfo{author}{K.~Pusch}, \bibinfo{author}{C.~Frey}, \bibinfo{author}{S.~C.
  Vogel}, \bibinfo{author}{A.~I. Saville}, \bibinfo{author}{C.~Torbet},
  \bibinfo{author}{A.~J. Clarke}, \bibinfo{author}{S.~Daly},
  \bibinfo{author}{T.~M. Pollock},
\newblock \bibinfo{title}{Single crystal elastic constants of additively
  manufactured components determined by resonant ultrasound spectroscopy},
\newblock \bibinfo{journal}{Materials Characterization} \bibinfo{volume}{192}
  (\bibinfo{year}{2022}) \bibinfo{pages}{112244}.
\bibitem[{Raissi et~al.(2017)Raissi, Perdikaris, and Karniadakis}]{RN192}
\bibinfo{author}{M.~Raissi}, \bibinfo{author}{P.~Perdikaris},
  \bibinfo{author}{G.~E. Karniadakis},
\newblock \bibinfo{title}{Physics informed deep learning (part i): Data-driven
  solutions of nonlinear partial differential equations},
\newblock \bibinfo{journal}{arXiv preprint arXiv:1711.10561}
  (\bibinfo{year}{2017}).
\bibitem[{Lagaris et~al.(1998)Lagaris, Likas, and Fotiadis}]{RN275}
\bibinfo{author}{I.~E. Lagaris}, \bibinfo{author}{A.~Likas},
  \bibinfo{author}{D.~I. Fotiadis},
\newblock \bibinfo{title}{Artificial neural networks for solving ordinary and
  partial differential equations},
\newblock \bibinfo{journal}{IEEE transactions on neural networks}
  \bibinfo{volume}{9} (\bibinfo{year}{1998}) \bibinfo{pages}{987--1000}.
\bibitem[{Psichogios and Ungar(1992)}]{RN276}
\bibinfo{author}{D.~C. Psichogios}, \bibinfo{author}{L.~H. Ungar},
\newblock \bibinfo{title}{A hybrid neural network‐first principles approach
  to process modeling},
\newblock \bibinfo{journal}{AIChE Journal} \bibinfo{volume}{38}
  (\bibinfo{year}{1992}) \bibinfo{pages}{1499--1511}.
\bibitem[{Cai et~al.(2021{\natexlab{a}})Cai, Wang, Wang, Perdikaris, and
  Karniadakis}]{RN148}
\bibinfo{author}{S.~Cai}, \bibinfo{author}{Z.~Wang}, \bibinfo{author}{S.~Wang},
  \bibinfo{author}{P.~Perdikaris}, \bibinfo{author}{G.~E. Karniadakis},
\newblock \bibinfo{title}{Physics-informed neural networks for heat transfer
  problems},
\newblock \bibinfo{journal}{Journal of Heat Transfer} \bibinfo{volume}{143}
  (\bibinfo{year}{2021}{\natexlab{a}}).
\bibitem[{Cai et~al.(2021{\natexlab{b}})Cai, Mao, Wang, Yin, and
  Karniadakis}]{RN147}
\bibinfo{author}{S.~Cai}, \bibinfo{author}{Z.~Mao}, \bibinfo{author}{Z.~Wang},
  \bibinfo{author}{M.~Yin}, \bibinfo{author}{G.~E. Karniadakis},
\newblock \bibinfo{title}{Physics-informed neural networks (pinns) for fluid
  mechanics: A review},
\newblock \bibinfo{journal}{Acta Mechanica Sinica} \bibinfo{volume}{37}
  (\bibinfo{year}{2021}{\natexlab{b}}) \bibinfo{pages}{1727--1738}.
\bibitem[{Raissi et~al.(2020)Raissi, Yazdani, and Karniadakis}]{RN193}
\bibinfo{author}{M.~Raissi}, \bibinfo{author}{A.~Yazdani},
  \bibinfo{author}{G.~E. Karniadakis},
\newblock \bibinfo{title}{Hidden fluid mechanics: Learning velocity and
  pressure fields from flow visualizations},
\newblock \bibinfo{journal}{Science} \bibinfo{volume}{367}
  (\bibinfo{year}{2020}) \bibinfo{pages}{1026--1030}.
\bibitem[{Mao et~al.(2020)Mao, Jagtap, and Karniadakis}]{RN149}
\bibinfo{author}{Z.~Mao}, \bibinfo{author}{A.~D. Jagtap},
  \bibinfo{author}{G.~E. Karniadakis},
\newblock \bibinfo{title}{Physics-informed neural networks for high-speed
  flows},
\newblock \bibinfo{journal}{Computer Methods in Applied Mechanics and
  Engineering} \bibinfo{volume}{360} (\bibinfo{year}{2020})
  \bibinfo{pages}{112789}.
\bibitem[{Rasht‐Behesht et~al.(2022)Rasht‐Behesht, Huber, Shukla, and
  Karniadakis}]{RN153}
\bibinfo{author}{M.~Rasht‐Behesht}, \bibinfo{author}{C.~Huber},
  \bibinfo{author}{K.~Shukla}, \bibinfo{author}{G.~E. Karniadakis},
\newblock \bibinfo{title}{Physics‐informed neural networks (pinns) for wave
  propagation and full waveform inversions},
\newblock \bibinfo{journal}{Journal of Geophysical Research: Solid Earth}
  \bibinfo{volume}{127} (\bibinfo{year}{2022}) \bibinfo{pages}{e2021JB023120}.
\bibitem[{Chen et~al.(2020)Chen, Lu, Karniadakis, and Dal~Negro}]{RN150}
\bibinfo{author}{Y.~Chen}, \bibinfo{author}{L.~Lu}, \bibinfo{author}{G.~E.
  Karniadakis}, \bibinfo{author}{L.~Dal~Negro},
\newblock \bibinfo{title}{Physics-informed neural networks for inverse problems
  in nano-optics and metamaterials},
\newblock \bibinfo{journal}{Optics express} \bibinfo{volume}{28}
  (\bibinfo{year}{2020}) \bibinfo{pages}{11618--11633}.
\bibitem[{Liao et~al.(2023)Liao, Xue, Jeong, Webster, Ehmann, and Cao}]{RN156}
\bibinfo{author}{S.~Liao}, \bibinfo{author}{T.~Xue},
  \bibinfo{author}{J.~Jeong}, \bibinfo{author}{S.~Webster},
  \bibinfo{author}{K.~Ehmann}, \bibinfo{author}{J.~Cao},
\newblock \bibinfo{title}{Hybrid thermal modeling of additive manufacturing
  processes using physics-informed neural networks for temperature prediction
  and parameter identification},
\newblock \bibinfo{journal}{Computational Mechanics}  (\bibinfo{year}{2023})
  \bibinfo{pages}{1--14}.
\bibitem[{Yin et~al.(2021)Yin, Zheng, Humphrey, and Karniadakis}]{RN151}
\bibinfo{author}{M.~Yin}, \bibinfo{author}{X.~Zheng}, \bibinfo{author}{J.~D.
  Humphrey}, \bibinfo{author}{G.~E. Karniadakis},
\newblock \bibinfo{title}{Non-invasive inference of thrombus material
  properties with physics-informed neural networks},
\newblock \bibinfo{journal}{Computer Methods in Applied Mechanics and
  Engineering} \bibinfo{volume}{375} (\bibinfo{year}{2021})
  \bibinfo{pages}{113603}.
\bibitem[{Lu et~al.(2021)Lu, Meng, Mao, and Karniadakis}]{RN152}
\bibinfo{author}{L.~Lu}, \bibinfo{author}{X.~Meng}, \bibinfo{author}{Z.~Mao},
  \bibinfo{author}{G.~E. Karniadakis},
\newblock \bibinfo{title}{Deepxde: A deep learning library for solving
  differential equations},
\newblock \bibinfo{journal}{SIAM review} \bibinfo{volume}{63}
  (\bibinfo{year}{2021}) \bibinfo{pages}{208--228}.
\bibitem[{Henkes et~al.(2022)Henkes, Wessels, and Mahnken}]{RN154}
\bibinfo{author}{A.~Henkes}, \bibinfo{author}{H.~Wessels},
  \bibinfo{author}{R.~Mahnken},
\newblock \bibinfo{title}{Physics informed neural networks for continuum
  micromechanics},
\newblock \bibinfo{journal}{Computer Methods in Applied Mechanics and
  Engineering} \bibinfo{volume}{393} (\bibinfo{year}{2022})
  \bibinfo{pages}{114790}.
\bibitem[{Haghighat et~al.(2021)Haghighat, Raissi, Moure, Gomez, and
  Juanes}]{RN194}
\bibinfo{author}{E.~Haghighat}, \bibinfo{author}{M.~Raissi},
  \bibinfo{author}{A.~Moure}, \bibinfo{author}{H.~Gomez},
  \bibinfo{author}{R.~Juanes},
\newblock \bibinfo{title}{A physics-informed deep learning framework for
  inversion and surrogate modeling in solid mechanics},
\newblock \bibinfo{journal}{Computer Methods in Applied Mechanics and
  Engineering} \bibinfo{volume}{379} (\bibinfo{year}{2021})
  \bibinfo{pages}{113741}.
\bibitem[{Bastek and Kochmann(2023)}]{RN195}
\bibinfo{author}{J.-H. Bastek}, \bibinfo{author}{D.~M. Kochmann},
\newblock \bibinfo{title}{Physics-informed neural networks for shell
  structures},
\newblock \bibinfo{journal}{European Journal of Mechanics-A/Solids}
  \bibinfo{volume}{97} (\bibinfo{year}{2023}) \bibinfo{pages}{104849}.
\bibitem[{Hornik et~al.(1989)Hornik, Stinchcombe, and White}]{RN196}
\bibinfo{author}{K.~Hornik}, \bibinfo{author}{M.~Stinchcombe},
  \bibinfo{author}{H.~White},
\newblock \bibinfo{title}{Multilayer feedforward networks are universal
  approximators},
\newblock \bibinfo{journal}{Neural networks} \bibinfo{volume}{2}
  (\bibinfo{year}{1989}) \bibinfo{pages}{359--366}.
\bibitem[{Chen and Chen(1995)}]{RN158}
\bibinfo{author}{T.~Chen}, \bibinfo{author}{H.~Chen},
\newblock \bibinfo{title}{Universal approximation to nonlinear operators by
  neural networks with arbitrary activation functions and its application to
  dynamical systems},
\newblock \bibinfo{journal}{IEEE Transactions on Neural Networks}
  \bibinfo{volume}{6} (\bibinfo{year}{1995}) \bibinfo{pages}{911--917}.
\bibitem[{Lu et~al.(2021)Lu, Jin, Pang, Zhang, and Karniadakis}]{RN157}
\bibinfo{author}{L.~Lu}, \bibinfo{author}{P.~Jin}, \bibinfo{author}{G.~Pang},
  \bibinfo{author}{Z.~Zhang}, \bibinfo{author}{G.~E. Karniadakis},
\newblock \bibinfo{title}{Learning nonlinear operators via deeponet based on
  the universal approximation theorem of operators},
\newblock \bibinfo{journal}{Nature machine intelligence} \bibinfo{volume}{3}
  (\bibinfo{year}{2021}) \bibinfo{pages}{218--229}.
\bibitem[{Kovachki et~al.(2021)Kovachki, Li, Liu, Azizzadenesheli,
  Bhattacharya, Stuart, and Anandkumar}]{kovachki2021neural}
\bibinfo{author}{N.~Kovachki}, \bibinfo{author}{Z.~Li},
  \bibinfo{author}{B.~Liu}, \bibinfo{author}{K.~Azizzadenesheli},
  \bibinfo{author}{K.~Bhattacharya}, \bibinfo{author}{A.~Stuart},
  \bibinfo{author}{A.~Anandkumar},
\newblock \bibinfo{title}{Neural operator: Learning maps between function
  spaces},
\newblock \bibinfo{journal}{arXiv preprint arXiv:2108.08481}
  (\bibinfo{year}{2021}).
\bibitem[{Li et~al.(2020)Li, Kovachki, Azizzadenesheli, Liu, Bhattacharya,
  Stuart, and Anandkumar}]{RN162}
\bibinfo{author}{Z.~Li}, \bibinfo{author}{N.~Kovachki},
  \bibinfo{author}{K.~Azizzadenesheli}, \bibinfo{author}{B.~Liu},
  \bibinfo{author}{K.~Bhattacharya}, \bibinfo{author}{A.~Stuart},
  \bibinfo{author}{A.~Anandkumar},
\newblock \bibinfo{title}{Fourier neural operator for parametric partial
  differential equations},
\newblock \bibinfo{journal}{arXiv preprint arXiv:2010.08895}
  (\bibinfo{year}{2020}).
\bibitem[{Goswami et~al.(2022)Goswami, Bora, Yu, and Karniadakis}]{RN159}
\bibinfo{author}{S.~Goswami}, \bibinfo{author}{A.~Bora},
  \bibinfo{author}{Y.~Yu}, \bibinfo{author}{G.~E. Karniadakis},
\newblock \bibinfo{title}{Physics-informed deep neural operators networks},
\newblock \bibinfo{journal}{arXiv preprint arXiv:2207.05748}
  (\bibinfo{year}{2022}).
\bibitem[{Lu et~al.(2022)Lu, Meng, Cai, Mao, Goswami, Zhang, and
  Karniadakis}]{RN197}
\bibinfo{author}{L.~Lu}, \bibinfo{author}{X.~Meng}, \bibinfo{author}{S.~Cai},
  \bibinfo{author}{Z.~Mao}, \bibinfo{author}{S.~Goswami},
  \bibinfo{author}{Z.~Zhang}, \bibinfo{author}{G.~E. Karniadakis},
\newblock \bibinfo{title}{A comprehensive and fair comparison of two neural
  operators (with practical extensions) based on fair data},
\newblock \bibinfo{journal}{Computer Methods in Applied Mechanics and
  Engineering} \bibinfo{volume}{393} (\bibinfo{year}{2022})
  \bibinfo{pages}{114778}.
\bibitem[{Li et~al.(2021)Li, Zheng, Kovachki, Jin, Chen, Liu, Azizzadenesheli,
  and Anandkumar}]{RN277}
\bibinfo{author}{Z.~Li}, \bibinfo{author}{H.~Zheng},
  \bibinfo{author}{N.~Kovachki}, \bibinfo{author}{D.~Jin},
  \bibinfo{author}{H.~Chen}, \bibinfo{author}{B.~Liu},
  \bibinfo{author}{K.~Azizzadenesheli}, \bibinfo{author}{A.~Anandkumar},
\newblock \bibinfo{title}{Physics-informed neural operator for learning partial
  differential equations},
\newblock \bibinfo{journal}{arXiv preprint arXiv:2111.03794}
  (\bibinfo{year}{2021}).
\bibitem[{Wang et~al.(2021)Wang, Wang, and Perdikaris}]{RN278}
\bibinfo{author}{S.~Wang}, \bibinfo{author}{H.~Wang},
  \bibinfo{author}{P.~Perdikaris},
\newblock \bibinfo{title}{Learning the solution operator of parametric partial
  differential equations with physics-informed deeponets},
\newblock \bibinfo{journal}{Science advances} \bibinfo{volume}{7}
  (\bibinfo{year}{2021}) \bibinfo{pages}{eabi8605}.
\bibitem[{Li et~al.(2022)Li, Huang, Liu, and Anandkumar}]{RN279}
\bibinfo{author}{Z.~Li}, \bibinfo{author}{D.~Z. Huang},
  \bibinfo{author}{B.~Liu}, \bibinfo{author}{A.~Anandkumar},
\newblock \bibinfo{title}{Fourier neural operator with learned deformations for
  pdes on general geometries},
\newblock \bibinfo{journal}{arXiv preprint arXiv:2207.05209}
  (\bibinfo{year}{2022}).
\bibitem[{Goswami et~al.(2022)Goswami, Yin, Yu, and Karniadakis}]{RN17}
\bibinfo{author}{S.~Goswami}, \bibinfo{author}{M.~Yin},
  \bibinfo{author}{Y.~Yu}, \bibinfo{author}{G.~E. Karniadakis},
\newblock \bibinfo{title}{A physics-informed variational deeponet for
  predicting crack path},
\newblock \bibinfo{journal}{Computer Methods in Applied Mechanics and
  Engineering} \bibinfo{volume}{391} (\bibinfo{year}{2022}).
  \DOIprefix\doi{10.1016/j.cma.2022.114587}.
\bibitem[{Zhang et~al.(2022)Zhang, Spronck, Humphrey, and Karniadakis}]{RN164}
\bibinfo{author}{E.~Zhang}, \bibinfo{author}{B.~Spronck},
  \bibinfo{author}{J.~D. Humphrey}, \bibinfo{author}{G.~E. Karniadakis},
\newblock \bibinfo{title}{G2Φnet: Relating genotype and biomechanical
  phenotype of tissues with deep learning},
\newblock \bibinfo{journal}{PLOS Computational Biology} \bibinfo{volume}{18}
  (\bibinfo{year}{2022}) \bibinfo{pages}{e1010660}.
\bibitem[{You et~al.(2022)You, Zhang, Ross, Lee, Hsu, and Yu}]{RN218}
\bibinfo{author}{H.~You}, \bibinfo{author}{Q.~Zhang}, \bibinfo{author}{C.~J.
  Ross}, \bibinfo{author}{C.-H. Lee}, \bibinfo{author}{M.-C. Hsu},
  \bibinfo{author}{Y.~Yu},
\newblock \bibinfo{title}{A physics-guided neural operator learning approach to
  model biological tissues from digital image correlation measurements},
\newblock \bibinfo{journal}{Journal of Biomechanical Engineering}
  \bibinfo{volume}{144} (\bibinfo{year}{2022}) \bibinfo{pages}{121012}.
\bibitem[{Yin et~al.(2022)Yin, Ban, Rego, Zhang, Cavinato, Humphrey, and
  Em~Karniadakis}]{RN165}
\bibinfo{author}{M.~Yin}, \bibinfo{author}{E.~Ban}, \bibinfo{author}{B.~V.
  Rego}, \bibinfo{author}{E.~Zhang}, \bibinfo{author}{C.~Cavinato},
  \bibinfo{author}{J.~D. Humphrey}, \bibinfo{author}{G.~Em~Karniadakis},
\newblock \bibinfo{title}{Simulating progressive intramural damage leading to
  aortic dissection using deeponet: an operator–regression neural network},
\newblock \bibinfo{journal}{Journal of the Royal Society Interface}
  \bibinfo{volume}{19} (\bibinfo{year}{2022}) \bibinfo{pages}{20210670}.
\bibitem[{Liu et~al.(2020)Liu, Athanasiou, Padture, Sheldon, and Gao}]{RN9}
\bibinfo{author}{X.~Liu}, \bibinfo{author}{C.~E. Athanasiou},
  \bibinfo{author}{N.~P. Padture}, \bibinfo{author}{B.~W. Sheldon},
  \bibinfo{author}{H.~J. Gao},
\newblock \bibinfo{title}{A machine learning approach to fracture mechanics
  problems},
\newblock \bibinfo{journal}{Acta Materialia} \bibinfo{volume}{190}
  (\bibinfo{year}{2020}) \bibinfo{pages}{105--112}.
  \DOIprefix\doi{10.1016/j.actamat.2020.03.016}.
\bibitem[{Liu et~al.(2021)Liu, Athanasiou, Padture, Sheldon, and Gao}]{RN10}
\bibinfo{author}{X.~Liu}, \bibinfo{author}{C.~E. Athanasiou},
  \bibinfo{author}{N.~P. Padture}, \bibinfo{author}{B.~W. Sheldon},
  \bibinfo{author}{H.~J. Gao},
\newblock \bibinfo{title}{Knowledge extraction and transfer in data-driven
  fracture mechanics},
\newblock \bibinfo{journal}{Proceedings of the National Academy of Sciences of
  the United States of America} \bibinfo{volume}{118} (\bibinfo{year}{2021}).
  \DOIprefix\doi{ARTN e2104765118 10.1073/pnas.2104765118}.
\bibitem[{Su et~al.(2021)Su, Peng, Yuan, and Li}]{RN14}
\bibinfo{author}{M.~Su}, \bibinfo{author}{H.~Peng}, \bibinfo{author}{M.~Yuan},
  \bibinfo{author}{S.~F. Li},
\newblock \bibinfo{title}{Identification of the interfacial cohesive law
  parameters of frp strips externally bonded to concrete using machine learning
  techniques},
\newblock \bibinfo{journal}{Engineering Fracture Mechanics}
  \bibinfo{volume}{247} (\bibinfo{year}{2021}). \DOIprefix\doi{ARTN 107643
  10.1016/j.engfracmech.2021.107643}.
\bibitem[{Ferdousi et~al.(2021)Ferdousi, Chen, Soltani, Zhu, Cao, Choi,
  Advincula, and Jiang}]{RN13}
\bibinfo{author}{S.~Ferdousi}, \bibinfo{author}{Q.~Chen},
  \bibinfo{author}{M.~Soltani}, \bibinfo{author}{J.~D. Zhu},
  \bibinfo{author}{P.~F. Cao}, \bibinfo{author}{W.~B. Choi},
  \bibinfo{author}{R.~Advincula}, \bibinfo{author}{Y.~J. Jiang},
\newblock \bibinfo{title}{Characterize traction-separation relation and
  interfacial imperfections by data-driven machine learning models},
\newblock \bibinfo{journal}{Scientific Reports} \bibinfo{volume}{11}
  (\bibinfo{year}{2021}). \DOIprefix\doi{ARTN 14330
  10.1038/s41598-021-93852-y}.
\bibitem[{Wei et~al.(2022)Wei, Zhang, Liechti, and Wu}]{wei2022deep}
\bibinfo{author}{C.~Wei}, \bibinfo{author}{J.~Zhang}, \bibinfo{author}{K.~M.
  Liechti}, \bibinfo{author}{C.~Wu},
\newblock \bibinfo{title}{Deep-green inversion to extract traction-separation
  relations at material interfaces},
\newblock \bibinfo{journal}{International Journal of Solids and Structures}
  \bibinfo{volume}{250} (\bibinfo{year}{2022}) \bibinfo{pages}{111698}.
\bibitem[{Niu and Srivastava(2022{\natexlab{a}})}]{RN19}
\bibinfo{author}{S.~Niu}, \bibinfo{author}{V.~Srivastava},
\newblock \bibinfo{title}{Simulation trained cnn for accurate embedded crack
  length, location, and orientation from ultrasound measurements},
\newblock \bibinfo{journal}{International Journal of Solids and Structures}
  \bibinfo{volume}{242} (\bibinfo{year}{2022}{\natexlab{a}}).
  \DOIprefix\doi{10.1016/j.ijsolstr.2022.111521}.
\bibitem[{Niu and Srivastava(2022{\natexlab{b}})}]{RN20}
\bibinfo{author}{S.~Niu}, \bibinfo{author}{V.~Srivastava},
\newblock \bibinfo{title}{Ultrasound classification of interacting flaws using
  finite element simulations and convolutional neural network},
\newblock \bibinfo{journal}{Engineering with Computers} \bibinfo{volume}{38}
  (\bibinfo{year}{2022}{\natexlab{b}}) \bibinfo{pages}{4653--4662}.
\bibitem[{Lew et~al.(2021)Lew, Yu, Hsu, and Buehler}]{RN16}
\bibinfo{author}{A.~J. Lew}, \bibinfo{author}{C.~H. Yu}, \bibinfo{author}{Y.~C.
  Hsu}, \bibinfo{author}{M.~J. Buehler},
\newblock \bibinfo{title}{Deep learning model to predict fracture mechanisms of
  graphene},
\newblock \bibinfo{journal}{Npj 2d Materials and Applications}
  \bibinfo{volume}{5} (\bibinfo{year}{2021}). \DOIprefix\doi{ARTN 48
  10.1038/s41699-021-00228-x}.
\bibitem[{Lew and Buehler(2021)}]{RN24}
\bibinfo{author}{A.~J. Lew}, \bibinfo{author}{M.~J. Buehler},
\newblock \bibinfo{title}{A deep learning augmented genetic algorithm approach
  to polycrystalline 2d material fracture discovery and design},
\newblock \bibinfo{journal}{Applied Physics Reviews} \bibinfo{volume}{8}
  (\bibinfo{year}{2021}). \DOIprefix\doi{Artn 041414 10.1063/5.0057162}.
\bibitem[{Athanasiou et~al.(2023)Athanasiou, Liu, Zhang, Cai, Ramirez, Padture,
  Lou, Sheldon, and Gao}]{RN274}
\bibinfo{author}{C.~E. Athanasiou}, \bibinfo{author}{X.~Liu},
  \bibinfo{author}{B.~Zhang}, \bibinfo{author}{T.~Cai},
  \bibinfo{author}{C.~Ramirez}, \bibinfo{author}{N.~P. Padture},
  \bibinfo{author}{J.~Lou}, \bibinfo{author}{B.~W. Sheldon},
  \bibinfo{author}{H.~Gao},
\newblock \bibinfo{title}{Integrated simulation, machine learning, and
  experimental approach to characterizing fracture instability in indentation
  pillar-splitting of materials},
\newblock \bibinfo{journal}{Journal of the Mechanics and Physics of Solids}
  \bibinfo{volume}{170} (\bibinfo{year}{2023}) \bibinfo{pages}{105092}.
\bibitem[{Komaris et~al.(2019)Komaris, Pérez-Valero, Jordan, Barton, Hennessy,
  O’Flynn, and Tedesco}]{RN168}
\bibinfo{author}{D.-S. Komaris}, \bibinfo{author}{E.~Pérez-Valero},
  \bibinfo{author}{L.~Jordan}, \bibinfo{author}{J.~Barton},
  \bibinfo{author}{L.~Hennessy}, \bibinfo{author}{B.~O’Flynn},
  \bibinfo{author}{S.~Tedesco},
\newblock \bibinfo{title}{Predicting three-dimensional ground reaction forces
  in running by using artificial neural networks and lower body kinematics},
\newblock \bibinfo{journal}{IEEE Access} \bibinfo{volume}{7}
  (\bibinfo{year}{2019}) \bibinfo{pages}{156779--156786}.
\bibitem[{Eerdekens et~al.(2020)Eerdekens, Deruyck, Fontaine, Martens,
  De~Poorter, and Joseph}]{RN167}
\bibinfo{author}{A.~Eerdekens}, \bibinfo{author}{M.~Deruyck},
  \bibinfo{author}{J.~Fontaine}, \bibinfo{author}{L.~Martens},
  \bibinfo{author}{E.~De~Poorter}, \bibinfo{author}{W.~Joseph},
\newblock \bibinfo{title}{Automatic equine activity detection by convolutional
  neural networks using accelerometer data},
\newblock \bibinfo{journal}{Computers and electronics in agriculture}
  \bibinfo{volume}{168} (\bibinfo{year}{2020}) \bibinfo{pages}{105139}.
\bibitem[{Holzapfel et~al.(2021)Holzapfel, Linka, Sherifova, and Cyron}]{RN181}
\bibinfo{author}{G.~A. Holzapfel}, \bibinfo{author}{K.~Linka},
  \bibinfo{author}{S.~Sherifova}, \bibinfo{author}{C.~J. Cyron},
\newblock \bibinfo{title}{Predictive constitutive modelling of arteries by deep
  learning},
\newblock \bibinfo{journal}{Journal of the Royal Society Interface}
  \bibinfo{volume}{18} (\bibinfo{year}{2021}) \bibinfo{pages}{20210411}.
\bibitem[{Liu et~al.(2019)Liu, Liang, and Sun}]{RN183}
\bibinfo{author}{M.~Liu}, \bibinfo{author}{L.~Liang}, \bibinfo{author}{W.~Sun},
\newblock \bibinfo{title}{Estimation of in vivo constitutive parameters of the
  aortic wall using a machine learning approach},
\newblock \bibinfo{journal}{Computer methods in applied mechanics and
  engineering} \bibinfo{volume}{347} (\bibinfo{year}{2019})
  \bibinfo{pages}{201--217}.
\bibitem[{Kamali et~al.(2023)Kamali, Sarabian, and Laksari}]{RN202}
\bibinfo{author}{A.~Kamali}, \bibinfo{author}{M.~Sarabian},
  \bibinfo{author}{K.~Laksari},
\newblock \bibinfo{title}{Elasticity imaging using physics-informed neural
  networks: Spatial discovery of elastic modulus and poisson's ratio},
\newblock \bibinfo{journal}{Acta Biomaterialia} \bibinfo{volume}{155}
  (\bibinfo{year}{2023}) \bibinfo{pages}{400--409}.
\bibitem[{Goswami et~al.(2022)Goswami, Li, Rego, Latorre, Humphrey, and
  Karniadakis}]{RN220}
\bibinfo{author}{S.~Goswami}, \bibinfo{author}{D.~S. Li},
  \bibinfo{author}{B.~V. Rego}, \bibinfo{author}{M.~Latorre},
  \bibinfo{author}{J.~D. Humphrey}, \bibinfo{author}{G.~E. Karniadakis},
\newblock \bibinfo{title}{Neural operator learning of heterogeneous
  mechanobiological insults contributing to aortic aneurysms},
\newblock \bibinfo{journal}{Journal of the Royal Society Interface}
  \bibinfo{volume}{19} (\bibinfo{year}{2022}) \bibinfo{pages}{20220410}.
\bibitem[{Mukherjee et~al.(2022)Mukherjee, Park, Pathak, Patino, Bao, and
  Espinosa}]{RN89}
\bibinfo{author}{P.~Mukherjee}, \bibinfo{author}{S.~H. Park},
  \bibinfo{author}{N.~Pathak}, \bibinfo{author}{C.~A. Patino},
  \bibinfo{author}{G.~Bao}, \bibinfo{author}{H.~D. Espinosa},
\newblock \bibinfo{title}{Integrating micro and nano technologies for cell
  engineering and analysis: Toward the next generation of cell therapy
  workflows},
\newblock \bibinfo{journal}{Acs Nano} \bibinfo{volume}{16}
  (\bibinfo{year}{2022}) \bibinfo{pages}{15653--15680}.
  \DOIprefix\doi{10.1021/acsnano.2c05494}.
\bibitem[{Patino et~al.(2021)Patino, Mukherjee, Lemaitre, Pathak, and
  Espinosa}]{RN246}
\bibinfo{author}{C.~A. Patino}, \bibinfo{author}{P.~Mukherjee},
  \bibinfo{author}{V.~Lemaitre}, \bibinfo{author}{N.~Pathak},
  \bibinfo{author}{H.~D. Espinosa},
\newblock \bibinfo{title}{Deep learning and computer vision strategies for
  automated gene editing with a single-cell electroporation platform},
\newblock \bibinfo{journal}{SLAS TECHNOLOGY: Translating Life Sciences
  Innovation} \bibinfo{volume}{26} (\bibinfo{year}{2021})
  \bibinfo{pages}{26--36}.
\bibitem[{Muliana et~al.(2002)Muliana, Haj-Ali, Steward, and
  Saxena}]{muliana2002artificial}
\bibinfo{author}{A.~Muliana}, \bibinfo{author}{R.~M. Haj-Ali},
  \bibinfo{author}{R.~Steward}, \bibinfo{author}{A.~Saxena},
\newblock \bibinfo{title}{Artificial neural network and finite element modeling
  of nanoindentation tests},
\newblock \bibinfo{journal}{Metallurgical and Materials Transactions A}
  \bibinfo{volume}{33} (\bibinfo{year}{2002}) \bibinfo{pages}{1939--1947}.
\bibitem[{Huber et~al.(2001)Huber, Konstantinidis, and Tsakmakis}]{RN62}
\bibinfo{author}{N.~Huber}, \bibinfo{author}{A.~Konstantinidis},
  \bibinfo{author}{C.~Tsakmakis},
\newblock \bibinfo{title}{Determination of poisson's ratio by spherical
  indentation using neural networks - part i: Theory},
\newblock \bibinfo{journal}{Journal of Applied Mechanics-Transactions of the
  Asme} \bibinfo{volume}{68} (\bibinfo{year}{2001}) \bibinfo{pages}{218--223}.
  \DOIprefix\doi{Doi 10.1115/1.1354624}.
\bibitem[{Huber and Tsakmakis(2001)}]{RN63}
\bibinfo{author}{N.~Huber}, \bibinfo{author}{C.~Tsakmakis},
\newblock \bibinfo{title}{Determination of poisson's ratio by spherical
  indentation using neural networks - part ii: Identification method},
\newblock \bibinfo{journal}{Journal of Applied Mechanics-Transactions of the
  Asme} \bibinfo{volume}{68} (\bibinfo{year}{2001}) \bibinfo{pages}{224--229}.
  \DOIprefix\doi{Doi 10.1115/1.1355032}.
\bibitem[{Zhang and Needleman(2021)}]{RN84}
\bibinfo{author}{Y.~Zhang}, \bibinfo{author}{A.~Needleman},
\newblock \bibinfo{title}{Characterization of plastically compressible solids
  via spherical indentation},
\newblock \bibinfo{journal}{Journal of the Mechanics and Physics of Solids}
  \bibinfo{volume}{148} (\bibinfo{year}{2021}).
  \DOIprefix\doi{10.1016/j.jmps.2020.104283}.
\bibitem[{Fernandez-Zelaia et~al.(2018)Fernandez-Zelaia, Joseph, Kalidindi, and
  Melkote}]{RN81}
\bibinfo{author}{P.~Fernandez-Zelaia}, \bibinfo{author}{V.~R. Joseph},
  \bibinfo{author}{S.~R. Kalidindi}, \bibinfo{author}{S.~N. Melkote},
\newblock \bibinfo{title}{Estimating mechanical properties from spherical
  indentation using bayesian approaches},
\newblock \bibinfo{journal}{Materials \& Design} \bibinfo{volume}{147}
  (\bibinfo{year}{2018}) \bibinfo{pages}{92--105}.
  \DOIprefix\doi{10.1016/j.matdes.2018.03.037}.
\bibitem[{Chandrashekar et~al.(2022)Chandrashekar, Belardinelli, Bessa,
  Staufer, and Alijani}]{RN91}
\bibinfo{author}{A.~Chandrashekar}, \bibinfo{author}{P.~Belardinelli},
  \bibinfo{author}{M.~A. Bessa}, \bibinfo{author}{U.~Staufer},
  \bibinfo{author}{F.~Alijani},
\newblock \bibinfo{title}{Quantifying nanoscale forces using machine learning
  in dynamic atomic force microscopy},
\newblock \bibinfo{journal}{Nanoscale Advances} \bibinfo{volume}{4}
  (\bibinfo{year}{2022}) \bibinfo{pages}{2134--2143}.
  \DOIprefix\doi{10.1039/d2na00011c}.
\bibitem[{Herriott and Spear(2020)}]{RN229}
\bibinfo{author}{C.~Herriott}, \bibinfo{author}{A.~D. Spear},
\newblock \bibinfo{title}{Predicting microstructure-dependent mechanical
  properties in additively manufactured metals with machine-and deep-learning
  methods},
\newblock \bibinfo{journal}{Computational Materials Science}
  \bibinfo{volume}{175} (\bibinfo{year}{2020}) \bibinfo{pages}{109599}.
\bibitem[{Sepasdar et~al.(2022)Sepasdar, Karpatne, and Shakiba}]{RN243}
\bibinfo{author}{R.~Sepasdar}, \bibinfo{author}{A.~Karpatne},
  \bibinfo{author}{M.~Shakiba},
\newblock \bibinfo{title}{A data-driven approach to full-field nonlinear stress
  distribution and failure pattern prediction in composites using deep
  learning},
\newblock \bibinfo{journal}{Computer Methods in Applied Mechanics and
  Engineering} \bibinfo{volume}{397} (\bibinfo{year}{2022})
  \bibinfo{pages}{115126}.
\bibitem[{Bulgarevich et~al.(2018)Bulgarevich, Tsukamoto, Kasuya, Demura, and
  Watanabe}]{RN231}
\bibinfo{author}{D.~S. Bulgarevich}, \bibinfo{author}{S.~Tsukamoto},
  \bibinfo{author}{T.~Kasuya}, \bibinfo{author}{M.~Demura},
  \bibinfo{author}{M.~Watanabe},
\newblock \bibinfo{title}{Pattern recognition with machine learning on optical
  microscopy images of typical metallurgical microstructures},
\newblock \bibinfo{journal}{Scientific reports} \bibinfo{volume}{8}
  (\bibinfo{year}{2018}) \bibinfo{pages}{1--8}.
\bibitem[{Alderete et~al.(2022)Alderete, Pathak, and Espinosa}]{RN38}
\bibinfo{author}{N.~A. Alderete}, \bibinfo{author}{N.~Pathak},
  \bibinfo{author}{H.~D. Espinosa},
\newblock \bibinfo{title}{Machine learning assisted design of
  shape-programmable 3d kirigami metamaterials},
\newblock \bibinfo{journal}{Npj Computational Materials} \bibinfo{volume}{8}
  (\bibinfo{year}{2022}). \DOIprefix\doi{10.1038/s41524-022-00873-w}.
\bibitem[{Ma et~al.(2022)Ma, Chang, Wu, and Zhao}]{RN40}
\bibinfo{author}{C.~Ma}, \bibinfo{author}{Y.~Chang}, \bibinfo{author}{S.~Wu},
  \bibinfo{author}{R.~R. Zhao},
\newblock \bibinfo{title}{Deep learning-accelerated designs of tunable
  magneto-mechanical metamaterials},
\newblock \bibinfo{journal}{Acs Applied Materials \& Interfaces}
  \bibinfo{volume}{14} (\bibinfo{year}{2022}) \bibinfo{pages}{33892--33902}.
  \DOIprefix\doi{10.1021/acsami.2c0905233892}.
\bibitem[{Hsu et~al.(2021)Hsu, Epting, Kim, Abernathy, Hackett, Rollett,
  Salvador, and Holm}]{RN244}
\bibinfo{author}{T.~Hsu}, \bibinfo{author}{W.~K. Epting},
  \bibinfo{author}{H.~Kim}, \bibinfo{author}{H.~W. Abernathy},
  \bibinfo{author}{G.~A. Hackett}, \bibinfo{author}{A.~D. Rollett},
  \bibinfo{author}{P.~A. Salvador}, \bibinfo{author}{E.~A. Holm},
\newblock \bibinfo{title}{Microstructure generation via generative adversarial
  network for heterogeneous, topologically complex 3d materials},
\newblock \bibinfo{journal}{Jom} \bibinfo{volume}{73} (\bibinfo{year}{2021})
  \bibinfo{pages}{90--102}.
\bibitem[{Zhang et~al.(2021)Zhang, Nguyen, Paci, Sankaranarayanan,
  Mendoza-Cortes, and Espinosa}]{zhang2021multi}
\bibinfo{author}{X.~Zhang}, \bibinfo{author}{H.~Nguyen}, \bibinfo{author}{J.~T.
  Paci}, \bibinfo{author}{S.~K. Sankaranarayanan}, \bibinfo{author}{J.~L.
  Mendoza-Cortes}, \bibinfo{author}{H.~D. Espinosa},
\newblock \bibinfo{title}{Multi-objective parametrization of interatomic
  potentials for large deformation pathways and fracture of two-dimensional
  materials},
\newblock \bibinfo{journal}{npj Computational Materials} \bibinfo{volume}{7}
  (\bibinfo{year}{2021}) \bibinfo{pages}{113}.
\bibitem[{Zhang et~al.(2022)Zhang, Nguyen, Zhang, Ajayan, Wen, and
  Espinosa}]{zhang2022atomistic}
\bibinfo{author}{X.~Zhang}, \bibinfo{author}{H.~Nguyen},
  \bibinfo{author}{X.~Zhang}, \bibinfo{author}{P.~M. Ajayan},
  \bibinfo{author}{J.~Wen}, \bibinfo{author}{H.~D. Espinosa},
\newblock \bibinfo{title}{Atomistic measurement and modeling of intrinsic
  fracture toughness of two-dimensional materials},
\newblock \bibinfo{journal}{Proceedings of the National Academy of Sciences}
  \bibinfo{volume}{119} (\bibinfo{year}{2022}) \bibinfo{pages}{e2206756119}.
\bibitem[{Zhu and Joyce(2012)}]{RN8}
\bibinfo{author}{X.~K. Zhu}, \bibinfo{author}{J.~A. Joyce},
\newblock \bibinfo{title}{Review of fracture toughness (g, k, j, ctod, ctoa)
  testing and standardization},
\newblock \bibinfo{journal}{Engineering Fracture Mechanics}
  \bibinfo{volume}{85} (\bibinfo{year}{2012}) \bibinfo{pages}{1--46}.
  \DOIprefix\doi{10.1016/j.engfracmech.2012.02.001}.
\bibitem[{Zhang et~al.(2018)Zhang, Bai, Morelle, and Suo}]{RN272}
\bibinfo{author}{E.~Zhang}, \bibinfo{author}{R.~Bai}, \bibinfo{author}{X.~P.
  Morelle}, \bibinfo{author}{Z.~Suo},
\newblock \bibinfo{title}{Fatigue fracture of nearly elastic hydrogels},
\newblock \bibinfo{journal}{Soft matter} \bibinfo{volume}{14}
  (\bibinfo{year}{2018}) \bibinfo{pages}{3563--3571}.
\bibitem[{Sun et~al.(2012)Sun, Zhao, Illeperuma, Chaudhuri, Oh, Mooney,
  Vlassak, and Suo}]{RN273}
\bibinfo{author}{J.-Y. Sun}, \bibinfo{author}{X.~Zhao}, \bibinfo{author}{W.~R.
  Illeperuma}, \bibinfo{author}{O.~Chaudhuri}, \bibinfo{author}{K.~H. Oh},
  \bibinfo{author}{D.~J. Mooney}, \bibinfo{author}{J.~J. Vlassak},
  \bibinfo{author}{Z.~Suo},
\newblock \bibinfo{title}{Highly stretchable and tough hydrogels},
\newblock \bibinfo{journal}{Nature} \bibinfo{volume}{489}
  (\bibinfo{year}{2012}) \bibinfo{pages}{133--136}.
\bibitem[{Liu(2020)}]{RN22}
\bibinfo{author}{Z.~Liu},
\newblock \bibinfo{title}{Deep material network with cohesive layers:
  Multi-stage training and interfacial failure analysis},
\newblock \bibinfo{journal}{Computer Methods in Applied Mechanics and
  Engineering} \bibinfo{volume}{363} (\bibinfo{year}{2020}).
  \DOIprefix\doi{10.1016/j.cma.2020.112913}.
\bibitem[{Wang and Sun(2019)}]{RN23}
\bibinfo{author}{K.~Wang}, \bibinfo{author}{W.~C. Sun},
\newblock \bibinfo{title}{Meta-modeling game for deriving theory-consistent,
  microstructure-based traction-separation laws via deep reinforcement
  learning},
\newblock \bibinfo{journal}{Computer Methods in Applied Mechanics and
  Engineering} \bibinfo{volume}{346} (\bibinfo{year}{2019})
  \bibinfo{pages}{216--241}. \DOIprefix\doi{10.1016/j.cma.2018.11.026}.
\bibitem[{Hsu et~al.(2020)Hsu, Yu, and Buehler}]{RN15}
\bibinfo{author}{Y.~C. Hsu}, \bibinfo{author}{C.~H. Yu}, \bibinfo{author}{M.~J.
  Buehler},
\newblock \bibinfo{title}{Using deep learning to predict fracture patterns in
  crystalline solids},
\newblock \bibinfo{journal}{Matter} \bibinfo{volume}{3} (\bibinfo{year}{2020})
  \bibinfo{pages}{197--211}. \DOIprefix\doi{10.1016/j.matt.2020.04.019}.
\bibitem[{Worthington and Chew(2023)}]{worthington2023crack}
\bibinfo{author}{M.~Worthington}, \bibinfo{author}{H.~B. Chew},
\newblock \bibinfo{title}{Crack path predictions in heterogeneous media by
  machine learning},
\newblock \bibinfo{journal}{Journal of the Mechanics and Physics of Solids}
  \bibinfo{volume}{172} (\bibinfo{year}{2023}) \bibinfo{pages}{105188}.
\bibitem[{Chon et~al.(2017)Chon, Daly, Wang, Xiao, Zaheri, Meyers, and
  Espinosa}]{chon2017lamellae}
\bibinfo{author}{M.~J. Chon}, \bibinfo{author}{M.~Daly},
  \bibinfo{author}{B.~Wang}, \bibinfo{author}{X.~Xiao},
  \bibinfo{author}{A.~Zaheri}, \bibinfo{author}{M.~A. Meyers},
  \bibinfo{author}{H.~D. Espinosa},
\newblock \bibinfo{title}{Lamellae spatial distribution modulates fracture
  behavior and toughness of african pangolin scales},
\newblock \bibinfo{journal}{Journal of the mechanical behavior of biomedical
  materials} \bibinfo{volume}{76} (\bibinfo{year}{2017})
  \bibinfo{pages}{30--37}.
\bibitem[{Garc{\'\i}a-Moreno et~al.(2019)Garc{\'\i}a-Moreno, Kamm, Neu,
  B{\"u}lk, Mokso, Schlep{\"u}tz, Stampanoni, and Banhart}]{garcia2019using}
\bibinfo{author}{F.~Garc{\'\i}a-Moreno}, \bibinfo{author}{P.~H. Kamm},
  \bibinfo{author}{T.~R. Neu}, \bibinfo{author}{F.~B{\"u}lk},
  \bibinfo{author}{R.~Mokso}, \bibinfo{author}{C.~M. Schlep{\"u}tz},
  \bibinfo{author}{M.~Stampanoni}, \bibinfo{author}{J.~Banhart},
\newblock \bibinfo{title}{Using x-ray tomoscopy to explore the dynamics of
  foaming metal},
\newblock \bibinfo{journal}{Nature communications} \bibinfo{volume}{10}
  (\bibinfo{year}{2019}) \bibinfo{pages}{3762}.
\bibitem[{Espinosa et~al.(2012)Espinosa, Bernal, and Minary-Jolandan}]{RN54}
\bibinfo{author}{H.~D. Espinosa}, \bibinfo{author}{R.~A. Bernal},
  \bibinfo{author}{M.~Minary-Jolandan},
\newblock \bibinfo{title}{A review of mechanical and electromechanical
  properties of piezoelectric nanowires},
\newblock \bibinfo{journal}{Advanced Materials} \bibinfo{volume}{24}
  (\bibinfo{year}{2012}) \bibinfo{pages}{4656--4675}.
  \DOIprefix\doi{10.1002/adma.201104810}.
\bibitem[{Ramachandramoorthy et~al.(2016)Ramachandramoorthy, Gao, Bernal, and
  Espinosa}]{RN261}
\bibinfo{author}{R.~Ramachandramoorthy}, \bibinfo{author}{W.~Gao},
  \bibinfo{author}{R.~Bernal}, \bibinfo{author}{H.~Espinosa},
\newblock \bibinfo{title}{High strain rate tensile testing of silver nanowires:
  rate-dependent brittle-to-ductile transition},
\newblock \bibinfo{journal}{Nano letters} \bibinfo{volume}{16}
  (\bibinfo{year}{2016}) \bibinfo{pages}{255--263}.
\bibitem[{Ramachandramoorthy et~al.(2015)Ramachandramoorthy, Bernal, and
  Espinosa}]{ramachandramoorthy2015pushing}
\bibinfo{author}{R.~Ramachandramoorthy}, \bibinfo{author}{R.~Bernal},
  \bibinfo{author}{H.~D. Espinosa},
\newblock \bibinfo{title}{Pushing the envelope of in situ transmission electron
  microscopy},
\newblock \bibinfo{journal}{ACS nano} \bibinfo{volume}{9}
  (\bibinfo{year}{2015}) \bibinfo{pages}{4675--4685}.
\bibitem[{Sharma et~al.(2022)Sharma, Mukhopadhyay, and Kushvaha}]{RN26}
\bibinfo{author}{A.~Sharma}, \bibinfo{author}{T.~Mukhopadhyay},
  \bibinfo{author}{V.~Kushvaha},
\newblock \bibinfo{title}{Experimental data-driven uncertainty quantification
  for the dynamic fracture toughness of particulate polymer composites},
\newblock \bibinfo{journal}{Engineering Fracture Mechanics}
  \bibinfo{volume}{273} (\bibinfo{year}{2022}). \DOIprefix\doi{ARTN 108724
  10.1016/j.engfracmech.2022.108724}.
\bibitem[{Niu et~al.(2023)Niu, Zhang, Bazilevs, and Srivastava}]{RN27}
\bibinfo{author}{S.~Niu}, \bibinfo{author}{E.~Zhang},
  \bibinfo{author}{Y.~Bazilevs}, \bibinfo{author}{V.~Srivastava},
\newblock \bibinfo{title}{Modeling finite-strain plasticity using
  physics-informed neural network and assessment of the network performance},
\newblock \bibinfo{journal}{Journal of the Mechanics and Physics of Solids}
  \bibinfo{volume}{172} (\bibinfo{year}{2023}) \bibinfo{pages}{105177}.
\bibitem[{Xu et~al.(2022)Xu, Ye, Li, Zhao, and Feng}]{RN28}
\bibinfo{author}{B.-W. Xu}, \bibinfo{author}{S.~Ye}, \bibinfo{author}{M.~Li},
  \bibinfo{author}{H.-P. Zhao}, \bibinfo{author}{X.-Q. Feng},
\newblock \bibinfo{title}{Deep learning method for predicting the strengths of
  microcracked brittle materials},
\newblock \bibinfo{journal}{Engineering Fracture Mechanics}
  \bibinfo{volume}{271} (\bibinfo{year}{2022}).
  \DOIprefix\doi{10.1016/j.engfracmech.2022.108600}.
\bibitem[{Knudson and Knudson(2007)}]{RN166}
\bibinfo{author}{D.~V. Knudson}, \bibinfo{author}{D.~Knudson},
  \bibinfo{title}{Fundamentals of biomechanics}, volume \bibinfo{volume}{183},
  \bibinfo{publisher}{Springer}, \bibinfo{year}{2007}.
\bibitem[{Zhang et~al.(2022)Zhang, Li, Zhang, Shahabi, Xia, Deng, and
  Alshurafa}]{zhang2022deep}
\bibinfo{author}{S.~Zhang}, \bibinfo{author}{Y.~Li},
  \bibinfo{author}{S.~Zhang}, \bibinfo{author}{F.~Shahabi},
  \bibinfo{author}{S.~Xia}, \bibinfo{author}{Y.~Deng},
  \bibinfo{author}{N.~Alshurafa},
\newblock \bibinfo{title}{Deep learning in human activity recognition with
  wearable sensors: A review on advances},
\newblock \bibinfo{journal}{Sensors} \bibinfo{volume}{22}
  (\bibinfo{year}{2022}) \bibinfo{pages}{1476}.
\bibitem[{Halilaj et~al.(2018)Halilaj, Rajagopal, Fiterau, Hicks, Hastie, and
  Delp}]{RN169}
\bibinfo{author}{E.~Halilaj}, \bibinfo{author}{A.~Rajagopal},
  \bibinfo{author}{M.~Fiterau}, \bibinfo{author}{J.~L. Hicks},
  \bibinfo{author}{T.~J. Hastie}, \bibinfo{author}{S.~L. Delp},
\newblock \bibinfo{title}{Machine learning in human movement biomechanics: Best
  practices, common pitfalls, and new opportunities},
\newblock \bibinfo{journal}{Journal of biomechanics} \bibinfo{volume}{81}
  (\bibinfo{year}{2018}) \bibinfo{pages}{1--11}.
\bibitem[{Phellan et~al.(2021)Phellan, Hachem, Clin, Mac‐Thiong, and
  Duong}]{RN170}
\bibinfo{author}{R.~Phellan}, \bibinfo{author}{B.~Hachem},
  \bibinfo{author}{J.~Clin}, \bibinfo{author}{J.~Mac‐Thiong},
  \bibinfo{author}{L.~Duong},
\newblock \bibinfo{title}{Real‐time biomechanics using the finite element
  method and machine learning: Review and perspective},
\newblock \bibinfo{journal}{Medical Physics} \bibinfo{volume}{48}
  (\bibinfo{year}{2021}) \bibinfo{pages}{7--18}.
\bibitem[{Low et~al.(2022)Low, Chan, Chuah, Tee, Hum, Salim, and Lai}]{RN171}
\bibinfo{author}{W.~S. Low}, \bibinfo{author}{C.~K. Chan},
  \bibinfo{author}{J.~H. Chuah}, \bibinfo{author}{Y.~K. Tee},
  \bibinfo{author}{Y.~C. Hum}, \bibinfo{author}{M.~I.~M. Salim},
  \bibinfo{author}{K.~W. Lai},
\newblock \bibinfo{title}{A review of machine learning network in human motion
  biomechanics},
\newblock \bibinfo{journal}{Journal of Grid Computing} \bibinfo{volume}{20}
  (\bibinfo{year}{2022}) \bibinfo{pages}{4}.
\bibitem[{Mouloodi et~al.(2021)Mouloodi, Rahmanpanah, Gohari, Burvill, Tse, and
  Davies}]{RN172}
\bibinfo{author}{S.~Mouloodi}, \bibinfo{author}{H.~Rahmanpanah},
  \bibinfo{author}{S.~Gohari}, \bibinfo{author}{C.~Burvill},
  \bibinfo{author}{K.~M. Tse}, \bibinfo{author}{H.~M. Davies},
\newblock \bibinfo{title}{What can artificial intelligence and machine learning
  tell us? a review of applications to equine biomechanical research},
\newblock \bibinfo{journal}{Journal of the Mechanical Behavior of Biomedical
  Materials} \bibinfo{volume}{123} (\bibinfo{year}{2021})
  \bibinfo{pages}{104728}.
\bibitem[{Gerbin et~al.(2021)Gerbin, Grancharova, Donovan-Maiye, Hendershott,
  Anderson, Brown, Chen, Dinh, Gehring, Johnson et~al.}]{gerbin2021cell}
\bibinfo{author}{K.~A. Gerbin}, \bibinfo{author}{T.~Grancharova},
  \bibinfo{author}{R.~M. Donovan-Maiye}, \bibinfo{author}{M.~C. Hendershott},
  \bibinfo{author}{H.~G. Anderson}, \bibinfo{author}{J.~M. Brown},
  \bibinfo{author}{J.~Chen}, \bibinfo{author}{S.~Q. Dinh},
  \bibinfo{author}{J.~L. Gehring}, \bibinfo{author}{G.~R. Johnson}, et~al.,
\newblock \bibinfo{title}{Cell states beyond transcriptomics: integrating
  structural organization and gene expression in hipsc-derived cardiomyocytes},
\newblock \bibinfo{journal}{Cell Systems} \bibinfo{volume}{12}
  (\bibinfo{year}{2021}) \bibinfo{pages}{670--687}.
\bibitem[{Jin et~al.(2021)Jin, Landauer, and Kim}]{RN173}
\bibinfo{author}{H.~Jin}, \bibinfo{author}{A.~K. Landauer},
  \bibinfo{author}{K.-S. Kim},
\newblock \bibinfo{title}{Ruga mechanics of soft-orifice closure under external
  pressure},
\newblock \bibinfo{journal}{Proceedings of the Royal Society A}
  \bibinfo{volume}{477} (\bibinfo{year}{2021}) \bibinfo{pages}{20210238}.
\bibitem[{Diab et~al.(2013)Diab, Zhang, Zhao, Gao, and Kim}]{RN174}
\bibinfo{author}{M.~Diab}, \bibinfo{author}{T.~Zhang},
  \bibinfo{author}{R.~Zhao}, \bibinfo{author}{H.~Gao}, \bibinfo{author}{K.-S.
  Kim},
\newblock \bibinfo{title}{Ruga mechanics of creasing: from instantaneous to
  setback creases},
\newblock \bibinfo{journal}{Proceedings of the Royal Society A: Mathematical,
  Physical and Engineering Sciences} \bibinfo{volume}{469}
  (\bibinfo{year}{2013}) \bibinfo{pages}{20120753}.
\bibitem[{Zhao et~al.(2015)Zhao, Zhang, Diab, Gao, and Kim}]{RN175}
\bibinfo{author}{R.~Zhao}, \bibinfo{author}{T.~Zhang},
  \bibinfo{author}{M.~Diab}, \bibinfo{author}{H.~Gao}, \bibinfo{author}{K.-S.
  Kim},
\newblock \bibinfo{title}{The primary bilayer ruga-phase diagram i:
  Localizations in ruga evolution},
\newblock \bibinfo{journal}{Extreme Mechanics Letters} \bibinfo{volume}{4}
  (\bibinfo{year}{2015}) \bibinfo{pages}{76--82}.
\bibitem[{Zhao et~al.(2016)Zhao, Diab, and Kim}]{RN176}
\bibinfo{author}{R.~Zhao}, \bibinfo{author}{M.~Diab}, \bibinfo{author}{K.-S.
  Kim},
\newblock \bibinfo{title}{The primary bilayer ruga-phase diagram ii:
  Irreversibility in ruga evolution},
\newblock \bibinfo{journal}{Journal of Applied Mechanics} \bibinfo{volume}{83}
  (\bibinfo{year}{2016}) \bibinfo{pages}{091004}.
\bibitem[{Treloar(1948)}]{RN177}
\bibinfo{author}{L.~Treloar},
\newblock \bibinfo{title}{Stresses and birefringence in rubber subjected to
  general homogeneous strain},
\newblock \bibinfo{journal}{Proceedings of the Physical Society}
  \bibinfo{volume}{60} (\bibinfo{year}{1948}) \bibinfo{pages}{135}.
\bibitem[{Ogden(1972)}]{RN178}
\bibinfo{author}{R.~W. Ogden},
\newblock \bibinfo{title}{Large deformation isotropic elasticity–on the
  correlation of theory and experiment for incompressible rubberlike solids},
\newblock \bibinfo{journal}{Proceedings of the Royal Society of London. A.
  Mathematical and Physical Sciences} \bibinfo{volume}{326}
  (\bibinfo{year}{1972}) \bibinfo{pages}{565--584}.
\bibitem[{Fung(2013)}]{RN179}
\bibinfo{author}{Y.-C. Fung}, \bibinfo{title}{Biomechanics: motion, flow,
  stress, and growth}, \bibinfo{publisher}{Springer Science \& Business Media},
  \bibinfo{year}{2013}.
\bibitem[{Holzapfel et~al.(2000)Holzapfel, Gasser, and Ogden}]{RN180}
\bibinfo{author}{G.~A. Holzapfel}, \bibinfo{author}{T.~C. Gasser},
  \bibinfo{author}{R.~W. Ogden},
\newblock \bibinfo{title}{A new constitutive framework for arterial wall
  mechanics and a comparative study of material models},
\newblock \bibinfo{journal}{Journal of elasticity and the physical science of
  solids} \bibinfo{volume}{61} (\bibinfo{year}{2000}) \bibinfo{pages}{1--48}.
\bibitem[{Shi et~al.(2019)Shi, Yao, Gan, Zhao, Eugene~McKee, Vink, Wapner,
  Hendon, and Myers}]{RN198}
\bibinfo{author}{L.~Shi}, \bibinfo{author}{W.~Yao}, \bibinfo{author}{Y.~Gan},
  \bibinfo{author}{L.~Y. Zhao}, \bibinfo{author}{W.~Eugene~McKee},
  \bibinfo{author}{J.~Vink}, \bibinfo{author}{R.~J. Wapner},
  \bibinfo{author}{C.~P. Hendon}, \bibinfo{author}{K.~Myers},
\newblock \bibinfo{title}{Anisotropic material characterization of human cervix
  tissue based on indentation and inverse finite element analysis},
\newblock \bibinfo{journal}{Journal of biomechanical engineering}
  \bibinfo{volume}{141} (\bibinfo{year}{2019}).
\bibitem[{Kakaletsis et~al.(2021)Kakaletsis, Meador, Mathur, Sugerman, Jazwiec,
  Malinowski, Lejeune, Timek, and Rausch}]{RN199}
\bibinfo{author}{S.~Kakaletsis}, \bibinfo{author}{W.~D. Meador},
  \bibinfo{author}{M.~Mathur}, \bibinfo{author}{G.~P. Sugerman},
  \bibinfo{author}{T.~Jazwiec}, \bibinfo{author}{M.~Malinowski},
  \bibinfo{author}{E.~Lejeune}, \bibinfo{author}{T.~A. Timek},
  \bibinfo{author}{M.~K. Rausch},
\newblock \bibinfo{title}{Right ventricular myocardial mechanics: Multi-modal
  deformation, microstructure, modeling, and comparison to the left ventricle},
\newblock \bibinfo{journal}{Acta biomaterialia} \bibinfo{volume}{123}
  (\bibinfo{year}{2021}) \bibinfo{pages}{154--166}.
\bibitem[{Sugerman et~al.(2021)Sugerman, Kakaletsis, Thakkar, Chokshi, Parekh,
  and Rausch}]{RN200}
\bibinfo{author}{G.~P. Sugerman}, \bibinfo{author}{S.~Kakaletsis},
  \bibinfo{author}{P.~Thakkar}, \bibinfo{author}{A.~Chokshi},
  \bibinfo{author}{S.~H. Parekh}, \bibinfo{author}{M.~K. Rausch},
\newblock \bibinfo{title}{A whole blood thrombus mimic: constitutive behavior
  under simple shear},
\newblock \bibinfo{journal}{journal of the mechanical behavior of biomedical
  materials} \bibinfo{volume}{115} (\bibinfo{year}{2021})
  \bibinfo{pages}{104216}.
\bibitem[{Kakaletsis et~al.(2022)Kakaletsis, Lejeune, and Rausch}]{RN201}
\bibinfo{author}{S.~Kakaletsis}, \bibinfo{author}{E.~Lejeune},
  \bibinfo{author}{M.~K. Rausch},
\newblock \bibinfo{title}{Can machine learning accelerate soft material
  parameter identification from complex mechanical test data?},
\newblock \bibinfo{journal}{Biomechanics and Modeling in Mechanobiology}
  (\bibinfo{year}{2022}) \bibinfo{pages}{1--14}.
\bibitem[{Holzapfel et~al.(2015)Holzapfel, Niestrawska, Ogden, Reinisch, and
  Schriefl}]{RN182}
\bibinfo{author}{G.~A. Holzapfel}, \bibinfo{author}{J.~A. Niestrawska},
  \bibinfo{author}{R.~W. Ogden}, \bibinfo{author}{A.~J. Reinisch},
  \bibinfo{author}{A.~J. Schriefl},
\newblock \bibinfo{title}{Modelling non-symmetric collagen fibre dispersion in
  arterial walls},
\newblock \bibinfo{journal}{Journal of the royal society interface}
  \bibinfo{volume}{12} (\bibinfo{year}{2015}) \bibinfo{pages}{20150188}.
\bibitem[{Kirchdoerfer and Ortiz(2016)}]{RN204}
\bibinfo{author}{T.~Kirchdoerfer}, \bibinfo{author}{M.~Ortiz},
\newblock \bibinfo{title}{Data-driven computational mechanics},
\newblock \bibinfo{journal}{Computer Methods in Applied Mechanics and
  Engineering} \bibinfo{volume}{304} (\bibinfo{year}{2016})
  \bibinfo{pages}{81--101}.
\bibitem[{Eggersmann et~al.(2019)Eggersmann, Kirchdoerfer, Reese, Stainier, and
  Ortiz}]{RN205}
\bibinfo{author}{R.~Eggersmann}, \bibinfo{author}{T.~Kirchdoerfer},
  \bibinfo{author}{S.~Reese}, \bibinfo{author}{L.~Stainier},
  \bibinfo{author}{M.~Ortiz},
\newblock \bibinfo{title}{Model-free data-driven inelasticity},
\newblock \bibinfo{journal}{Computer Methods in Applied Mechanics and
  Engineering} \bibinfo{volume}{350} (\bibinfo{year}{2019})
  \bibinfo{pages}{81--99}.
\bibitem[{Stainier et~al.(2019)Stainier, Leygue, and Ortiz}]{RN206}
\bibinfo{author}{L.~Stainier}, \bibinfo{author}{A.~Leygue},
  \bibinfo{author}{M.~Ortiz},
\newblock \bibinfo{title}{Model-free data-driven methods in mechanics: material
  data identification and solvers},
\newblock \bibinfo{journal}{Computational Mechanics} \bibinfo{volume}{64}
  (\bibinfo{year}{2019}) \bibinfo{pages}{381--393}.
\bibitem[{Prume et~al.(2023)Prume, Reese, and Ortiz}]{RN207}
\bibinfo{author}{E.~Prume}, \bibinfo{author}{S.~Reese},
  \bibinfo{author}{M.~Ortiz},
\newblock \bibinfo{title}{Model-free data-driven inference in computational
  mechanics},
\newblock \bibinfo{journal}{Computer Methods in Applied Mechanics and
  Engineering} \bibinfo{volume}{403} (\bibinfo{year}{2023})
  \bibinfo{pages}{115704}.
\bibitem[{Tac et~al.(2022)Tac, Sree, Rausch, and Tepole}]{RN213}
\bibinfo{author}{V.~Tac}, \bibinfo{author}{V.~D. Sree}, \bibinfo{author}{M.~K.
  Rausch}, \bibinfo{author}{A.~B. Tepole},
\newblock \bibinfo{title}{Data-driven modeling of the mechanical behavior of
  anisotropic soft biological tissue},
\newblock \bibinfo{journal}{Engineering with Computers} \bibinfo{volume}{38}
  (\bibinfo{year}{2022}) \bibinfo{pages}{4167--4182}.
\bibitem[{He et~al.(2021)He, Laurence, Lee, and Chen}]{RN216}
\bibinfo{author}{Q.~He}, \bibinfo{author}{D.~W. Laurence},
  \bibinfo{author}{C.-H. Lee}, \bibinfo{author}{J.-S. Chen},
\newblock \bibinfo{title}{Manifold learning based data-driven modeling for soft
  biological tissues},
\newblock \bibinfo{journal}{Journal of biomechanics} \bibinfo{volume}{117}
  (\bibinfo{year}{2021}) \bibinfo{pages}{110124}.
\bibitem[{Li and Chen(2022)}]{RN210}
\bibinfo{author}{L.~Li}, \bibinfo{author}{C.~Chen},
\newblock \bibinfo{title}{Equilibrium-based convolution neural networks for
  constitutive modeling of hyperelastic materials},
\newblock \bibinfo{journal}{Journal of the Mechanics and Physics of Solids}
  \bibinfo{volume}{164} (\bibinfo{year}{2022}) \bibinfo{pages}{104931}.
\bibitem[{Wang et~al.(2021)Wang, Li, Cui, Hui, Yeo, and Zehnder}]{RN211}
\bibinfo{author}{J.~Wang}, \bibinfo{author}{T.~Li}, \bibinfo{author}{F.~Cui},
  \bibinfo{author}{C.-Y. Hui}, \bibinfo{author}{J.~Yeo}, \bibinfo{author}{A.~T.
  Zehnder},
\newblock \bibinfo{title}{Metamodeling of constitutive model using gaussian
  process machine learning},
\newblock \bibinfo{journal}{Journal of the Mechanics and Physics of Solids}
  \bibinfo{volume}{154} (\bibinfo{year}{2021}) \bibinfo{pages}{104532}.
\bibitem[{Liu et~al.(2020)Liu, Liang, and Sun}]{RN212}
\bibinfo{author}{M.~Liu}, \bibinfo{author}{L.~Liang}, \bibinfo{author}{W.~Sun},
\newblock \bibinfo{title}{A generic physics-informed neural network-based
  constitutive model for soft biological tissues},
\newblock \bibinfo{journal}{Computer methods in applied mechanics and
  engineering} \bibinfo{volume}{372} (\bibinfo{year}{2020})
  \bibinfo{pages}{113402}.
\bibitem[{Rao et~al.(2020)Rao, Jin, Engwall, Chason, and Kim}]{RN52}
\bibinfo{author}{Z.~Rao}, \bibinfo{author}{H.~Jin},
  \bibinfo{author}{A.~Engwall}, \bibinfo{author}{E.~Chason},
  \bibinfo{author}{K.-S. Kim},
\newblock \bibinfo{title}{Determination of stresses in incrementally deposited
  films from wafer-curvature measurements},
\newblock \bibinfo{journal}{Journal of Applied Mechanics} \bibinfo{volume}{87}
  (\bibinfo{year}{2020}). \DOIprefix\doi{10.1115/1.4047572}.
\bibitem[{Chason et~al.(2002)Chason, Sheldon, Freund, Floro, and Hearne}]{RN53}
\bibinfo{author}{E.~Chason}, \bibinfo{author}{B.~W. Sheldon},
  \bibinfo{author}{L.~B. Freund}, \bibinfo{author}{J.~A. Floro},
  \bibinfo{author}{S.~J. Hearne},
\newblock \bibinfo{title}{Origin of compressive residual stress in
  polycrystalline thin films},
\newblock \bibinfo{journal}{Physical Review Letters} \bibinfo{volume}{88}
  (\bibinfo{year}{2002}). \DOIprefix\doi{ARTN 156103
  10.1103/PhysRevLett.88.156103}.
\bibitem[{Espinosa et~al.(2002)Espinosa, Peng, Kim, Prorok, Moldovan, Xiao,
  Gerbi, Birrell, Auciello, and Carlisle}]{RN251}
\bibinfo{author}{H.~Espinosa}, \bibinfo{author}{B.~Peng},
  \bibinfo{author}{K.-H. Kim}, \bibinfo{author}{B.~Prorok},
  \bibinfo{author}{N.~Moldovan}, \bibinfo{author}{X.~Xiao},
  \bibinfo{author}{J.~Gerbi}, \bibinfo{author}{J.~Birrell},
  \bibinfo{author}{O.~Auciello}, \bibinfo{author}{J.~Carlisle},
\newblock \bibinfo{title}{Mechanical properties of ultrananocrystalline diamond
  thin films for mems applications},
\newblock \bibinfo{journal}{MRS Online Proceedings Library (OPL)}
  \bibinfo{volume}{741} (\bibinfo{year}{2002}).
\bibitem[{Espinosa et~al.(2003{\natexlab{a}})Espinosa, Prorok, and
  Fischer}]{RN247}
\bibinfo{author}{H.~Espinosa}, \bibinfo{author}{B.~Prorok},
  \bibinfo{author}{M.~Fischer},
\newblock \bibinfo{title}{A methodology for determining mechanical properties
  of freestanding thin films and mems materials},
\newblock \bibinfo{journal}{Journal of the Mechanics and Physics of Solids}
  \bibinfo{volume}{51} (\bibinfo{year}{2003}{\natexlab{a}})
  \bibinfo{pages}{47--67}.
\bibitem[{Espinosa et~al.(2003{\natexlab{b}})Espinosa, Prorok, Peng, Kim,
  Moldovan, Auciello, Carlisle, Gruen, and Mancini}]{RN252}
\bibinfo{author}{H.~Espinosa}, \bibinfo{author}{B.~Prorok},
  \bibinfo{author}{B.~Peng}, \bibinfo{author}{K.~Kim},
  \bibinfo{author}{N.~Moldovan}, \bibinfo{author}{O.~Auciello},
  \bibinfo{author}{J.~Carlisle}, \bibinfo{author}{D.~Gruen},
  \bibinfo{author}{D.~Mancini},
\newblock \bibinfo{title}{Mechanical properties of ultrananocrystalline diamond
  thin films relevant to mems/nems devices},
\newblock \bibinfo{journal}{Experimental Mechanics} \bibinfo{volume}{43}
  (\bibinfo{year}{2003}{\natexlab{b}}) \bibinfo{pages}{256--268}.
\bibitem[{Espinosa et~al.(2003{\natexlab{c}})Espinosa, Peng, Prorok, Moldovan,
  Auciello, Carlisle, Gruen, and Mancini}]{RN255}
\bibinfo{author}{H.~Espinosa}, \bibinfo{author}{B.~Peng},
  \bibinfo{author}{B.~Prorok}, \bibinfo{author}{N.~Moldovan},
  \bibinfo{author}{O.~Auciello}, \bibinfo{author}{J.~Carlisle},
  \bibinfo{author}{D.~Gruen}, \bibinfo{author}{D.~Mancini},
\newblock \bibinfo{title}{Fracture strength of ultrananocrystalline diamond
  thin films—identification of weibull parameters},
\newblock \bibinfo{journal}{Journal of Applied Physics} \bibinfo{volume}{94}
  (\bibinfo{year}{2003}{\natexlab{c}}) \bibinfo{pages}{6076--6084}.
\bibitem[{Espinosa et~al.(2004)Espinosa, Prorok, and Peng}]{RN254}
\bibinfo{author}{H.~Espinosa}, \bibinfo{author}{B.~Prorok},
  \bibinfo{author}{B.~Peng},
\newblock \bibinfo{title}{Plasticity size effects in free-standing submicron
  polycrystalline fcc films subjected to pure tension},
\newblock \bibinfo{journal}{Journal of the Mechanics and Physics of Solids}
  \bibinfo{volume}{52} (\bibinfo{year}{2004}) \bibinfo{pages}{667--689}.
\bibitem[{Pugno et~al.(2005)Pugno, Peng, and Espinosa}]{RN256}
\bibinfo{author}{N.~Pugno}, \bibinfo{author}{B.~Peng},
  \bibinfo{author}{H.~Espinosa},
\newblock \bibinfo{title}{Predictions of strength in mems components with
  defects––a novel experimental–theoretical approach},
\newblock \bibinfo{journal}{International journal of solids and structures}
  \bibinfo{volume}{42} (\bibinfo{year}{2005}) \bibinfo{pages}{647--661}.
\bibitem[{Freund and Suresh(2004)}]{RN257}
\bibinfo{author}{L.~B. Freund}, \bibinfo{author}{S.~Suresh},
  \bibinfo{title}{Thin film materials: stress, defect formation and surface
  evolution}, \bibinfo{publisher}{Cambridge university press},
  \bibinfo{year}{2004}.
\bibitem[{Greer and Nix(2006)}]{greer2006nanoscale}
\bibinfo{author}{J.~R. Greer}, \bibinfo{author}{W.~D. Nix},
\newblock \bibinfo{title}{Nanoscale gold pillars strengthened through
  dislocation starvation},
\newblock \bibinfo{journal}{Physical Review B} \bibinfo{volume}{73}
  (\bibinfo{year}{2006}) \bibinfo{pages}{245410}.
\bibitem[{Chen et~al.(2023)Chen, Sologubenko, and Wheeler}]{chen2023exploring}
\bibinfo{author}{M.~Chen}, \bibinfo{author}{A.~S. Sologubenko},
  \bibinfo{author}{J.~M. Wheeler},
\newblock \bibinfo{title}{Exploring defect behavior and size effects in
  micron-scale germanium from cryogenic to elevated temperatures},
\newblock \bibinfo{journal}{Matter}  (\bibinfo{year}{2023}).
\bibitem[{Greer et~al.(2006)Greer, Oliver, and Nix}]{RN59}
\bibinfo{author}{J.~R. Greer}, \bibinfo{author}{W.~C. Oliver},
  \bibinfo{author}{W.~D. Nix},
\newblock \bibinfo{title}{Size dependence in mechanical properties of gold at
  the micron scale in the absence of strain gradients (vol 53, pg 1821, 2005)},
\newblock \bibinfo{journal}{Acta Materialia} \bibinfo{volume}{54}
  (\bibinfo{year}{2006}) \bibinfo{pages}{1705--1705}.
  \DOIprefix\doi{10.1016/j.actamat.2005.12.004}.
\bibitem[{Zhao et~al.(2022)Zhao, Li, Gao, and Lu}]{RN60}
\bibinfo{author}{H.~Zhao}, \bibinfo{author}{Z.~Li}, \bibinfo{author}{H.~Gao},
  \bibinfo{author}{L.~Lu},
\newblock \bibinfo{title}{Fracture and toughening mechanisms in nanotwinned and
  nanolayered materials},
\newblock \bibinfo{journal}{Mrs Bulletin} \bibinfo{volume}{47}
  (\bibinfo{year}{2022}) \bibinfo{pages}{839--847}.
  \DOIprefix\doi{10.1557/s43577-022-00376-5}.
\bibitem[{Jin et~al.(2018)Jin, Zhou, and Chen}]{RN223}
\bibinfo{author}{H.~Jin}, \bibinfo{author}{J.~Zhou}, \bibinfo{author}{Y.~Chen},
\newblock \bibinfo{title}{Grain size gradient and length scale effect on
  mechanical behaviors of surface nanocrystalline metals},
\newblock \bibinfo{journal}{Materials Science and Engineering: A}
  \bibinfo{volume}{725} (\bibinfo{year}{2018}) \bibinfo{pages}{1--7}.
\bibitem[{Greer and De~Hosson(2011)}]{RN258}
\bibinfo{author}{J.~R. Greer}, \bibinfo{author}{J.~T.~M. De~Hosson},
\newblock \bibinfo{title}{Plasticity in small-sized metallic systems: Intrinsic
  versus extrinsic size effect},
\newblock \bibinfo{journal}{Progress in Materials Science} \bibinfo{volume}{56}
  (\bibinfo{year}{2011}) \bibinfo{pages}{654--724}.
\bibitem[{Li et~al.(2020)Li, Lu, Li, Zhang, and Gao}]{RN259}
\bibinfo{author}{X.~Li}, \bibinfo{author}{L.~Lu}, \bibinfo{author}{J.~Li},
  \bibinfo{author}{X.~Zhang}, \bibinfo{author}{H.~Gao},
\newblock \bibinfo{title}{Mechanical properties and deformation mechanisms of
  gradient nanostructured metals and alloys},
\newblock \bibinfo{journal}{Nature Reviews Materials} \bibinfo{volume}{5}
  (\bibinfo{year}{2020}) \bibinfo{pages}{706--723}.
\bibitem[{Jin and Zhou(2017)}]{jin2017distribution}
\bibinfo{author}{H.~Jin}, \bibinfo{author}{J.~Zhou},
\newblock \bibinfo{title}{Distribution effects of secondary twin lamellae on
  the global and local behavior of hierarchically nanotwinned metals},
\newblock \bibinfo{journal}{Journal of Materials Science} \bibinfo{volume}{52}
  (\bibinfo{year}{2017}) \bibinfo{pages}{4647--4657}.
\bibitem[{Lee et~al.(2009)Lee, Roh, and Park}]{RN55}
\bibinfo{author}{B.~Lee}, \bibinfo{author}{S.~Roh}, \bibinfo{author}{J.~Park},
\newblock \bibinfo{title}{Current status of micro- and nano-structured optical
  fiber sensors},
\newblock \bibinfo{journal}{Optical Fiber Technology} \bibinfo{volume}{15}
  (\bibinfo{year}{2009}) \bibinfo{pages}{209--221}.
  \DOIprefix\doi{10.1016/j.yofte.2009.02.006}.
\bibitem[{Ramachandramoorthy et~al.(2017)Ramachandramoorthy, Wang, Aghaei,
  Richter, Cai, and Espinosa}]{RN260}
\bibinfo{author}{R.~Ramachandramoorthy}, \bibinfo{author}{Y.~Wang},
  \bibinfo{author}{A.~Aghaei}, \bibinfo{author}{G.~Richter},
  \bibinfo{author}{W.~Cai}, \bibinfo{author}{H.~D. Espinosa},
\newblock \bibinfo{title}{Reliability of single crystal silver nanowire-based
  systems: stress assisted instabilities},
\newblock \bibinfo{journal}{ACS nano} \bibinfo{volume}{11}
  (\bibinfo{year}{2017}) \bibinfo{pages}{4768--4776}.
\bibitem[{Bernal et~al.(2015)Bernal, Aghaei, Lee, Ryu, Sohn, Huang, Cai, and
  Espinosa}]{RN262}
\bibinfo{author}{R.~A. Bernal}, \bibinfo{author}{A.~Aghaei},
  \bibinfo{author}{S.~Lee}, \bibinfo{author}{S.~Ryu},
  \bibinfo{author}{K.~Sohn}, \bibinfo{author}{J.~Huang},
  \bibinfo{author}{W.~Cai}, \bibinfo{author}{H.~Espinosa},
\newblock \bibinfo{title}{Intrinsic bauschinger effect and recoverable
  plasticity in pentatwinned silver nanowires tested in tension},
\newblock \bibinfo{journal}{Nano letters} \bibinfo{volume}{15}
  (\bibinfo{year}{2015}) \bibinfo{pages}{139--146}.
\bibitem[{Filleter et~al.(2012)Filleter, Ryu, Kang, Yin, Bernal, Sohn, Li,
  Huang, Cai, and Espinosa}]{RN263}
\bibinfo{author}{T.~Filleter}, \bibinfo{author}{S.~Ryu},
  \bibinfo{author}{K.~Kang}, \bibinfo{author}{J.~Yin}, \bibinfo{author}{R.~A.
  Bernal}, \bibinfo{author}{K.~Sohn}, \bibinfo{author}{S.~Li},
  \bibinfo{author}{J.~Huang}, \bibinfo{author}{W.~Cai}, \bibinfo{author}{H.~D.
  Espinosa},
\newblock \bibinfo{title}{Nucleation‐controlled distributed plasticity in
  penta‐twinned silver nanowires},
\newblock \bibinfo{journal}{Small} \bibinfo{volume}{8} (\bibinfo{year}{2012})
  \bibinfo{pages}{2986--2993}.
\bibitem[{Espinosa et~al.(2013)Espinosa, Bernal, and Filleter}]{RN264}
\bibinfo{author}{H.~D. Espinosa}, \bibinfo{author}{R.~A. Bernal},
  \bibinfo{author}{T.~Filleter},
\newblock \bibinfo{title}{In‐situ tem electromechanical testing of nanowires
  and nanotubes},
\newblock \bibinfo{journal}{Nano and Cell Mechanics: Fundamentals and
  Frontiers}  (\bibinfo{year}{2013}) \bibinfo{pages}{191--226}.
\bibitem[{Akinwande et~al.(2017)Akinwande, Brennan, Bunch, Egberts, Felts, Gao,
  Huang, Kim, Li, Li, Liechti, Lu, Park, Reed, Wang, Yakobson, Zhang, Zhang,
  Zhou, and Zhu}]{RN56}
\bibinfo{author}{D.~Akinwande}, \bibinfo{author}{C.~J. Brennan},
  \bibinfo{author}{J.~S. Bunch}, \bibinfo{author}{P.~Egberts},
  \bibinfo{author}{J.~R. Felts}, \bibinfo{author}{H.~J. Gao},
  \bibinfo{author}{R.~Huang}, \bibinfo{author}{J.~S. Kim},
  \bibinfo{author}{T.~Li}, \bibinfo{author}{Y.~Li}, \bibinfo{author}{K.~M.
  Liechti}, \bibinfo{author}{N.~S. Lu}, \bibinfo{author}{H.~S. Park},
  \bibinfo{author}{E.~J. Reed}, \bibinfo{author}{P.~Wang},
  \bibinfo{author}{B.~I. Yakobson}, \bibinfo{author}{T.~Zhang},
  \bibinfo{author}{Y.~W. Zhang}, \bibinfo{author}{Y.~Zhou},
  \bibinfo{author}{Y.~Zhu},
\newblock \bibinfo{title}{A review on mechanics and mechanical properties of 2d
  materials-graphene and beyond},
\newblock \bibinfo{journal}{Extreme Mechanics Letters} \bibinfo{volume}{13}
  (\bibinfo{year}{2017}) \bibinfo{pages}{42--77}.
  \DOIprefix\doi{10.1016/j.eml.2017.01.008}.
\bibitem[{Wei et~al.(2016)Wei, Meng, Ruiz, Xia, Lee, Kysar, Hone, Keten, and
  Espinosa}]{RN266}
\bibinfo{author}{X.~Wei}, \bibinfo{author}{Z.~Meng}, \bibinfo{author}{L.~Ruiz},
  \bibinfo{author}{W.~Xia}, \bibinfo{author}{C.~Lee}, \bibinfo{author}{J.~W.
  Kysar}, \bibinfo{author}{J.~C. Hone}, \bibinfo{author}{S.~Keten},
  \bibinfo{author}{H.~D. Espinosa},
\newblock \bibinfo{title}{Recoverable slippage mechanism in multilayer graphene
  leads to repeatable energy dissipation},
\newblock \bibinfo{journal}{ACS nano} \bibinfo{volume}{10}
  (\bibinfo{year}{2016}) \bibinfo{pages}{1820--1828}.
\bibitem[{Soler-Crespo et~al.(2019)Soler-Crespo, Mao, Wen, Nguyen, Zhang, Wei,
  Huang, Nguyen, and Espinosa}]{RN267}
\bibinfo{author}{R.~A. Soler-Crespo}, \bibinfo{author}{L.~Mao},
  \bibinfo{author}{J.~Wen}, \bibinfo{author}{H.~T. Nguyen},
  \bibinfo{author}{X.~Zhang}, \bibinfo{author}{X.~Wei},
  \bibinfo{author}{J.~Huang}, \bibinfo{author}{S.~T. Nguyen},
  \bibinfo{author}{H.~D. Espinosa},
\newblock \bibinfo{title}{Atomically thin polymer layer enhances toughness of
  graphene oxide monolayers},
\newblock \bibinfo{journal}{Matter} \bibinfo{volume}{1} (\bibinfo{year}{2019})
  \bibinfo{pages}{369--388}.
\bibitem[{Choi et~al.(2021)Choi, Zhang, Nguyen, Roenbeck, Mao, Soler-Crespo,
  Nguyen, and Espinosa}]{RN268}
\bibinfo{author}{J.~Y. Choi}, \bibinfo{author}{X.~Zhang},
  \bibinfo{author}{H.~T. Nguyen}, \bibinfo{author}{M.~R. Roenbeck},
  \bibinfo{author}{L.~Mao}, \bibinfo{author}{R.~Soler-Crespo},
  \bibinfo{author}{S.~T. Nguyen}, \bibinfo{author}{H.~D. Espinosa},
\newblock \bibinfo{title}{Atomistic mechanisms of adhesion and shear strength
  in graphene oxide-polymer interfaces},
\newblock \bibinfo{journal}{Journal of the Mechanics and Physics of Solids}
  \bibinfo{volume}{156} (\bibinfo{year}{2021}) \bibinfo{pages}{104578}.
\bibitem[{Yang et~al.(2021)Yang, Song, Lu, Zhang, Zhang, Ni, Wang, Li, Gu, Xie
  et~al.}]{yang2021intrinsic}
\bibinfo{author}{Y.~Yang}, \bibinfo{author}{Z.~Song}, \bibinfo{author}{G.~Lu},
  \bibinfo{author}{Q.~Zhang}, \bibinfo{author}{B.~Zhang},
  \bibinfo{author}{B.~Ni}, \bibinfo{author}{C.~Wang}, \bibinfo{author}{X.~Li},
  \bibinfo{author}{L.~Gu}, \bibinfo{author}{X.~Xie}, et~al.,
\newblock \bibinfo{title}{Intrinsic toughening and stable crack propagation in
  hexagonal boron nitride},
\newblock \bibinfo{journal}{Nature} \bibinfo{volume}{594}
  (\bibinfo{year}{2021}) \bibinfo{pages}{57--61}.
\bibitem[{Lin et~al.(2020)Lin, Novelino, Wei, Alderete, Paulino, Espinosa, and
  Krishnaswamy}]{RN265}
\bibinfo{author}{Z.~Lin}, \bibinfo{author}{L.~S. Novelino},
  \bibinfo{author}{H.~Wei}, \bibinfo{author}{N.~A. Alderete},
  \bibinfo{author}{G.~H. Paulino}, \bibinfo{author}{H.~D. Espinosa},
  \bibinfo{author}{S.~Krishnaswamy},
\newblock \bibinfo{title}{Folding at the microscale: Enabling multifunctional
  3d origami‐architected metamaterials},
\newblock \bibinfo{journal}{Small} \bibinfo{volume}{16} (\bibinfo{year}{2020})
  \bibinfo{pages}{2002229}.
\bibitem[{Bauer et~al.(2017)Bauer, Meza, Schaedler, Schwaiger, Zheng, and
  Valdevit}]{RN57}
\bibinfo{author}{J.~Bauer}, \bibinfo{author}{L.~R. Meza},
  \bibinfo{author}{T.~A. Schaedler}, \bibinfo{author}{R.~Schwaiger},
  \bibinfo{author}{X.~Y. Zheng}, \bibinfo{author}{L.~Valdevit},
\newblock \bibinfo{title}{Nanolattices: An emerging class of mechanical
  metamaterials},
\newblock \bibinfo{journal}{Advanced Materials} \bibinfo{volume}{29}
  (\bibinfo{year}{2017}). \DOIprefix\doi{ARTN 1701850 10.1002/adma.201701850}.
\bibitem[{Vyatskikh et~al.(2018)Vyatskikh, Delalande, Kudo, Zhang, Portela, and
  Greer}]{RN270}
\bibinfo{author}{A.~Vyatskikh}, \bibinfo{author}{S.~Delalande},
  \bibinfo{author}{A.~Kudo}, \bibinfo{author}{X.~Zhang}, \bibinfo{author}{C.~M.
  Portela}, \bibinfo{author}{J.~R. Greer},
\newblock \bibinfo{title}{Additive manufacturing of 3d nano-architected
  metals},
\newblock \bibinfo{journal}{Nature communications} \bibinfo{volume}{9}
  (\bibinfo{year}{2018}) \bibinfo{pages}{593}.
\bibitem[{Meza et~al.(2014)Meza, Das, and Greer}]{RN271}
\bibinfo{author}{L.~R. Meza}, \bibinfo{author}{S.~Das}, \bibinfo{author}{J.~R.
  Greer},
\newblock \bibinfo{title}{Strong, lightweight, and recoverable
  three-dimensional ceramic nanolattices},
\newblock \bibinfo{journal}{Science} \bibinfo{volume}{345}
  (\bibinfo{year}{2014}) \bibinfo{pages}{1322--1326}.
\bibitem[{Bang et~al.(2009)Bang, Jeong, Ryu, Russell, and Hawker}]{RN58}
\bibinfo{author}{J.~Bang}, \bibinfo{author}{U.~Jeong}, \bibinfo{author}{D.~Y.
  Ryu}, \bibinfo{author}{T.~P. Russell}, \bibinfo{author}{C.~J. Hawker},
\newblock \bibinfo{title}{Block copolymer nanolithography: Translation of
  molecular level control to nanoscale patterns},
\newblock \bibinfo{journal}{Advanced Materials} \bibinfo{volume}{21}
  (\bibinfo{year}{2009}) \bibinfo{pages}{4769--4792}.
  \DOIprefix\doi{10.1002/adma.200803302}.
\bibitem[{Jin et~al.(2022)Jin, Machnicki, Hegarty, Clifton, and Kim}]{RN221}
\bibinfo{author}{H.~Jin}, \bibinfo{author}{C.~Machnicki},
  \bibinfo{author}{J.~Hegarty}, \bibinfo{author}{R.~J. Clifton},
  \bibinfo{author}{K.-S. Kim},
\newblock \bibinfo{title}{Understanding the nanoscale deformation mechanisms of
  polyurea from in situ afm tensile experiments},
\newblock in: \bibinfo{booktitle}{Challenges in Mechanics of Time Dependent
  Materials, Mechanics of Biological Systems and Materials \& Micro-and
  Nanomechanics, Volume 2: Proceedings of the 2021 Annual Conference \&
  Exposition on Experimental and Applied Mechanics},
  \bibinfo{publisher}{Springer}, \bibinfo{year}{2022}, pp.
  \bibinfo{pages}{45--51}.
\bibitem[{Kim et~al.(2021)Kim, Jin, Jiao, and Clifton}]{RN222}
\bibinfo{author}{K.-S. Kim}, \bibinfo{author}{H.~Jin},
  \bibinfo{author}{T.~Jiao}, \bibinfo{author}{R.~J. Clifton},
\newblock \bibinfo{title}{Dynamic fracture-toughness testing of a
  hierarchically nano-structured solid},
\newblock in: \bibinfo{booktitle}{Fracture, Fatigue, Failure and Damage
  Evolution, Volume 3: Proceedings of the 2020 Annual Conference on
  Experimental and Applied Mechanics}, \bibinfo{publisher}{Springer},
  \bibinfo{year}{2021}, pp. \bibinfo{pages}{97--100}.
\bibitem[{Xia et~al.(2009)Xia, Qi, Perry, and Kim}]{RN61}
\bibinfo{author}{S.~Xia}, \bibinfo{author}{Y.~Qi}, \bibinfo{author}{T.~Perry},
  \bibinfo{author}{K.-S. Kim},
\newblock \bibinfo{title}{Strength characterization of al/si interfaces: A
  hybrid method of nanoindentation and finite element analysis},
\newblock \bibinfo{journal}{Acta Materialia} \bibinfo{volume}{57}
  (\bibinfo{year}{2009}) \bibinfo{pages}{695--707}.
  \DOIprefix\doi{10.1016/j.actamat.2008.10.011}.
\bibitem[{Shen et~al.(2001)Shen, Williams, Piotrowski, Chawla, and Guo}]{RN79}
\bibinfo{author}{Y.-L. Shen}, \bibinfo{author}{J.~Williams},
  \bibinfo{author}{G.~Piotrowski}, \bibinfo{author}{N.~Chawla},
  \bibinfo{author}{Y.~Guo},
\newblock \bibinfo{title}{Correlation between tensile and indentation behavior
  of particle-reinforced metal matrix composites: An experimental and numerical
  study},
\newblock \bibinfo{journal}{Acta Materialia} \bibinfo{volume}{49}
  (\bibinfo{year}{2001}) \bibinfo{pages}{3219--3229}.
  \DOIprefix\doi{10.1016/S1359-6454(01)00226-9}.
\bibitem[{Nix and Gao(1998)}]{nix1998indentation}
\bibinfo{author}{W.~D. Nix}, \bibinfo{author}{H.~Gao},
\newblock \bibinfo{title}{Indentation size effects in crystalline materials: a
  law for strain gradient plasticity},
\newblock \bibinfo{journal}{Journal of the Mechanics and Physics of Solids}
  \bibinfo{volume}{46} (\bibinfo{year}{1998}) \bibinfo{pages}{411--425}.
\bibitem[{Papakyriakou et~al.(2022)Papakyriakou, Lu, and
  Xia}]{papakyriakou2022nanoindentation}
\bibinfo{author}{M.~Papakyriakou}, \bibinfo{author}{M.~Lu},
  \bibinfo{author}{S.~Xia},
\newblock \bibinfo{title}{Nanoindentation size effects in lithiated and
  sodiated battery electrode materials},
\newblock \bibinfo{journal}{Journal of Applied Mechanics} \bibinfo{volume}{89}
  (\bibinfo{year}{2022}) \bibinfo{pages}{071007}.
\bibitem[{Oliver and Pharr(1992)}]{oliver1992improved}
\bibinfo{author}{W.~C. Oliver}, \bibinfo{author}{G.~M. Pharr},
\newblock \bibinfo{title}{An improved technique for determining hardness and
  elastic modulus using load and displacement sensing indentation experiments},
\newblock \bibinfo{journal}{Journal of materials research} \bibinfo{volume}{7}
  (\bibinfo{year}{1992}) \bibinfo{pages}{1564--1583}.
\bibitem[{Oliver and Pharr(2004)}]{oliver2004measurement}
\bibinfo{author}{W.~C. Oliver}, \bibinfo{author}{G.~M. Pharr},
\newblock \bibinfo{title}{Measurement of hardness and elastic modulus by
  instrumented indentation: Advances in understanding and refinements to
  methodology},
\newblock \bibinfo{journal}{Journal of materials research} \bibinfo{volume}{19}
  (\bibinfo{year}{2004}) \bibinfo{pages}{3--20}.
\bibitem[{Alkorta et~al.(2005)Alkorta, Martinez-Esnaola, and
  Sevillano}]{alkorta2005absence}
\bibinfo{author}{J.~Alkorta}, \bibinfo{author}{J.~Martinez-Esnaola},
  \bibinfo{author}{J.~G. Sevillano},
\newblock \bibinfo{title}{Absence of one-to-one correspondence between
  elastoplastic properties and sharp-indentation load--penetration data},
\newblock \bibinfo{journal}{Journal of materials research} \bibinfo{volume}{20}
  (\bibinfo{year}{2005}) \bibinfo{pages}{432--437}.
\bibitem[{Chen et~al.(2007)Chen, Ogasawara, Zhao, and
  Chiba}]{chen2007uniqueness}
\bibinfo{author}{X.~Chen}, \bibinfo{author}{N.~Ogasawara},
  \bibinfo{author}{M.~Zhao}, \bibinfo{author}{N.~Chiba},
\newblock \bibinfo{title}{On the uniqueness of measuring elastoplastic
  properties from indentation: the indistinguishable mystical materials},
\newblock \bibinfo{journal}{Journal of the Mechanics and Physics of Solids}
  \bibinfo{volume}{55} (\bibinfo{year}{2007}) \bibinfo{pages}{1618--1660}.
\bibitem[{Cheng and Cheng(1999)}]{cheng1999can}
\bibinfo{author}{Y.-T. Cheng}, \bibinfo{author}{C.-M. Cheng},
\newblock \bibinfo{title}{Can stress--strain relationships be obtained from
  indentation curves using conical and pyramidal indenters?},
\newblock \bibinfo{journal}{Journal of Materials Research} \bibinfo{volume}{14}
  (\bibinfo{year}{1999}) \bibinfo{pages}{3493--3496}.
\bibitem[{Bhowmick et~al.(2019)Bhowmick, Espinosa, Jungjohann, Pardoen, and
  Pierron}]{RN248}
\bibinfo{author}{S.~Bhowmick}, \bibinfo{author}{H.~Espinosa},
  \bibinfo{author}{K.~Jungjohann}, \bibinfo{author}{T.~Pardoen},
  \bibinfo{author}{O.~Pierron},
\newblock \bibinfo{title}{Advanced microelectromechanical systems-based
  nanomechanical testing: beyond stress and strain measurements},
\newblock \bibinfo{journal}{Mrs Bulletin} \bibinfo{volume}{44}
  (\bibinfo{year}{2019}) \bibinfo{pages}{487--493}.
\bibitem[{Needleman et~al.(2015)Needleman, Tvergaard, and Van~der
  Giessen}]{needleman2015indentation}
\bibinfo{author}{A.~Needleman}, \bibinfo{author}{V.~Tvergaard},
  \bibinfo{author}{E.~Van~der Giessen},
\newblock \bibinfo{title}{Indentation of elastically soft and plastically
  compressible solids},
\newblock \bibinfo{journal}{Acta Mechanica Sinica} \bibinfo{volume}{31}
  (\bibinfo{year}{2015}) \bibinfo{pages}{473--480}.
\bibitem[{Bower et~al.(1993)Bower, Fleck, Needleman, and
  Ogbonna}]{bower1993indentation}
\bibinfo{author}{A.~Bower}, \bibinfo{author}{N.~A. Fleck},
  \bibinfo{author}{A.~Needleman}, \bibinfo{author}{N.~Ogbonna},
\newblock \bibinfo{title}{Indentation of a power law creeping solid},
\newblock \bibinfo{journal}{Proceedings of the Royal Society of London. Series
  A: Mathematical and Physical Sciences} \bibinfo{volume}{441}
  (\bibinfo{year}{1993}) \bibinfo{pages}{97--124}.
\bibitem[{Lee et~al.(2019)Lee, Huen, Vimonsatit, and Mendis}]{RN67}
\bibinfo{author}{H.~Lee}, \bibinfo{author}{W.~Y. Huen},
  \bibinfo{author}{V.~Vimonsatit}, \bibinfo{author}{P.~Mendis},
\newblock \bibinfo{title}{An investigation of nanomechanical properties of
  materials using nanoindentation and artificial neural network},
\newblock \bibinfo{journal}{Scientific Reports} \bibinfo{volume}{9}
  (\bibinfo{year}{2019}). \DOIprefix\doi{10.1038/s41598-019-49780-z}.
\bibitem[{Konstantopoulos et~al.(2020)Konstantopoulos, Koumoulos, and
  Charitidis}]{RN68}
\bibinfo{author}{G.~Konstantopoulos}, \bibinfo{author}{E.~P. Koumoulos},
  \bibinfo{author}{C.~A. Charitidis},
\newblock \bibinfo{title}{Classification of mechanism of reinforcement in the
  fiber-matrix interface: Application of machine learning on nanoindentation
  data},
\newblock \bibinfo{journal}{Materials \& Design} \bibinfo{volume}{192}
  (\bibinfo{year}{2020}). \DOIprefix\doi{10.1016/j.matdes.2020.108705}.
\bibitem[{Weng et~al.(2020)Weng, Lindvall, Zhuang, St{\aa}hl, Ding, and
  Zhou}]{RN69}
\bibinfo{author}{J.~Weng}, \bibinfo{author}{R.~Lindvall},
  \bibinfo{author}{K.~Zhuang}, \bibinfo{author}{J.-E. St{\aa}hl},
  \bibinfo{author}{H.~Ding}, \bibinfo{author}{J.~Zhou},
\newblock \bibinfo{title}{A machine learning based approach for determining the
  stress-strain relation of grey cast iron from nanoindentation},
\newblock \bibinfo{journal}{Mechanics of Materials} \bibinfo{volume}{148}
  (\bibinfo{year}{2020}). \DOIprefix\doi{10.1016/j.mechmat.2020.103522}.
\bibitem[{Haj-Ali et~al.(2008)Haj-Ali, Kim, Koh, Saxena, and Tummala}]{RN70}
\bibinfo{author}{R.~Haj-Ali}, \bibinfo{author}{H.~K. Kim},
  \bibinfo{author}{S.~W. Koh}, \bibinfo{author}{A.~Saxena},
  \bibinfo{author}{R.~Tummala},
\newblock \bibinfo{title}{Nonlinear constitutive models from nanoindentation
  tests using artificial neural networks},
\newblock \bibinfo{journal}{International Journal of Plasticity}
  \bibinfo{volume}{24} (\bibinfo{year}{2008}) \bibinfo{pages}{371--396}.
  \DOIprefix\doi{10.1016/j.ijplas.2007.02.001}.
\bibitem[{Han et~al.(2022)Han, Marimuthu, and Lee}]{RN71}
\bibinfo{author}{G.~Han}, \bibinfo{author}{K.~P. Marimuthu},
  \bibinfo{author}{H.~Lee},
\newblock \bibinfo{title}{Evaluation of thin film material properties using a
  deep nanoindentation and ann},
\newblock \bibinfo{journal}{Materials \& Design} \bibinfo{volume}{221}
  (\bibinfo{year}{2022}). \DOIprefix\doi{10.1016/j.matdes.2022.111000}.
\bibitem[{Jeong et~al.(2022)Jeong, Lee, Lee, Kang, Jung, Lee, Kwak, Kwon, and
  Han}]{RN72}
\bibinfo{author}{K.~Jeong}, \bibinfo{author}{K.~Lee}, \bibinfo{author}{S.~Lee},
  \bibinfo{author}{S.~G. Kang}, \bibinfo{author}{J.~Jung},
  \bibinfo{author}{H.~Lee}, \bibinfo{author}{N.~Kwak},
  \bibinfo{author}{D.~Kwon}, \bibinfo{author}{H.~N. Han},
\newblock \bibinfo{title}{Deep learning-based indentation plastometry in
  anisotropic materials},
\newblock \bibinfo{journal}{International Journal of Plasticity}
  \bibinfo{volume}{157} (\bibinfo{year}{2022}).
  \DOIprefix\doi{10.1016/j.ijplas.2022.103403}.
\bibitem[{Li et~al.(2016)Li, Gutierrez, Toda, Kuwazuru, Liu, Hangai, Kobayashi,
  and Batres}]{RN73}
\bibinfo{author}{H.~Li}, \bibinfo{author}{L.~Gutierrez},
  \bibinfo{author}{H.~Toda}, \bibinfo{author}{O.~Kuwazuru},
  \bibinfo{author}{W.~L. Liu}, \bibinfo{author}{Y.~Hangai},
  \bibinfo{author}{M.~Kobayashi}, \bibinfo{author}{R.~Batres},
\newblock \bibinfo{title}{Identification of material properties using
  nanoindentation and surrogate modeling},
\newblock \bibinfo{journal}{International Journal of Solids and Structures}
  \bibinfo{volume}{81} (\bibinfo{year}{2016}) \bibinfo{pages}{151--159}.
  \DOIprefix\doi{10.1016/j.ijsolstr.2015.11.022}.
\bibitem[{Kim et~al.(2022)Kim, Gu, Asghari-Rad, Noh, Rho, Seo, and Kim}]{RN74}
\bibinfo{author}{Y.~Kim}, \bibinfo{author}{G.~H. Gu},
  \bibinfo{author}{P.~Asghari-Rad}, \bibinfo{author}{J.~Noh},
  \bibinfo{author}{J.~Rho}, \bibinfo{author}{M.~H. Seo}, \bibinfo{author}{H.~S.
  Kim},
\newblock \bibinfo{title}{Novel deep learning approach for practical
  applications of indentation},
\newblock \bibinfo{journal}{Materials Today Advances} \bibinfo{volume}{13}
  (\bibinfo{year}{2022}). \DOIprefix\doi{10.1016/j.mtadv.2022.100207}.
\bibitem[{Xia et~al.(2022)Xia, Won, Kim, Lee, and Yoon}]{RN75}
\bibinfo{author}{J.~Xia}, \bibinfo{author}{C.~Won}, \bibinfo{author}{H.~Kim},
  \bibinfo{author}{W.~Lee}, \bibinfo{author}{J.~Yoon},
\newblock \bibinfo{title}{Artificial neural networks for predicting plastic
  anisotropy of sheet metals based on indentation test},
\newblock \bibinfo{journal}{Materials} \bibinfo{volume}{15}
  (\bibinfo{year}{2022}). \DOIprefix\doi{10.3390/ma15051714}.
\bibitem[{Tyulyukovskiy and Huber(2006)}]{RN76}
\bibinfo{author}{E.~Tyulyukovskiy}, \bibinfo{author}{N.~Huber},
\newblock \bibinfo{title}{Identification of viscoplastic material parameters
  from spherical indentation data: Part i. neural networks},
\newblock \bibinfo{journal}{Journal of Materials Research} \bibinfo{volume}{21}
  (\bibinfo{year}{2006}) \bibinfo{pages}{664--676}.
  \DOIprefix\doi{10.1557/Jmr.2006.0076}.
\bibitem[{Meng and Karniadakis(2020)}]{RN78}
\bibinfo{author}{X.~Meng}, \bibinfo{author}{G.~E. Karniadakis},
\newblock \bibinfo{title}{A composite neural network that learns from
  multi-fidelity data: Application to function approximation and inverse pde
  problems},
\newblock \bibinfo{journal}{Journal of Computational Physics}
  \bibinfo{volume}{401} (\bibinfo{year}{2020}).
  \DOIprefix\doi{10.1016/j.jcp.2019.109020}.
\bibitem[{Bayes(1763)}]{RN80}
\bibinfo{author}{T.~Bayes},
\newblock \bibinfo{title}{Lii. an essay towards solving a problem in the
  doctrine of chances. by the late rev. mr. bayes, frs communicated by mr.
  price, in a letter to john canton, amfr s},
\newblock \bibinfo{journal}{Philosophical transactions of the Royal Society of
  London}  (\bibinfo{year}{1763}) \bibinfo{pages}{370--418}.
\bibitem[{Castillo and Kalidindi(2019)}]{RN83}
\bibinfo{author}{A.~Castillo}, \bibinfo{author}{S.~R. Kalidindi},
\newblock \bibinfo{title}{A bayesian framework for the estimation of the single
  crystal elastic parameters from spherical indentation stress-strain
  measurements},
\newblock \bibinfo{journal}{Frontiers in Materials} \bibinfo{volume}{6}
  (\bibinfo{year}{2019}). \DOIprefix\doi{10.3389/fmats.2019.00136}.
\bibitem[{Wang and Wu(2019)}]{RN82}
\bibinfo{author}{M.~Wang}, \bibinfo{author}{J.~Wu},
\newblock \bibinfo{title}{Identification of plastic properties of metal
  materials using spherical indentation experiment and bayesian model updating
  approach},
\newblock \bibinfo{journal}{International Journal of Mechanical Sciences}
  \bibinfo{volume}{151} (\bibinfo{year}{2019}) \bibinfo{pages}{733--745}.
  \DOIprefix\doi{10.1016/j.ijmecsci.2018.12.027}.
\bibitem[{Zhang and Needleman(2020)}]{RN86}
\bibinfo{author}{Y.~Zhang}, \bibinfo{author}{A.~Needleman},
\newblock \bibinfo{title}{Influence of assumed strain hardening relation on
  plastic stress-strain response identification from conical indentation},
\newblock \bibinfo{journal}{Journal of Engineering Materials and Technology}
  \bibinfo{volume}{142} (\bibinfo{year}{2020}).
\bibitem[{Zhang and Needleman(2021)}]{RN87}
\bibinfo{author}{Y.~Zhang}, \bibinfo{author}{A.~Needleman},
\newblock \bibinfo{title}{On the identification of power-law creep parameters
  from conical indentation},
\newblock \bibinfo{journal}{Proceedings of the Royal Society A}
  \bibinfo{volume}{477} (\bibinfo{year}{2021}).
  \DOIprefix\doi{10.1098/rspa.2021.0233}.
\bibitem[{Asgari et~al.(2022)Asgari, Latifi, Giovanniello, Espinosa, and
  Amabili}]{RN90}
\bibinfo{author}{M.~Asgari}, \bibinfo{author}{N.~Latifi},
  \bibinfo{author}{F.~Giovanniello}, \bibinfo{author}{H.~D. Espinosa},
  \bibinfo{author}{M.~Amabili},
\newblock \bibinfo{title}{Revealing layer-specific ultrastructure and
  nanomechanics of fibrillar collagen in human aorta via atomic force
  microscopy testing: Implications on tissue mechanics at macroscopic scale},
\newblock \bibinfo{journal}{Advanced Nanobiomed Research} \bibinfo{volume}{2}
  (\bibinfo{year}{2022}). \DOIprefix\doi{10.1002/anbr.202100159}.
\bibitem[{Broussard et~al.(2017)Broussard, Yang, Huang, Nathamgari, Beese,
  Godsel, Hegazy, Lee, Zhou, and Sniadecki}]{RN249}
\bibinfo{author}{J.~A. Broussard}, \bibinfo{author}{R.~Yang},
  \bibinfo{author}{C.~Huang}, \bibinfo{author}{S.~S.~P. Nathamgari},
  \bibinfo{author}{A.~M. Beese}, \bibinfo{author}{L.~M. Godsel},
  \bibinfo{author}{M.~H. Hegazy}, \bibinfo{author}{S.~Lee},
  \bibinfo{author}{F.~Zhou}, \bibinfo{author}{N.~J. Sniadecki},
\newblock \bibinfo{title}{The desmoplakin–intermediate filament linkage
  regulates cell mechanics},
\newblock \bibinfo{journal}{Molecular biology of the cell} \bibinfo{volume}{28}
  (\bibinfo{year}{2017}) \bibinfo{pages}{3156--3164}.
\bibitem[{Broussard et~al.(2020)Broussard, Jaiganesh, Zarkoob, Conway, Dunn,
  Espinosa, Janmey, and Green}]{RN250}
\bibinfo{author}{J.~A. Broussard}, \bibinfo{author}{A.~Jaiganesh},
  \bibinfo{author}{H.~Zarkoob}, \bibinfo{author}{D.~E. Conway},
  \bibinfo{author}{A.~R. Dunn}, \bibinfo{author}{H.~D. Espinosa},
  \bibinfo{author}{P.~A. Janmey}, \bibinfo{author}{K.~J. Green},
\newblock \bibinfo{title}{Scaling up single-cell mechanics to multicellular
  tissues–the role of the intermediate filament–desmosome network},
\newblock \bibinfo{journal}{Journal of cell science} \bibinfo{volume}{133}
  (\bibinfo{year}{2020}) \bibinfo{pages}{jcs228031}.
\bibitem[{Rajabifar et~al.(2022)Rajabifar, Meyers, Wagner, and Raman}]{RN93}
\bibinfo{author}{B.~Rajabifar}, \bibinfo{author}{G.~F. Meyers},
  \bibinfo{author}{R.~Wagner}, \bibinfo{author}{A.~Raman},
\newblock \bibinfo{title}{Machine learning approach to characterize the
  adhesive and mechanical properties of soft polymers using peakforce tapping
  afm},
\newblock \bibinfo{journal}{Macromolecules} \bibinfo{volume}{55}
  (\bibinfo{year}{2022}) \bibinfo{pages}{8731--8740}.
  \DOIprefix\doi{10.1021/acs.macromol.2c00147}.
\bibitem[{Nguyen and Liu(2022)}]{RN92}
\bibinfo{author}{L.~P. Nguyen}, \bibinfo{author}{B.~Liu},
\newblock \bibinfo{title}{Machine learning approach for reducing uncertainty in
  afm nanomechanical measurements through selection of appropriate contact
  model},
\newblock \bibinfo{journal}{European Journal of Mechanics a-Solids}
  \bibinfo{volume}{94} (\bibinfo{year}{2022}).
  \DOIprefix\doi{10.1016/j.euromechsol.2022.104579}.
\bibitem[{Chan et~al.(2020)Chan, Cherukara, Loeffler, Narayanan, and
  Sankaranarayanan}]{RN225}
\bibinfo{author}{H.~Chan}, \bibinfo{author}{M.~Cherukara},
  \bibinfo{author}{T.~D. Loeffler}, \bibinfo{author}{B.~Narayanan},
  \bibinfo{author}{S.~K. Sankaranarayanan},
\newblock \bibinfo{title}{Machine learning enabled autonomous microstructural
  characterization in 3d samples},
\newblock \bibinfo{journal}{npj Computational Materials} \bibinfo{volume}{6}
  (\bibinfo{year}{2020}) \bibinfo{pages}{1}.
\bibitem[{Bostanabad et~al.(2018)Bostanabad, Zhang, Li, Kearney, Brinson,
  Apley, Liu, and Chen}]{RN226}
\bibinfo{author}{R.~Bostanabad}, \bibinfo{author}{Y.~Zhang},
  \bibinfo{author}{X.~Li}, \bibinfo{author}{T.~Kearney}, \bibinfo{author}{L.~C.
  Brinson}, \bibinfo{author}{D.~W. Apley}, \bibinfo{author}{W.~K. Liu},
  \bibinfo{author}{W.~Chen},
\newblock \bibinfo{title}{Computational microstructure characterization and
  reconstruction: Review of the state-of-the-art techniques},
\newblock \bibinfo{journal}{Progress in Materials Science} \bibinfo{volume}{95}
  (\bibinfo{year}{2018}) \bibinfo{pages}{1--41}.
\bibitem[{Ge et~al.(2020)Ge, Su, Zhao, and Su}]{RN230}
\bibinfo{author}{M.~Ge}, \bibinfo{author}{F.~Su}, \bibinfo{author}{Z.~Zhao},
  \bibinfo{author}{D.~Su},
\newblock \bibinfo{title}{Deep learning analysis on microscopic imaging in
  materials science},
\newblock \bibinfo{journal}{Materials Today Nano} \bibinfo{volume}{11}
  (\bibinfo{year}{2020}) \bibinfo{pages}{100087}.
\bibitem[{Chowdhury et~al.(2016)Chowdhury, Kautz, Yener, and Lewis}]{RN235}
\bibinfo{author}{A.~Chowdhury}, \bibinfo{author}{E.~Kautz},
  \bibinfo{author}{B.~Yener}, \bibinfo{author}{D.~Lewis},
\newblock \bibinfo{title}{Image driven machine learning methods for
  microstructure recognition},
\newblock \bibinfo{journal}{Computational Materials Science}
  \bibinfo{volume}{123} (\bibinfo{year}{2016}) \bibinfo{pages}{176--187}.
\bibitem[{DeCost et~al.(2019)DeCost, Lei, Francis, and Holm}]{RN228}
\bibinfo{author}{B.~L. DeCost}, \bibinfo{author}{B.~Lei},
  \bibinfo{author}{T.~Francis}, \bibinfo{author}{E.~A. Holm},
\newblock \bibinfo{title}{High throughput quantitative metallography for
  complex microstructures using deep learning: A case study in ultrahigh carbon
  steel},
\newblock \bibinfo{journal}{Microscopy and Microanalysis} \bibinfo{volume}{25}
  (\bibinfo{year}{2019}) \bibinfo{pages}{21--29}.
\bibitem[{Chen and Daly(2018)}]{chen2018deformation}
\bibinfo{author}{Z.~Chen}, \bibinfo{author}{S.~Daly},
\newblock \bibinfo{title}{Deformation twin identification in magnesium through
  clustering and computer vision},
\newblock \bibinfo{journal}{Materials Science and Engineering: A}
  \bibinfo{volume}{736} (\bibinfo{year}{2018}) \bibinfo{pages}{61--75}.
\bibitem[{Durmaz et~al.(2021)Durmaz, M{\"u}ller, Lei, Thomas, Britz, Holm,
  Eberl, M{\"u}cklich, and Gumbsch}]{durmaz2021deep}
\bibinfo{author}{A.~R. Durmaz}, \bibinfo{author}{M.~M{\"u}ller},
  \bibinfo{author}{B.~Lei}, \bibinfo{author}{A.~Thomas},
  \bibinfo{author}{D.~Britz}, \bibinfo{author}{E.~A. Holm},
  \bibinfo{author}{C.~Eberl}, \bibinfo{author}{F.~M{\"u}cklich},
  \bibinfo{author}{P.~Gumbsch},
\newblock \bibinfo{title}{A deep learning approach for complex microstructure
  inference},
\newblock \bibinfo{journal}{Nature communications} \bibinfo{volume}{12}
  (\bibinfo{year}{2021}) \bibinfo{pages}{6272}.
\bibitem[{Stuckner et~al.(2022)Stuckner, Harder, and
  Smith}]{stuckner2022microstructure}
\bibinfo{author}{J.~Stuckner}, \bibinfo{author}{B.~Harder},
  \bibinfo{author}{T.~M. Smith},
\newblock \bibinfo{title}{Microstructure segmentation with deep learning
  encoders pre-trained on a large microscopy dataset},
\newblock \bibinfo{journal}{npj Computational Materials} \bibinfo{volume}{8}
  (\bibinfo{year}{2022}) \bibinfo{pages}{200}.
\bibitem[{Liu et~al.(2020)Liu, Wu, Paul, He, Peng, Gludovatz, Kruzic, Wang, and
  Li}]{RN232}
\bibinfo{author}{Q.~Liu}, \bibinfo{author}{H.~Wu}, \bibinfo{author}{M.~J.
  Paul}, \bibinfo{author}{P.~He}, \bibinfo{author}{Z.~Peng},
  \bibinfo{author}{B.~Gludovatz}, \bibinfo{author}{J.~J. Kruzic},
  \bibinfo{author}{C.~H. Wang}, \bibinfo{author}{X.~Li},
\newblock \bibinfo{title}{Machine-learning assisted laser powder bed fusion
  process optimization for alsi10mg: New microstructure description indices and
  fracture mechanisms},
\newblock \bibinfo{journal}{Acta Materialia} \bibinfo{volume}{201}
  (\bibinfo{year}{2020}) \bibinfo{pages}{316--328}.
\bibitem[{Müller et~al.(2021)Müller, Karathanasopoulos, Roth, and
  Mohr}]{RN236}
\bibinfo{author}{A.~Müller}, \bibinfo{author}{N.~Karathanasopoulos},
  \bibinfo{author}{C.~C. Roth}, \bibinfo{author}{D.~Mohr},
\newblock \bibinfo{title}{Machine learning classifiers for surface crack
  detection in fracture experiments},
\newblock \bibinfo{journal}{International Journal of Mechanical Sciences}
  \bibinfo{volume}{209} (\bibinfo{year}{2021}) \bibinfo{pages}{106698}.
\bibitem[{Hashemi and Kalidindi(2021)}]{RN233}
\bibinfo{author}{S.~Hashemi}, \bibinfo{author}{S.~R. Kalidindi},
\newblock \bibinfo{title}{A machine learning framework for the temporal
  evolution of microstructure during static recrystallization of
  polycrystalline materials simulated by cellular automaton},
\newblock \bibinfo{journal}{Computational Materials Science}
  \bibinfo{volume}{188} (\bibinfo{year}{2021}) \bibinfo{pages}{110132}.
\bibitem[{Li and Li(2022)}]{RN234}
\bibinfo{author}{Y.~Li}, \bibinfo{author}{S.~Li},
\newblock \bibinfo{title}{Deep learning based phase transformation model for
  the prediction of microstructure and mechanical properties of hot-stamped
  parts},
\newblock \bibinfo{journal}{International Journal of Mechanical Sciences}
  \bibinfo{volume}{220} (\bibinfo{year}{2022}) \bibinfo{pages}{107134}.
\bibitem[{Alberi et~al.(2019)Alberi, Nardelli, Zakutayev, Mitas, Curtarolo,
  Jain, Fornari, Marzari, Takeuchi, Green, Kanatzidis, Toney, Butenko, Meredig,
  Lany, Kattner, Davydov, Toberer, Stevanovic, Walsh, Park, Aspuru-Guzik,
  Tabor, Nelson, Murphy, Setlur, Gregoire, Li, Xiao, Ludwig, Martin, Rappe,
  Wei, and Perkins}]{RN29}
\bibinfo{author}{K.~Alberi}, \bibinfo{author}{M.~B. Nardelli},
  \bibinfo{author}{A.~Zakutayev}, \bibinfo{author}{L.~Mitas},
  \bibinfo{author}{S.~Curtarolo}, \bibinfo{author}{A.~Jain},
  \bibinfo{author}{M.~Fornari}, \bibinfo{author}{N.~Marzari},
  \bibinfo{author}{I.~Takeuchi}, \bibinfo{author}{M.~L. Green},
  \bibinfo{author}{M.~Kanatzidis}, \bibinfo{author}{M.~F. Toney},
  \bibinfo{author}{S.~Butenko}, \bibinfo{author}{B.~Meredig},
  \bibinfo{author}{S.~Lany}, \bibinfo{author}{U.~Kattner},
  \bibinfo{author}{A.~Davydov}, \bibinfo{author}{E.~S. Toberer},
  \bibinfo{author}{V.~Stevanovic}, \bibinfo{author}{A.~Walsh},
  \bibinfo{author}{N.~G. Park}, \bibinfo{author}{A.~Aspuru-Guzik},
  \bibinfo{author}{D.~P. Tabor}, \bibinfo{author}{J.~Nelson},
  \bibinfo{author}{J.~Murphy}, \bibinfo{author}{A.~Setlur},
  \bibinfo{author}{J.~Gregoire}, \bibinfo{author}{H.~Li},
  \bibinfo{author}{R.~J. Xiao}, \bibinfo{author}{A.~Ludwig},
  \bibinfo{author}{L.~W. Martin}, \bibinfo{author}{A.~M. Rappe},
  \bibinfo{author}{S.~H. Wei}, \bibinfo{author}{J.~Perkins},
\newblock \bibinfo{title}{The 2019 materials by design roadmap},
\newblock \bibinfo{journal}{Journal of Physics D-Applied Physics}
  \bibinfo{volume}{52} (\bibinfo{year}{2019}).
  \DOIprefix\doi{10.1088/1361-6463/aad926}.
\bibitem[{Zhang et~al.(2019)Zhang, Vyatskikh, Gao, Greer, and Li}]{RN34}
\bibinfo{author}{X.~Zhang}, \bibinfo{author}{A.~Vyatskikh},
  \bibinfo{author}{H.~J. Gao}, \bibinfo{author}{J.~R. Greer},
  \bibinfo{author}{X.~Y. Li},
\newblock \bibinfo{title}{Lightweight, flaw-tolerant, and ultrastrong
  nanoarchitected carbon},
\newblock \bibinfo{journal}{Proceedings of the National Academy of Sciences of
  the United States of America} \bibinfo{volume}{116} (\bibinfo{year}{2019})
  \bibinfo{pages}{6665--6672}. \DOIprefix\doi{10.1073/pnas.1817309116}.
\bibitem[{Portela et~al.(2020)Portela, Vidyasagar, Krodel, Weissenbach, Yee,
  Greer, and Kochmann}]{RN33}
\bibinfo{author}{C.~M. Portela}, \bibinfo{author}{A.~Vidyasagar},
  \bibinfo{author}{S.~Krodel}, \bibinfo{author}{T.~Weissenbach},
  \bibinfo{author}{D.~W. Yee}, \bibinfo{author}{J.~R. Greer},
  \bibinfo{author}{D.~M. Kochmann},
\newblock \bibinfo{title}{Extreme mechanical resilience of self-assembled
  nanolabyrinthine materials},
\newblock \bibinfo{journal}{Proceedings of the National Academy of Sciences of
  the United States of America} \bibinfo{volume}{117} (\bibinfo{year}{2020})
  \bibinfo{pages}{5686--5693}. \DOIprefix\doi{10.1073/pnas.1916817117}.
\bibitem[{Portela et~al.(2021)Portela, Edwards, Veysset, Sun, Nelson, Kochmann,
  and Greer}]{RN32}
\bibinfo{author}{C.~M. Portela}, \bibinfo{author}{B.~W. Edwards},
  \bibinfo{author}{D.~Veysset}, \bibinfo{author}{Y.~C. Sun},
  \bibinfo{author}{K.~A. Nelson}, \bibinfo{author}{D.~M. Kochmann},
  \bibinfo{author}{J.~R. Greer},
\newblock \bibinfo{title}{Supersonic impact resilience of nanoarchitected
  carbon},
\newblock \bibinfo{journal}{Nature Materials} \bibinfo{volume}{20}
  (\bibinfo{year}{2021}) \bibinfo{pages}{1491--+}.
  \DOIprefix\doi{10.1038/s41563-021-01033-z}.
\bibitem[{Gu et~al.(2018)Gu, Chen, Richmond, and Buehler}]{RN41}
\bibinfo{author}{G.~X. Gu}, \bibinfo{author}{C.~T. Chen},
  \bibinfo{author}{D.~J. Richmond}, \bibinfo{author}{M.~J. Buehler},
\newblock \bibinfo{title}{Bioinspired hierarchical composite design using
  machine learning: simulation, additive manufacturing, and experiment},
\newblock \bibinfo{journal}{Materials Horizons} \bibinfo{volume}{5}
  (\bibinfo{year}{2018}) \bibinfo{pages}{939--945}.
  \DOIprefix\doi{10.1039/c8mh00653a}.
\bibitem[{Bastek et~al.(2022)Bastek, Kumar, Telgen, Glaesener, and
  Kochmann}]{RN43}
\bibinfo{author}{J.~H. Bastek}, \bibinfo{author}{S.~Kumar},
  \bibinfo{author}{B.~Telgen}, \bibinfo{author}{R.~N. Glaesener},
  \bibinfo{author}{D.~M. Kochmann},
\newblock \bibinfo{title}{Inverting the structure-property map of truss
  metamaterials by deep learning},
\newblock \bibinfo{journal}{Proceedings of the National Academy of Sciences of
  the United States of America} \bibinfo{volume}{119} (\bibinfo{year}{2022}).
  \DOIprefix\doi{10.1073/pnas.2111505119}.
\bibitem[{Wu et~al.(2020)Wu, Liu, Wang, Zhai, Zhuang, Krishnaraju, Wang, and
  Jiang}]{RN44}
\bibinfo{author}{L.~Wu}, \bibinfo{author}{L.~Liu}, \bibinfo{author}{Y.~Wang},
  \bibinfo{author}{Z.~Zhai}, \bibinfo{author}{H.~Zhuang},
  \bibinfo{author}{D.~Krishnaraju}, \bibinfo{author}{Q.~Wang},
  \bibinfo{author}{H.~Jiang},
\newblock \bibinfo{title}{A machine learning -based method to design modular
  metamaterials},
\newblock \bibinfo{journal}{Extreme Mechanics Letters} \bibinfo{volume}{36}
  (\bibinfo{year}{2020}). \DOIprefix\doi{10.1016/j.eml.2020.100657}.
\bibitem[{Wang et~al.(2020)Wang, Chan, Ahmed, Liu, Zhu, and Chen}]{RN45}
\bibinfo{author}{L.~Wang}, \bibinfo{author}{Y.-C. Chan},
  \bibinfo{author}{F.~Ahmed}, \bibinfo{author}{Z.~Liu},
  \bibinfo{author}{P.~Zhu}, \bibinfo{author}{W.~Chen},
\newblock \bibinfo{title}{Deep generative modeling for mechanistic-based
  learning and design of metamaterial systems},
\newblock \bibinfo{journal}{Computer Methods in Applied Mechanics and
  Engineering} \bibinfo{volume}{372} (\bibinfo{year}{2020}).
  \DOIprefix\doi{10.1016/j.cma.2020.113377}.
\bibitem[{Cecen et~al.(2018)Cecen, Dai, Yabansu, Kalidindi, and Song}]{RN47}
\bibinfo{author}{A.~Cecen}, \bibinfo{author}{H.~J. Dai}, \bibinfo{author}{Y.~C.
  Yabansu}, \bibinfo{author}{S.~R. Kalidindi}, \bibinfo{author}{L.~Song},
\newblock \bibinfo{title}{Material structure-property linkages using
  three-dimensional convolutional neural networks},
\newblock \bibinfo{journal}{Acta Materialia} \bibinfo{volume}{146}
  (\bibinfo{year}{2018}) \bibinfo{pages}{76--84}.
  \DOIprefix\doi{10.1016/j.actamat.2017.11.053}.
\bibitem[{Deng et~al.(2022)Deng, Zareei, Ding, Weaver, Rycroft, and
  Bertoldi}]{RN51}
\bibinfo{author}{B.~Deng}, \bibinfo{author}{A.~Zareei},
  \bibinfo{author}{X.~Ding}, \bibinfo{author}{J.~C. Weaver},
  \bibinfo{author}{C.~H. Rycroft}, \bibinfo{author}{K.~Bertoldi},
\newblock \bibinfo{title}{Inverse design of mechanical metamaterials with
  target nonlinear response via a neural accelerated evolution strategy},
\newblock \bibinfo{journal}{Advanced Materials} \bibinfo{volume}{34}
  (\bibinfo{year}{2022}). \DOIprefix\doi{10.1002/adma.202206238}.
\bibitem[{Muhammad et~al.(2022)Muhammad, Kennedy, and Lim}]{RN37}
\bibinfo{author}{Muhammad}, \bibinfo{author}{J.~Kennedy},
  \bibinfo{author}{C.~W. Lim},
\newblock \bibinfo{title}{Machine learning and deep learning in phononic
  crystals and metamaterials-a review},
\newblock \bibinfo{journal}{Materials Today Communications}
  \bibinfo{volume}{33} (\bibinfo{year}{2022}).
  \DOIprefix\doi{10.1016/j.mtcomm.2022.104606}.
\bibitem[{Wang et~al.(2022)Wang, Zeng, Wang, Li, and Fang}]{RN50}
\bibinfo{author}{Y.~Wang}, \bibinfo{author}{Q.~Zeng},
  \bibinfo{author}{J.~Wang}, \bibinfo{author}{Y.~Li},
  \bibinfo{author}{D.~Fang},
\newblock \bibinfo{title}{Inverse design of shell-based mechanical metamaterial
  with customized loading curves based on machine learning and genetic
  algorithm},
\newblock \bibinfo{journal}{Computer Methods in Applied Mechanics and
  Engineering} \bibinfo{volume}{401} (\bibinfo{year}{2022}).
  \DOIprefix\doi{10.1016/j.cma.2022.115571}.
\bibitem[{Alderete et~al.(2021)Alderete, Medina, Lamberti, Sciammarella, and
  Espinosa}]{RN39}
\bibinfo{author}{N.~A. Alderete}, \bibinfo{author}{L.~Medina},
  \bibinfo{author}{L.~Lamberti}, \bibinfo{author}{C.~Sciammarella},
  \bibinfo{author}{H.~D. Espinosa},
\newblock \bibinfo{title}{Programmable 3d structures via kirigami engineering
  and controlled stretching},
\newblock \bibinfo{journal}{Extreme Mechanics Letters} \bibinfo{volume}{43}
  (\bibinfo{year}{2021}). \DOIprefix\doi{10.1016/j.eml.2020.101146}.
\bibitem[{Bessa et~al.(2019)Bessa, Glowacki, and Houlder}]{RN48}
\bibinfo{author}{M.~A. Bessa}, \bibinfo{author}{P.~Glowacki},
  \bibinfo{author}{M.~Houlder},
\newblock \bibinfo{title}{Bayesian machine learning in metamaterial design:
  Fragile becomes supercompressible},
\newblock \bibinfo{journal}{Advanced Materials} \bibinfo{volume}{31}
  (\bibinfo{year}{2019}). \DOIprefix\doi{10.1002/adma.201904845}.
\bibitem[{Gongora et~al.(2020)Gongora, Xu, Perry, Okoye, Riley, Reyes, Morgan,
  and Brown}]{gongora2020bayesian}
\bibinfo{author}{A.~E. Gongora}, \bibinfo{author}{B.~Xu},
  \bibinfo{author}{W.~Perry}, \bibinfo{author}{C.~Okoye},
  \bibinfo{author}{P.~Riley}, \bibinfo{author}{K.~G. Reyes},
  \bibinfo{author}{E.~F. Morgan}, \bibinfo{author}{K.~A. Brown},
\newblock \bibinfo{title}{A bayesian experimental autonomous researcher for
  mechanical design},
\newblock \bibinfo{journal}{Science advances} \bibinfo{volume}{6}
  (\bibinfo{year}{2020}) \bibinfo{pages}{eaaz1708}.
\bibitem[{Gongora et~al.(2021)Gongora, Snapp, Whiting, Riley, Reyes, Morgan,
  and Brown}]{gongora2021using}
\bibinfo{author}{A.~E. Gongora}, \bibinfo{author}{K.~L. Snapp},
  \bibinfo{author}{E.~Whiting}, \bibinfo{author}{P.~Riley},
  \bibinfo{author}{K.~G. Reyes}, \bibinfo{author}{E.~F. Morgan},
  \bibinfo{author}{K.~A. Brown},
\newblock \bibinfo{title}{Using simulation to accelerate autonomous
  experimentation: A case study using mechanics},
\newblock \bibinfo{journal}{Iscience} \bibinfo{volume}{24}
  (\bibinfo{year}{2021}) \bibinfo{pages}{102262}.
\bibitem[{Stach et~al.(2021)Stach, DeCost, Kusne, Hattrick-Simpers, Brown,
  Reyes, Schrier, Billinge, Buonassisi, Foster et~al.}]{stach2021autonomous}
\bibinfo{author}{E.~Stach}, \bibinfo{author}{B.~DeCost}, \bibinfo{author}{A.~G.
  Kusne}, \bibinfo{author}{J.~Hattrick-Simpers}, \bibinfo{author}{K.~A. Brown},
  \bibinfo{author}{K.~G. Reyes}, \bibinfo{author}{J.~Schrier},
  \bibinfo{author}{S.~Billinge}, \bibinfo{author}{T.~Buonassisi},
  \bibinfo{author}{I.~Foster}, et~al.,
\newblock \bibinfo{title}{Autonomous experimentation systems for materials
  development: A community perspective},
\newblock \bibinfo{journal}{Matter} \bibinfo{volume}{4} (\bibinfo{year}{2021})
  \bibinfo{pages}{2702--2726}.
\bibitem[{Lew and Buehler(2022)}]{RN49}
\bibinfo{author}{A.~J. Lew}, \bibinfo{author}{M.~J. Buehler},
\newblock \bibinfo{title}{Deepbuckle: Extracting physical behavior directly
  from empirical observation for a material agnostic approach to analyze and
  predict buckling},
\newblock \bibinfo{journal}{Journal of the Mechanics and Physics of Solids}
  \bibinfo{volume}{164} (\bibinfo{year}{2022}).
  \DOIprefix\doi{10.1016/j.jmps.2022.104909}.
\bibitem[{Akinwande et~al.(2019)Akinwande, Huyghebaert, Wang, Serna, Goossens,
  Li, Wong, and Koppens}]{akinwande2019graphene}
\bibinfo{author}{D.~Akinwande}, \bibinfo{author}{C.~Huyghebaert},
  \bibinfo{author}{C.-H. Wang}, \bibinfo{author}{M.~I. Serna},
  \bibinfo{author}{S.~Goossens}, \bibinfo{author}{L.-J. Li},
  \bibinfo{author}{H.-S.~P. Wong}, \bibinfo{author}{F.~H. Koppens},
\newblock \bibinfo{title}{Graphene and two-dimensional materials for silicon
  technology},
\newblock \bibinfo{journal}{Nature} \bibinfo{volume}{573}
  (\bibinfo{year}{2019}) \bibinfo{pages}{507--518}.
\bibitem[{Geim(2009)}]{geim2009graphene}
\bibinfo{author}{A.~K. Geim},
\newblock \bibinfo{title}{Graphene: status and prospects},
\newblock \bibinfo{journal}{science} \bibinfo{volume}{324}
  (\bibinfo{year}{2009}) \bibinfo{pages}{1530--1534}.
\bibitem[{Dong et~al.(2022)Dong, Zhang, Nathamgari, Krayev, Zhang, Hwang,
  Ajayan, and Espinosa}]{dong2022facile}
\bibinfo{author}{S.~Dong}, \bibinfo{author}{X.~Zhang},
  \bibinfo{author}{S.~S.~P. Nathamgari}, \bibinfo{author}{A.~Krayev},
  \bibinfo{author}{X.~Zhang}, \bibinfo{author}{J.~W. Hwang},
  \bibinfo{author}{P.~M. Ajayan}, \bibinfo{author}{H.~D. Espinosa},
\newblock \bibinfo{title}{Facile fabrication of 2d material multilayers and vdw
  heterostructures with multimodal microscopy and afm characterization},
\newblock \bibinfo{journal}{Materials Today} \bibinfo{volume}{52}
  (\bibinfo{year}{2022}) \bibinfo{pages}{31--42}.
\bibitem[{Ni et~al.(2022)Ni, Steinbach, Yang, Lew, Zhang, Fang, Buehler, and
  Lou}]{ni2022fracture}
\bibinfo{author}{B.~Ni}, \bibinfo{author}{D.~Steinbach},
  \bibinfo{author}{Z.~Yang}, \bibinfo{author}{A.~Lew},
  \bibinfo{author}{B.~Zhang}, \bibinfo{author}{Q.~Fang}, \bibinfo{author}{M.~J.
  Buehler}, \bibinfo{author}{J.~Lou},
\newblock \bibinfo{title}{Fracture at the two-dimensional limit},
\newblock \bibinfo{journal}{MRS Bulletin} \bibinfo{volume}{47}
  (\bibinfo{year}{2022}) \bibinfo{pages}{848--862}.
\bibitem[{Zhang et~al.(2014)Zhang, Ma, Fan, Zeng, Peng, Loya, Liu, Gong, Zhang,
  Zhang et~al.}]{zhang2014fracture}
\bibinfo{author}{P.~Zhang}, \bibinfo{author}{L.~Ma}, \bibinfo{author}{F.~Fan},
  \bibinfo{author}{Z.~Zeng}, \bibinfo{author}{C.~Peng}, \bibinfo{author}{P.~E.
  Loya}, \bibinfo{author}{Z.~Liu}, \bibinfo{author}{Y.~Gong},
  \bibinfo{author}{J.~Zhang}, \bibinfo{author}{X.~Zhang}, et~al.,
\newblock \bibinfo{title}{Fracture toughness of graphene},
\newblock \bibinfo{journal}{Nature communications} \bibinfo{volume}{5}
  (\bibinfo{year}{2014}) \bibinfo{pages}{3782}.
\bibitem[{Jiang et~al.(2013)Jiang, Park, and Rabczuk}]{jiang2013molecular}
\bibinfo{author}{J.-W. Jiang}, \bibinfo{author}{H.~S. Park},
  \bibinfo{author}{T.~Rabczuk},
\newblock \bibinfo{title}{Molecular dynamics simulations of single-layer
  molybdenum disulphide (mos2): Stillinger-weber parametrization, mechanical
  properties, and thermal conductivity},
\newblock \bibinfo{journal}{Journal of Applied Physics} \bibinfo{volume}{114}
  (\bibinfo{year}{2013}) \bibinfo{pages}{064307}.
\bibitem[{Ostadhossein et~al.(2017)Ostadhossein, Rahnamoun, Wang, Zhao, Zhang,
  Crespi, and Van~Duin}]{ostadhossein2017reaxff}
\bibinfo{author}{A.~Ostadhossein}, \bibinfo{author}{A.~Rahnamoun},
  \bibinfo{author}{Y.~Wang}, \bibinfo{author}{P.~Zhao},
  \bibinfo{author}{S.~Zhang}, \bibinfo{author}{V.~H. Crespi},
  \bibinfo{author}{A.~C. Van~Duin},
\newblock \bibinfo{title}{Reaxff reactive force-field study of molybdenum
  disulfide (mos2)},
\newblock \bibinfo{journal}{The journal of physical chemistry letters}
  \bibinfo{volume}{8} (\bibinfo{year}{2017}) \bibinfo{pages}{631--640}.
\bibitem[{Wen et~al.(2017)Wen, Shirodkar, Plech{\'a}{\v{c}}, Kaxiras, Elliott,
  and Tadmor}]{wen2017force}
\bibinfo{author}{M.~Wen}, \bibinfo{author}{S.~N. Shirodkar},
  \bibinfo{author}{P.~Plech{\'a}{\v{c}}}, \bibinfo{author}{E.~Kaxiras},
  \bibinfo{author}{R.~S. Elliott}, \bibinfo{author}{E.~B. Tadmor},
\newblock \bibinfo{title}{A force-matching stillinger-weber potential for mos2:
  Parameterization and fisher information theory based sensitivity analysis},
\newblock \bibinfo{journal}{Journal of Applied Physics} \bibinfo{volume}{122}
  (\bibinfo{year}{2017}) \bibinfo{pages}{244301}.
\bibitem[{Botu et~al.(2017)Botu, Batra, Chapman, and
  Ramprasad}]{botu2017machine}
\bibinfo{author}{V.~Botu}, \bibinfo{author}{R.~Batra},
  \bibinfo{author}{J.~Chapman}, \bibinfo{author}{R.~Ramprasad},
\newblock \bibinfo{title}{Machine learning force fields: construction,
  validation, and outlook},
\newblock \bibinfo{journal}{The Journal of Physical Chemistry C}
  \bibinfo{volume}{121} (\bibinfo{year}{2017}) \bibinfo{pages}{511--522}.
\bibitem[{Li et~al.(2017)Li, Li, Pickard~IV, Narayanan, Sen, Chan,
  Sankaranarayanan, Brooks, and Roux}]{li2017machine}
\bibinfo{author}{Y.~Li}, \bibinfo{author}{H.~Li}, \bibinfo{author}{F.~C.
  Pickard~IV}, \bibinfo{author}{B.~Narayanan}, \bibinfo{author}{F.~G. Sen},
  \bibinfo{author}{M.~K. Chan}, \bibinfo{author}{S.~K. Sankaranarayanan},
  \bibinfo{author}{B.~R. Brooks}, \bibinfo{author}{B.~Roux},
\newblock \bibinfo{title}{Machine learning force field parameters from ab
  initio data},
\newblock \bibinfo{journal}{Journal of chemical theory and computation}
  \bibinfo{volume}{13} (\bibinfo{year}{2017}) \bibinfo{pages}{4492--4503}.
\bibitem[{Shorten and Khoshgoftaar(2019)}]{shorten2019survey}
\bibinfo{author}{C.~Shorten}, \bibinfo{author}{T.~M. Khoshgoftaar},
\newblock \bibinfo{title}{A survey on image data augmentation for deep
  learning},
\newblock \bibinfo{journal}{Journal of big data} \bibinfo{volume}{6}
  (\bibinfo{year}{2019}) \bibinfo{pages}{1--48}.
\bibitem[{Abadi et~al.(2016)Abadi, Agarwal, Barham, Brevdo, Chen, Citro,
  Corrado, Davis, Dean, and Devin}]{RN237}
\bibinfo{author}{M.~Abadi}, \bibinfo{author}{A.~Agarwal},
  \bibinfo{author}{P.~Barham}, \bibinfo{author}{E.~Brevdo},
  \bibinfo{author}{Z.~Chen}, \bibinfo{author}{C.~Citro}, \bibinfo{author}{G.~S.
  Corrado}, \bibinfo{author}{A.~Davis}, \bibinfo{author}{J.~Dean},
  \bibinfo{author}{M.~Devin},
\newblock \bibinfo{title}{Tensorflow: Large-scale machine learning on
  heterogeneous distributed systems},
\newblock \bibinfo{journal}{arXiv preprint arXiv:1603.04467}
  (\bibinfo{year}{2016}).
\bibitem[{Paszke et~al.(2019)Paszke, Gross, Massa, Lerer, Bradbury, Chanan,
  Killeen, Lin, Gimelshein, and Antiga}]{RN239}
\bibinfo{author}{A.~Paszke}, \bibinfo{author}{S.~Gross},
  \bibinfo{author}{F.~Massa}, \bibinfo{author}{A.~Lerer},
  \bibinfo{author}{J.~Bradbury}, \bibinfo{author}{G.~Chanan},
  \bibinfo{author}{T.~Killeen}, \bibinfo{author}{Z.~Lin},
  \bibinfo{author}{N.~Gimelshein}, \bibinfo{author}{L.~Antiga},
\newblock \bibinfo{title}{Pytorch: An imperative style, high-performance deep
  learning library},
\newblock \bibinfo{journal}{Advances in neural information processing systems}
  \bibinfo{volume}{32} (\bibinfo{year}{2019}).
\bibitem[{Bradbury et~al.(2018)Bradbury, Frostig, Hawkins, Johnson, Leary,
  Maclaurin, Necula, Paszke, VanderPlas, Wanderman-Milne et~al.}]{RN280}
\bibinfo{author}{J.~Bradbury}, \bibinfo{author}{R.~Frostig},
  \bibinfo{author}{P.~Hawkins}, \bibinfo{author}{M.~Johnson},
  \bibinfo{author}{C.~Leary}, \bibinfo{author}{D.~Maclaurin},
  \bibinfo{author}{G.~Necula}, \bibinfo{author}{A.~Paszke},
  \bibinfo{author}{J.~VanderPlas}, \bibinfo{author}{S.~Wanderman-Milne},
  et~al.,
\newblock \bibinfo{title}{Jax: composable transformations of python+ numpy
  programs, v0. 2.5},
\newblock \bibinfo{journal}{Software available from https://github.
  com/google/jax}  (\bibinfo{year}{2018}).
\bibitem[{Kohavi et~al.(1995)}]{kohavi1995study}
\bibinfo{author}{R.~Kohavi}, et~al.,
\newblock \bibinfo{title}{A study of cross-validation and bootstrap for
  accuracy estimation and model selection},
\newblock in: \bibinfo{booktitle}{Ijcai}, volume~\bibinfo{volume}{14},
  \bibinfo{organization}{Montreal, Canada}, \bibinfo{year}{1995}, pp.
  \bibinfo{pages}{1137--1145}.
\bibitem[{Ngiam et~al.(2011)Ngiam, Khosla, Kim, Nam, Lee, and Ng}]{RN281}
\bibinfo{author}{J.~Ngiam}, \bibinfo{author}{A.~Khosla},
  \bibinfo{author}{M.~Kim}, \bibinfo{author}{J.~Nam}, \bibinfo{author}{H.~Lee},
  \bibinfo{author}{A.~Y. Ng},
\newblock \bibinfo{title}{Multimodal deep learning},
\newblock in: \bibinfo{booktitle}{Proceedings of the 28th international
  conference on machine learning (ICML-11)}, \bibinfo{year}{2011}, pp.
  \bibinfo{pages}{689--696}.
\bibitem[{Trask et~al.(2022)Trask, Martinez, Lee, and Boyce}]{RN282}
\bibinfo{author}{N.~Trask}, \bibinfo{author}{C.~Martinez},
  \bibinfo{author}{K.~Lee}, \bibinfo{author}{B.~Boyce},
\newblock \bibinfo{title}{Unsupervised physics-informed disentanglement of
  multimodal data for high-throughput scientific discovery},
\newblock \bibinfo{journal}{arXiv preprint arXiv:2202.03242}
  (\bibinfo{year}{2022}).
\bibitem[{Smith(2013)}]{smith2013uncertainty}
\bibinfo{author}{R.~C. Smith}, \bibinfo{title}{Uncertainty quantification:
  theory, implementation, and applications}, volume~\bibinfo{volume}{12},
  \bibinfo{publisher}{Siam}, \bibinfo{year}{2013}.
\bibitem[{Abdar et~al.(2021)Abdar, Pourpanah, Hussain, Rezazadegan, Liu,
  Ghavamzadeh, Fieguth, Cao, Khosravi, Acharya et~al.}]{abdar2021review}
\bibinfo{author}{M.~Abdar}, \bibinfo{author}{F.~Pourpanah},
  \bibinfo{author}{S.~Hussain}, \bibinfo{author}{D.~Rezazadegan},
  \bibinfo{author}{L.~Liu}, \bibinfo{author}{M.~Ghavamzadeh},
  \bibinfo{author}{P.~Fieguth}, \bibinfo{author}{X.~Cao},
  \bibinfo{author}{A.~Khosravi}, \bibinfo{author}{U.~R. Acharya}, et~al.,
\newblock \bibinfo{title}{A review of uncertainty quantification in deep
  learning: Techniques, applications and challenges},
\newblock \bibinfo{journal}{Information Fusion} \bibinfo{volume}{76}
  (\bibinfo{year}{2021}) \bibinfo{pages}{243--297}.
\bibitem[{Soize(2017)}]{soize2017uncertainty}
\bibinfo{author}{C.~Soize}, \bibinfo{title}{Uncertainty quantification},
  \bibinfo{publisher}{Springer}, \bibinfo{year}{2017}.
\bibitem[{Sullivan(2015)}]{sullivan2015introduction}
\bibinfo{author}{T.~J. Sullivan}, \bibinfo{title}{Introduction to uncertainty
  quantification}, volume~\bibinfo{volume}{63}, \bibinfo{publisher}{Springer},
  \bibinfo{year}{2015}.
\bibitem[{Cicci et~al.(2023)Cicci, Fresca, Guo, Manzoni, and
  Zunino}]{cicci2023uncertainty}
\bibinfo{author}{L.~Cicci}, \bibinfo{author}{S.~Fresca},
  \bibinfo{author}{M.~Guo}, \bibinfo{author}{A.~Manzoni},
  \bibinfo{author}{P.~Zunino},
\newblock \bibinfo{title}{Uncertainty quantification for nonlinear solid
  mechanics using reduced order models with gaussian process regression},
\newblock \bibinfo{journal}{arXiv preprint arXiv:2302.08216}
  (\bibinfo{year}{2023}).
\bibitem[{Liang et~al.(2022)Liang, Chang, Wan, Gan, Schlangen, and
  {\v{S}}avija}]{liang2022interpretable}
\bibinfo{author}{M.~Liang}, \bibinfo{author}{Z.~Chang},
  \bibinfo{author}{Z.~Wan}, \bibinfo{author}{Y.~Gan},
  \bibinfo{author}{E.~Schlangen}, \bibinfo{author}{B.~{\v{S}}avija},
\newblock \bibinfo{title}{Interpretable ensemble-machine-learning models for
  predicting creep behavior of concrete},
\newblock \bibinfo{journal}{Cement and Concrete Composites}
  \bibinfo{volume}{125} (\bibinfo{year}{2022}) \bibinfo{pages}{104295}.
\bibitem[{de~Oca~Zapiain et~al.(2022)de~Oca~Zapiain, Lim, Park, and
  Pourboghrat}]{de2022predicting}
\bibinfo{author}{D.~M. de~Oca~Zapiain}, \bibinfo{author}{H.~Lim},
  \bibinfo{author}{T.~Park}, \bibinfo{author}{F.~Pourboghrat},
\newblock \bibinfo{title}{Predicting plastic anisotropy using crystal
  plasticity and bayesian neural network surrogate models},
\newblock \bibinfo{journal}{Materials Science and Engineering: A}
  \bibinfo{volume}{833} (\bibinfo{year}{2022}) \bibinfo{pages}{142472}.
\bibitem[{Pyrialakos et~al.(2021)Pyrialakos, Kalogeris, Sotiropoulos, and
  Papadopoulos}]{pyrialakos2021neural}
\bibinfo{author}{S.~Pyrialakos}, \bibinfo{author}{I.~Kalogeris},
  \bibinfo{author}{G.~Sotiropoulos}, \bibinfo{author}{V.~Papadopoulos},
\newblock \bibinfo{title}{A neural network-aided bayesian identification
  framework for multiscale modeling of nanocomposites},
\newblock \bibinfo{journal}{Computer Methods in Applied Mechanics and
  Engineering} \bibinfo{volume}{384} (\bibinfo{year}{2021})
  \bibinfo{pages}{113937}.
\bibitem[{Nguyen and Kim(2021)}]{nguyen2021hybrid}
\bibinfo{author}{M.~S.~T. Nguyen}, \bibinfo{author}{S.-E. Kim},
\newblock \bibinfo{title}{A hybrid machine learning approach in prediction and
  uncertainty quantification of ultimate compressive strength of rcfst
  columns},
\newblock \bibinfo{journal}{Construction and Building Materials}
  \bibinfo{volume}{302} (\bibinfo{year}{2021}) \bibinfo{pages}{124208}.
\bibitem[{Huang et~al.(2022)Huang, Liu, Wu, and Chen}]{huang2022microstructure}
\bibinfo{author}{T.~Huang}, \bibinfo{author}{Z.~Liu}, \bibinfo{author}{C.~Wu},
  \bibinfo{author}{W.~Chen},
\newblock \bibinfo{title}{Microstructure-guided deep material network for rapid
  nonlinear material modeling and uncertainty quantification},
\newblock \bibinfo{journal}{Computer Methods in Applied Mechanics and
  Engineering} \bibinfo{volume}{398} (\bibinfo{year}{2022})
  \bibinfo{pages}{115197}.
\bibitem[{Huang et~al.(2020)Huang, Xu, Farhat, and Darve}]{RN283}
\bibinfo{author}{D.~Z. Huang}, \bibinfo{author}{K.~Xu},
  \bibinfo{author}{C.~Farhat}, \bibinfo{author}{E.~Darve},
\newblock \bibinfo{title}{Learning constitutive relations from indirect
  observations using deep neural networks},
\newblock \bibinfo{journal}{Journal of Computational Physics}
  \bibinfo{volume}{416} (\bibinfo{year}{2020}) \bibinfo{pages}{109491}.
\bibitem[{Ravindran et~al.(2022)Ravindran, Gandhi, Joshi, and
  Ravichandran}]{RN240}
\bibinfo{author}{S.~Ravindran}, \bibinfo{author}{V.~Gandhi},
  \bibinfo{author}{A.~Joshi}, \bibinfo{author}{G.~Ravichandran},
\newblock \bibinfo{title}{Three dimensional full-field velocity measurements in
  shock compression experiments using stereo digital image correlation},
\newblock \bibinfo{journal}{arXiv preprint arXiv:2210.12568}
  (\bibinfo{year}{2022}).
\bibitem[{Saccone et~al.(2022)Saccone, Gallivan, Narita, Yee, and
  Greer}]{RN241}
\bibinfo{author}{M.~A. Saccone}, \bibinfo{author}{R.~A. Gallivan},
  \bibinfo{author}{K.~Narita}, \bibinfo{author}{D.~W. Yee},
  \bibinfo{author}{J.~R. Greer},
\newblock \bibinfo{title}{Additive manufacturing of micro-architected metals
  via hydrogel infusion},
\newblock \bibinfo{journal}{Nature}  (\bibinfo{year}{2022})
  \bibinfo{pages}{1--2}.
\bibitem[{Kagias et~al.(2023)Kagias, Lee, Friedman, Zheng, Veysset, Faraon, and
  Greer}]{kagiasmetasurface}
\bibinfo{author}{M.~Kagias}, \bibinfo{author}{S.~Lee}, \bibinfo{author}{A.~C.
  Friedman}, \bibinfo{author}{T.~Zheng}, \bibinfo{author}{D.~Veysset},
  \bibinfo{author}{A.~Faraon}, \bibinfo{author}{J.~R. Greer},
\newblock \bibinfo{title}{Metasurface-enabled holographic lithography for
  impact-absorbing nano-architected sheets},
\newblock \bibinfo{journal}{Advanced Materials}  (\bibinfo{year}{2023})
  \bibinfo{pages}{2209153}.

\end{thebibliography}

\end{document}